\documentclass{article} % For LaTeX2e
\usepackage{iclr2027_conference,times}

\usepackage{hyperref}
\hypersetup{hidelinks}
\usepackage{booktabs,tabularx,graphicx}
\usepackage{amsmath,amssymb,amsthm,aliascnt}
\usepackage{microtype,enumitem,placeins}
\usepackage{xcolor}
\newcommand{\rev}[1]{{\color{black}#1}}
\usepackage{titletoc}
\newtheorem{theorem}{Theorem}
\newaliascnt{lemma}{theorem}
\newtheorem{lemma}[lemma]{Lemma}
\aliascntresetthe{lemma}
\newaliascnt{proposition}{theorem}
\newtheorem{proposition}[proposition]{Proposition}
\aliascntresetthe{proposition}
\newaliascnt{corollary}{theorem}
\newtheorem{corollary}[corollary]{Corollary}
\aliascntresetthe{corollary}
\theoremstyle{definition}
\newtheorem{definition}{Definition}
\usepackage[nameinlink,noabbrev]{cleveref}
\crefname{lemma}{lemma}{lemmas}
\Crefname{lemma}{Lemma}{Lemmas}
\crefname{proposition}{proposition}{propositions}
\Crefname{proposition}{Proposition}{Propositions}
\crefname{corollary}{corollary}{corollaries}
\Crefname{corollary}{Corollary}{Corollaries}
\crefname{appendix}{Appendix}{Appendices}
\Crefname{appendix}{Appendix}{Appendices}
\newcommand{\R}{\mathbb R}
\newcommand{\E}{\mathbb E}
\newcommand{\Prob}{\mathbb P}
\newcommand{\Sph}{\mathbb S}
\newcommand{\eps}{\varepsilon}
\newcommand{\ang}{\operatorname{angle}}
\newcommand{\sgn}{\operatorname{sign}}
\newcommand{\cond}{\operatorname{cond}}
\newcommand{\tr}{\operatorname{tr}}

\newcommand{\norm}[1]{\left\lVert #1\right\rVert}

\newcommand{\cC}{\mathcal C}
\newcommand{\cB}{\mathcal B}
\newcommand{\cQ}{\mathcal Q_{d,\kappa}}
\DeclareMathOperator*{\argmax}{arg\,max}
\DeclareMathOperator*{\argmin}{arg\,min}

\title{When Do Rankings Beat Signs? \\ Gradient Direction Recovery under \\ Even Nuisance}

\author{%
\makebox[\dimexpr\textwidth-2\tabcolsep\relax][c]{%
\begin{tabular}{@{}c@{}}
{\normalfont\normalsize
Zhuangzhuang Pan\textsuperscript{1}
\qquad
Yan Xia\textsuperscript{2,1}
\qquad
Chee Seng Chan\textsuperscript{1,3*}}
\\[0.8em]
\resizebox{0.96\textwidth}{!}{%
\normalfont\small
\textsuperscript{1}Universiti Malaya, Malaysia
\quad
\textsuperscript{2}Suzhou University of Technology, China
\quad
\textsuperscript{3}VinUniversity, Vietnam}
\\[0.5em]
{%
\normalfont\small
\{23072126,23078403\}@siswa.um.edu.my} \\
{\normalfont\small
\textsuperscript{*}Corresponding author: cs.chan@um.edu.my}
\end{tabular}%
}%
}

\iclrfinalcopy
\begin{document}
\maketitle

\lhead{Preprint}

\begin{abstract}
\rev{Comparison-based optimization relies on ordinal feedback to minimize an objective when numerical function values are unavailable. Existing sign-based baselines evaluate symmetric perturbation pairs in isolation, retaining only within-pair comparison signs and discarding all cross-pair orderings from the same ranked batch. In this paper, we ask when complete batch rankings provably outperform within-pair signs under an unknown differentiable even nuisance, such as quadratic curvature. We show that reversed cross-pair comparisons cancel shared nuisance offsets algebraically without estimating a Hessian, and introduce an algorithm-agnostic geometric framework characterizing three nested candidate direction sets, namely \emph{Mirror}, \emph{Reverse}, and \emph{Full}. Rather than total curvature magnitude or condition number, we identify the \emph{relative shell variation} $\tau_a$, defined as the nuisance oscillation range on a radius-$a$ sphere divided by $a\|g\|$ for gradient $g\ne 0$ in $d$ dimensions, as a key quantity governing one-round localization. For a target angular accuracy $\varepsilon$ in radians, whenever $\tau_a\le c\varepsilon/\sqrt d$ for a sufficiently small absolute constant $c$, one-round \emph{Reverse} constraints from independent uniform query directions localize every surviving direction within angle $\varepsilon$ of the true gradient using $\widetilde O(d+\sqrt{d/\varepsilon})$ endpoint evaluations, compared with $\Theta(d/\varepsilon)$ for sign-only recovery on the same class. This whole-cell guarantee requires no variation bound as input to the decoder and matches a lower bound up to logarithmic factors. Conversely, over well-conditioned positive-definite quadratics with unbounded curvature relative to the gradient scale, complete rankings offer no asymptotic minimax evaluation advantage over signs, even across multiple feedback rounds. Finally, experiments certify strict set separation between \emph{Full} and \emph{Reverse}. On two ridge-regression tasks under joint evaluation and feedback-round caps, although \emph{Full} does not establish an additional optimization benefit beyond reversed cross-pair constraints, \emph{Reverse} consistently attains lower terminal relative optimization gaps than tuned \emph{Mirror} and unit-normalized ranking-based zeroth-order stochastic gradient descent (ZO-RankSGD) in separate cohorts.}
\end{abstract}

%%%%%%%%%%%%%%%%%%%%%%%%%%%%%%%%%%%%%%%%%%%%%%%%%%%%%%%%
\section{Introduction}
\label{sec:introduction}

\begin{figure}[!t]
\centering
\includegraphics[width=\linewidth]{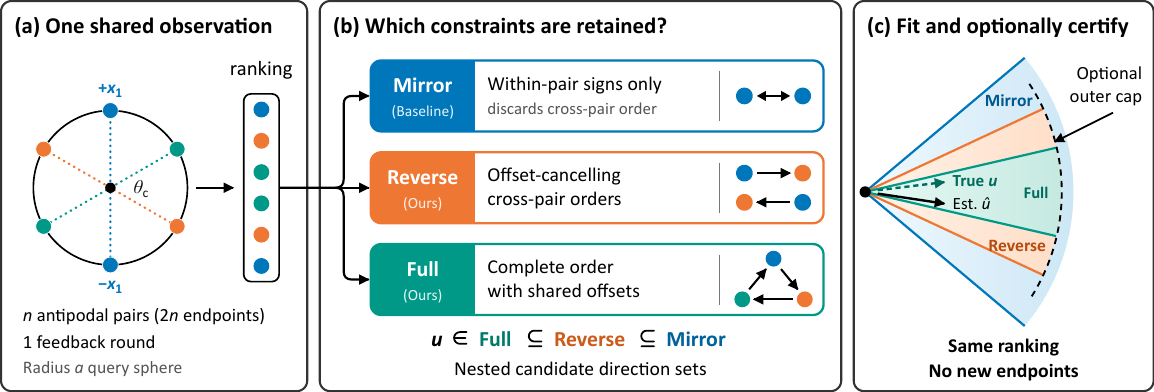}
\caption{One ranking, three uses of its constraints. (a) All methods receive the same ordering of $2n$ antipodal endpoints ($n$ opposite pairs) around center $\theta_c$ on a radius-$a$ query sphere in one feedback round. (b) Baseline Mirror keeps only within-pair signs, whereas Reverse adds offset-cancelling cross-pair orders, and Full enforces the complete ranking with shared offsets, forming nested candidate sets that contain $u$. The arrows represent information reuse, not additional queries. (c) Fitting selects a feasible estimate $\widehat{u}$ alongside the true direction $u$, and an optional outer cap encloses Reverse and hence Full. Wedges are schematic, may coincide, and do not depict measured accuracy or runtime.}
\label{fig:overview}
\end{figure}

\rev{In many black-box optimization and preference-learning applications, direct access to numerical objective values is unavailable, and the optimizer must rely solely on ordinal comparisons \citep{cai2022onebit,zhang2025zpg,ren2026riemannian,wang2026stationary}. When candidate points are evaluated in parallel batches, a ranking oracle returns their ordering in a single feedback round \citep{tang2024ranking,saha2024multiway}. In practice, query evaluations and synchronization rounds represent distinct resource budgets. Evaluating a larger batch consumes more queries in parallel, whereas adapting subsequent queries requires additional feedback rounds.

Despite recent progress in comparison-based optimization, existing methods face a sharp trade-off between feedback rounds and sample efficiency. To recover a local gradient direction in dimension $d$ within target angular error $\eps$ (in radians), sequential adaptive schemes attain the optimal $\Theta(d\log(1/\eps))$ comparison rate given supplied smoothness and gradient-norm bounds \citep{tao2026comparison}, but require many sequential feedback rounds. Conversely, single-round sign-based baselines \citep{lobanov2024acceleration,saha2025general} query symmetric antipodal perturbation pairs $\pm x_i$ around the current iterate and keep only within-pair signs, discarding all cross-pair comparisons returned in the same batch. The core obstacle is unknown nonlinear curvature. While opposite points cancel symmetric curvature within a single pair, comparing points across different pairs entangles gradient projections with unknown curvature differences. There remains a clear gap in understanding when cross-pair rankings can be exploited safely to outperform signs in a single feedback round.

In this paper, we ask when batch rankings provably outperform within-pair signs. At a fixed recovery center $\theta_c\in\R^d$ of an objective $F:\R^d\to\R$, we study the local function $f(x)=F(\theta_c+x)=b+g^\top x+E(x)$, where $x\in\R^d$ is a query displacement, $b=F(\theta_c)$, $g=\nabla F(\theta_c)\ne 0$, $u=g/\norm{g}$ is the target unit gradient direction, and $E$ is an unknown differentiable even nuisance ($E(-x)=E(x)$, $E(0)=0$). Every quadratic satisfies this decomposition identically without knowing the Hessian, and general smooth objectives fit it via calibrated odd-remainder bounds (\Cref{sec:calibration}). Within each pair, opposite points cancel $E$ ($f(x_i)-f(-x_i)=2g^\top x_i$), whereas comparing distinct pairs mixes $g^\top(x_i-x_j)$ with $E(x_i)-E(x_j)$. Our key insight is that reversed cross-pair orderings cancel this shared offset difference algebraically without estimating curvature. To formalize this mechanism, we organize ranking information into three nested candidate direction sets (\Cref{fig:overview}), where \emph{Mirror} retains within-pair signs, \emph{Reverse} adds offset-cancelling cross-pair comparisons, and \emph{Full} enforces the complete batch ranking with shared pair offsets.

Answering when rankings beat signs requires identifying the right measure of curvature. Neither curvature magnitude nor condition number governs this transition. Arbitrarily large isotropic curvature causes zero interference across a sphere, whereas arbitrarily well-conditioned quadratics with condition number $\kappa$ close to $1$ can eliminate all ranking gains. Instead, we characterize the favorable regime through the \emph{relative shell variation} $\tau_a$, measuring the oscillation of $E$ across a radius-$a$ sphere relative to the linear signal $a\norm{g}$. Whenever $\tau_a\le c\eps/\sqrt{d}$ for a small universal constant $c$, one-round \emph{Reverse} alone breaks the $\Theta(d/\eps)$ sign-only barrier down to $\widetilde O(d+\sqrt{d/\eps})$ evaluations for whole-cell recovery, without supplying $\tau_a$ to the decoder. Conversely, without controlling relative curvature scale, complete rankings lose any uniform minimax evaluation advantage over signs, even across adaptive rounds. Their practical trade-off also depends on constraint count, solver runtime, and tuning.}

\rev{Our key contributions are summarized below.
\begin{enumerate}[leftmargin=*,itemsep=2pt,topsep=2pt]
\item We characterize \emph{Mirror}, \emph{Reverse}, and \emph{Full} as nested consistency sets on a pair-offset difference graph, proving how reversed cross-pair orders cancel unknown even curvature without Hessian estimation (\Cref{sec:model}).
\item We introduce relative shell variation $\tau_a$ and prove a nearly optimal $\widetilde O(d+\sqrt{d/\eps})$ one-round whole-cell recovery bound whenever $\tau_a\le c\eps/\sqrt{d}$ (versus $\Theta(d/\eps)$ for signs), paired with a tight $\Theta(dR(\eps^{-1/R}-1))$ minimax endpoint--round barrier across $R$ feedback rounds under unbounded relative curvature (\Cref{thm:global,thm:barrier}).
\item We evaluate set information, point recovery, optimization progress, and solver cost separately across synthetic, ridge-regression, and calibrated non-even logistic benchmarks, certifying strict \emph{Full}-over-\emph{Reverse} set separation while confirming \emph{Reverse}'s consistent optimization gains over tuned \emph{Mirror} and unit-normalized \textsc{ZO-RankSGD} under joint caps (\Cref{sec:experiments,sec:discussion}).
\end{enumerate}}

%%%%%%%%%%%%%%%%%%%%%%%%%%%%%%%%%%%%%%%%%%%%%%%%%%%%%%%%%%
\section{Problem Setup and Ranking Information}
\label{sec:model}

\rev{This section formalizes how complete batch rankings extract directional constraints beyond isolated within-pair signs. We first illustrate shared-offset cancellation on a minimal two-pair example (\Cref{sec:four-endpoint-example}), define the local objective and ranking oracle (\Cref{sec:observation-model}), construct the nested \emph{Mirror}, \emph{Reverse}, and \emph{Full} consistency sets (\Cref{sec:three-sets}), and establish their directed-cycle hierarchy and recovery procedures (\Cref{sec:cycles-main}).}

\subsection{A Two-Pair Example in Two Dimensions}
\label{sec:four-endpoint-example}

For the linear-plus-even model, query $\pm x_1,\pm x_2$ with $x_1=(1,0)$ and $x_2=(0,1)$. Suppose their observed ranking is
\begin{equation}\label{eq:four-ranking}
f(-x_2)<f(-x_1)<f(x_1)<f(x_2).
\end{equation}
Within-pair differences cancel the even component, giving $f(x_i)-f(-x_i)=2g^\top x_i$. The two signs therefore constrain a candidate unit direction $v=(v_1,v_2)$ to $v_1,v_2\ge0$, but do not order its components. Both $(2,1)/\sqrt5$ and $(1,2)/\sqrt5$ satisfy these sign constraints.

The cross-pair orders $f(x_2)>f(x_1)$ and $f(-x_2)<f(-x_1)$ provide more information. With one unknown candidate offset $z_i$ shared by each opposite pair, they require
\begin{equation}\label{eq:four-cancellation}
\begin{aligned}
v_2+z_2&\ge v_1+z_1,\\
-v_2+z_2&\le-v_1+z_1
\end{aligned}
\quad\Longrightarrow\quad v_2-v_1\ge|z_2-z_1|\ge0.
\end{equation}
Thus the same four evaluations exclude $(2,1)/\sqrt5$ without knowing or equating the offsets. We now formalize this cancellation and the direction sets it induces.

\subsection{Objective and Ranking Feedback}
\label{sec:observation-model}

At a fixed recovery center $\theta_c\in\R^d$ of a loss $F:\R^d\to\R$, write $f(x)=F(\theta_c+x)$ and assume the exact-even decomposition
\begin{equation}\label{eq:objective}
f(x)=b+g^\top x+E(x),\qquad g\ne0,\qquad E(-x)=E(x),\qquad E(0)=0,
\end{equation}
where $b=F(\theta_c)$ and $E:\R^d\to\R$ is differentiable. Evenness implies $\nabla E(0)=0$, so $g=\nabla F(\theta_c)$ and the decoder recovers an estimate $\widehat u$ of $u=g/\norm g$ \rev{on the unit sphere $\Sph^{d-1}$} for a proposed update $\theta_{c+1}=\theta_c-\eta\widehat u$ \rev{with step size $\eta>0$}. Here $v\in\Sph^{d-1}$ is a candidate direction with radian angle $\ang(v,u)=\arccos(v^\top u)$.

\rev{The decomposition is exact for quadratics. For $f(x)=b+g^\top x+\frac12x^\top Hx$ expanded about $\theta_c$, $E(x)=x^\top Hx/2$ is even for any unknown anisotropic Hessian $H$.} Antipodal queries exploit this symmetry. Central differences, simultaneous perturbation stochastic approximation (SPSA), and two-point estimators evaluate $\pm x_i$ so even curvature cancels in $f(x_i)-f(-x_i)$ when numerical values are observed \citep{spall1992multivariate,nesterov2017random,duchi2015two}. Evolution strategies also use mirrored perturbations (\citealp{salimans2017evolution}). We study the information that survives this cancellation when only rankings are observed. The model also permits higher even-order terms; \Cref{sec:calibration} accounts for the odd remainder when exact evenness fails.

A batch contains $X=\{\sigma x_i:i=1,\ldots,n,\ \sigma\in\{-1,+1\}\}$, chosen before feedback. Common-shell queries \rev{at radius $a>0$} have $x_i=aW_i$ with $W_i\in\Sph^{d-1}$. The oracle returns a weak order $\pi$ without numerical objective values. The relation $(i,\sigma)\preceq_\pi(j,\zeta)$ means that $\sigma x_i$ ranks no higher than $\zeta x_j$; ties give both relations. A \emph{batch oracle} ranks the current endpoints, whereas a \emph{cumulative oracle} ranks all endpoints evaluated so far.

Each endpoint costs one function evaluation (FE), including antipodes, initialization, and repeated evaluations. A feedback round is one synchronization; sorting, fitting, and certification are separate computational costs. Minimax procedures may use measurable query and output rules with independent private randomness, subject to hard budgets of $N$ endpoints and $R$ rounds on every history. They must satisfy
\begin{equation}\label{eq:success}
\Prob\{\ang(\widehat u,u)\le\eps\}\ge3/4
\quad\text{for every fixed instance.}
\end{equation}
Upper constructions use batch rankings; lower bounds allow the stronger cumulative oracle.

\subsection{Ranking Constraints and Consistency Sets}
\label{sec:three-sets}

Subtracting $b$ and dividing by $\norm g>0$ preserve the ranking and give latent scores
\begin{equation}\label{eq:scores}
s_{i,\sigma}=\sigma u^\top x_i+e_i,\qquad e_i=E(x_i)/\norm g,\qquad \sigma\in\{-1,+1\}.
\end{equation}
Within a pair, the offset cancels and reveals $\sgn(u^\top x_i)$. Across pairs, $s_{j,+}>s_{i,+}$ and $s_{j,-}<s_{i,-}$ imply
\begin{equation}\label{eq:reverse-cancellation}
\begin{aligned}
v^\top(x_j-x_i)+(z_j-z_i)&\ge0,\\
v^\top(x_j-x_i)-(z_j-z_i)&\ge0
\end{aligned}
\quad\Longrightarrow\quad v^\top(x_j-x_i)\ge0.
\end{equation}
Adding the inequalities eliminates the offset difference; reorienting either pair gives the other signed relations. Throughout, strict observations impose weak candidate inequalities to obtain closed sets, and observed ties impose both inequalities.

\begin{definition}[Within-pair consistency (Mirror)]\label{def:mirror}
Mirror retains only within-pair orders, defining
\begin{equation}\label{eq:mirror-set}
\cC_{\rm mirror}(X,\pi)=\bigl\{v\in\Sph^{d-1}:\sigma v^\top x_i\le\zeta v^\top x_i
\text{ whenever }(i,\sigma)\preceq_\pi(i,\zeta)\bigr\}.
\end{equation}
\end{definition}

\begin{definition}[Cross-pair consistency (Reverse)]\label{def:reverse}
A reversed relation has $i\ne j$ and $\sigma,\zeta\in\{-1,+1\}$ with
$(i,\sigma)\preceq_\pi(j,\zeta)$ and $(j,-\zeta)\preceq_\pi(i,-\sigma)$. Adding these relations to Mirror defines
\begin{equation}\label{eq:reverse-set}
\begin{aligned}
\cC_{\rm rev}(X,\pi)=\bigl\{v\in\cC_{\rm mirror}(X,\pi):\ &\zeta v^\top x_j-\sigma v^\top x_i\ge0\\
&\text{for every reversed relation}\bigr\}.
\end{aligned}
\end{equation}
\end{definition}

\begin{definition}[Complete-ranking consistency (Full)]\label{def:full}
Full retains all ranking inequalities with one unrestricted offset per physical pair. For $(v,z)\in\R^d\times\R^n$, define
\begin{equation}\label{eq:joint}
\begin{aligned}
K(X,\pi)&=\bigl\{(v,z):\sigma v^\top x_i+z_i\le\zeta v^\top x_j+z_j
\text{ whenever }(i,\sigma)\preceq_\pi(j,\zeta)\bigr\},\\
\cC(X,\pi)&=\operatorname{proj}_v K(X,\pi)\cap\Sph^{d-1}.
\end{aligned}
\end{equation}
\end{definition}
The projection includes $v$ whenever some shared offsets explain the order. Adjacent-order inequalities suffice, with both directions for ties. Repeated physical pairs share one offset, and a common additive gauge can be fixed. On one shell, positive rescaling yields $\cC(W,\pi)$; across shells, we retain physical coordinates and the same gradient.

\subsection{Information Hierarchy and Recovery}
\label{sec:cycles-main}\label{sec:decoder}

\begin{proposition}[Validity and information hierarchy]\label{prop:validity}
For every differentiable even component and its observed ranking,
\begin{equation}\label{eq:inclusions}
u\in\cC(X,\pi)\subseteq\cC_{\rm rev}(X,\pi)\subseteq\cC_{\rm mirror}(X,\pi).
\end{equation}
For fixed $v$, Full's inequalities are difference constraints in $z$, feasible exactly when every directed cycle has nonnegative weight\rev{, where} Mirror retains self-loops, Reverse retains two-edge cycles, and Full enforces all directed cycles.
\end{proposition}
Writing each inequality in \Cref{eq:joint} as $z_i - z_j \le w_{ji}(v)$ defines a directed edge $j\to i$ of weight $w_{ji}(v)=\zeta v^\top x_j - \sigma v^\top x_i$ on pair vertices $\{1,\ldots,n\}$. Summing around any directed cycle cancels $z$ identically; a negative-weight cycle excludes $v$, while shortest-path potentials prove feasibility when none exists (\Cref{app:geometry}).

\rev{\paragraph{Information beyond Reverse (a three-pair cycle example).}\label{app:cycle-example}
Longer cycles can impose directional constraints beyond Reverse. Consider the three coordinate directions $e_1,e_2,e_3$ with $g=(1/5,4/5,1/5)$, $H=\operatorname{diag}(7/5,2,13/5)$, and $f(x)=g^\top x+x^\top Hx/2-1$. Their observed ascending ranking is
\[
\begin{aligned}
f(-e_2)=-4/5 &< f(-e_1)=-1/2 < f(e_1)=-1/10 \\
&< f(-e_3)=1/10 < f(e_3)=1/2 < f(e_2)=4/5.
\end{aligned}
\]
Mirror requires $v_1,v_2,v_3\ge 0$, and the two-edge Reverse relations require $v_2\ge\max\{v_1,v_3\}$. A candidate proportional to $(1/2,4/5,1/2)$ satisfies all Reverse constraints. However, the complete ranking nests the two disjoint score intervals of pairs 1 and 3 (widths $2v_1$ and $2v_3$) inside the interval of pair 2 (width $2v_2$), forcing Full to require $v_2\ge v_1+v_3$, which $(1/2,4/5,1/2)$ violates. For this candidate, the directed three-edge cycle $1\to 2\to 3\to 1$ has edge weights $3/10$, $3/10$, and $-1$, summing to $-2/5<0$ and certifying that no shared offsets can explain the ranking.}

\paragraph{Recovering a direction.}
The one-round design queries $N=2n$ endpoints $\theta_c\pm aW_i$ with independent uniform $W_i$. \rev{For strict spanning designs, the numerical decoder computes a bounded-section analytic center \citep{goffin2002analytic}. Let $B$ encode the $m$ row-normalized homogeneous ranking inequalities, with variables $y=v$ for Mirror and Reverse, or $y=(v,z)$ with one fixed gauge offset for Full. Given strictly positive weights $w$, setting $\ell=B^\top w$ fixes the scale on a bounded affine section and yields the convex log-barrier program
\begin{equation}\label{eq:decoder-center}
\widehat y=\argmin_{\ell^\top y=1,\ By>0}
\left[-\sum_{j=1}^m\log(By)_j\right],\qquad
\widehat u=\widehat y_g/\norm{\widehat y_g},
\end{equation}
where $\widehat y_g$ denotes the gradient component of $\widehat y$.} A feasibility check validates $\widehat u$, including shared offsets for Full, and optional residual-checked caps enclose Reverse and hence Full. These numerical procedures use no extra endpoints or rounds (solver and certification details appear in \Cref{app:analytic-center,app:enclosing-center,app:certificate}). The recovery guarantee applies to any measurable consistent selector (\Cref{app:consistent-selector}).

Mirror stores $O(n)$ constraints in $d$ variables; Reverse can require $O(n^2)$. Full has $2n-1$ adjacent rows but $d+n-1$ variables after gauge fixing, for distinct pairs and a strict ranking (ties require both directions). Its Newton step uses dense QR factorization (orthogonal and upper-triangular factors) at cost $O(n(d+n)^2)$, excluding iteration counts and bit complexity. The gradient certificate can store $O(n^2d)$ coefficients (\Cref{app:computation}).

\section{When Rankings Beat Signs}
\label{sec:localization}

\subsection{Recovery under Small Shell Variation}
\label{sec:ranking-gain}

\rev{Set inclusion alone does not guarantee angular accuracy, because a consistent set may remain wide even when it shrinks strictly inside Mirror. To determine when cross-pair rankings localize the gradient direction, we must measure how much the unknown even nuisance fluctuates across the query sphere relative to the linear gradient signal. We formalize this through the \emph{relative shell variation}, defined as}
\begin{equation}\label{eq:tau}
\Omega_a(E)=\sup_{\norm w=\norm z=1}|E(aw)-E(az)|,
\qquad \tau_a=\frac{\Omega_a(E)}{a\norm g}.
\end{equation}
Write $\tau=\tau_a$ on a fixed shell. \rev{For a quadratic $E(x)=x^\top Hx/2$, this simplifies to $\tau_a=a(\lambda_{\max}(H)-\lambda_{\min}(H))/(2\norm g)$. Notably, isotropic curvature ($H=\lambda I$) contributes zero shell variation regardless of how large $\norm{H}$ is, because constant shell shifts preserve every cross-pair ranking. Neither $\tau_a$ nor an upper bound on it is supplied to the decoder, and candidate offsets remain unrestricted.} Let $\cB(u,\eps)=\{v\in\Sph^{d-1}:\ang(v,u)\le\eps\}$.

\begin{theorem}[Whole-cell localization with unknown even nuisance]\label{thm:global}
There are absolute constants $c,C>0$ such that the following holds. Let $d\ge3$, $0<\eps\le1/10$, $0<\delta\le1/4$, and fix any even instance with $\tau\le c\eps/\sqrt d$. Set
\begin{equation}\label{eq:global-count}
L_\eps=d\log(C\sqrt d/\eps)+\log(C/\delta),\qquad
n\ge C\bigl(L_\eps+\sqrt{L_\eps/\eps}\bigr).
\end{equation}
The complete ranking of $2n$ paid endpoints $\pm aW_i$, with independent uniform $W_i\in\Sph^{d-1}$, satisfies
$\Prob\{u\in\cC\subseteq\cC_{\rm rev}\subseteq\cB(u,\eps)\}\ge1-\delta$.
\end{theorem}

\rev{\paragraph{Why $\tau_a\le c\eps/\sqrt d$ unlocks the speedup.}
\Cref{thm:global} delivers a whole-cell guarantee under which every Reverse-consistent estimate is $\eps$-accurate, and the entire feasible set has diameter at most $2\eps$. With $n$ pairs and $N=2n$ endpoints, the fixed-confidence cost is $\widetilde O(d+\sqrt{d/\eps})$. In contrast, one-round sign-only point recovery has minimax cost $\Theta(d/\eps)$ on the same class, even on its zero-variation isotropic subclass (\Cref{cor:mirror-lower,app:mirror}). In high-precision regimes ($\eps\ll 1$), cross-pair reversals break the linear $1/\eps$ precision barrier down to at most a square-root $\sqrt{1/\eps}$ dependence; for illustration, at $d=10^4$ and $\eps=10^{-4}$, the leading scaling drops from $d/\eps=10^8$ to $d+\sqrt{d/\eps}=2\times 10^4$ up to polylogarithmic factors and absolute constants. The sufficient condition $\tau_a\le c\eps/\sqrt d$ directly reflects high-dimensional spherical geometry, because random directions on $\Sph^{d-1}$ capture a gradient projection of typical order $1/\sqrt d$, so preserving reverse-order margins at angular resolution $\eps$ requires shell variation below scale $\eps/\sqrt d$. (This comparison concerns endpoint complexity excluding decoding cost, and Mirror's rate is a point-recovery bound rather than a whole-cell bound for independent and identically distributed (iid) queries.)}

The proof finds pairs with oppositely signed projection gaps for $u$ and a distant candidate, with margins proportional to $\eps/\sqrt d$. Shell variation below these margins preserves the reverse orders that exclude the candidate. Endpoint-disjoint witnesses control dependencies among comparisons. A failure exponent of order $\min\{n,n^2\eps\}$, combined with a margin-net argument, yields the square-root term without treating all comparisons as independent (\Cref{app:one-round-proof}).

\paragraph{Near-optimality and larger variation.}
Even the known isotropic family $q_u(x)=u^\top x+\norm x^2/2$ requires $\Omega(d+\sqrt{d/\eps})$ one-round endpoints, allowing arbitrary radii (\Cref{app:benign-lower}). The bounds therefore match up to logarithmic factors. For larger variation, \Cref{app:degradation} gives a sufficient enclosing radius increasing with $\sqrt d\,\tau_a$. This is not an accuracy floor. Exact Mirror signs still permit finer recovery with more evaluations.

\paragraph{Query savings versus constraint count.}
Reverse's $\widetilde O(d+\sqrt{d/\eps})$ pairs can generate up to $O(n^2)$ cross-pair constraints, compared with $O(n)$ sign rows for Mirror's $\Theta(d/\eps)$ pairs. Transitive reduction of Reverse's directed acyclic graph removes redundant constraints (\Cref{app:computation}). Runtime therefore depends on oracle cost, feedback latency, and fitting overhead. Rank-weighted estimates require $O(nd)$ arithmetic once ranks are available (\Cref{sec:exp-tasks}).

\subsection{Limits under Unbounded Relative Curvature}
\label{sec:barrier}

\rev{A natural question is whether standard conditioning assumptions alone guarantee this ranking advantage. We show that a small condition number does not ensure small relative shell variation and cannot replace control of relative curvature scale.} For any prescribed $\kappa>1$, consider
\begin{equation}\label{eq:quadratic}
\cQ=\{b+u^\top x+\tfrac12x^\top Hx:\norm u=1,\ H\succ0,\ \cond(H)<\kappa\},
\end{equation}
with no common bound on $\norm H$. Let $N_\star(d,\eps,R)$ be the least hard endpoint budget satisfying \Cref{eq:success} uniformly over $\cQ$ with at most $R$ rounds.

\begin{theorem}[Cumulative-ranking scale barrier]\label{thm:barrier}
For $d\ge3$, $0<\eps\le1/10$, and integer $R\ge1$, under either ranking oracle,
\begin{equation}\label{eq:barrier}
N_\star(d,\eps,R)=\Theta\!\bigl(dR(\eps^{-1/R}-1)\bigr),
\end{equation}
with absolute constants uniform over every prescribed $\kappa>1$. The lower bound allows arbitrary radii, adaptive batch sizes, early stopping, and randomized queries.
\end{theorem}

The lower bound constructs well-conditioned quadratics at a sufficiently large curvature scale. With high probability over the independent instance construction and private randomness, curvature hides all non-antipodal directional information in the finite query tree, leaving only sign information. A finite Mirror construction matches the bound (\Cref{app:scale-proof}). Thus complete rankings offer no minimax endpoint advantage on this class. One round costs $\Theta(d/\eps)$, while sufficiently many rounds reduce the cost to $\Theta(d\log(1/\eps))$ (\Cref{tab:theory}).

For each fixed differentiable even instance, $\tau_a\to0$ as $a\to0$. This convergence is not uniform over $\cQ$, so small radii help fixed instances without breaking the uniform barrier (\Cref{app:small-radius}). \rev{Together, \Cref{thm:global,thm:barrier} establish why relative shell variation, rather than total curvature magnitude or condition number, governs when batch rankings beat signs.}

\begin{table}[!t]
\centering\footnotesize
\caption{Recovery bounds at fixed confidence. Reverse and Mirror concern the same small-variation class; unbounded quadratic scale defines a different class. Only the pairwise benchmark counts comparisons; all other rows count endpoints. Constraint processing is excluded from every row; constraint counts and measured solve times appear in \Cref{sec:decoder,app:computation}.}\label{tab:theory}
\begin{tabularx}{\linewidth}{@{}Xll@{}}
\toprule
Instance / information used & Rounds & Reported bound\\
\midrule
Even nuisance, $\tau\le c\eps/\sqrt d$; Reverse & $1$ & $O(L_\eps+\sqrt{L_\eps/\eps})$\\
Known isotropic quadratic; complete ranking & $1$ & $\Omega(d+\sqrt{d/\eps})$\\
Even nuisance, $\tau\le c\eps/\sqrt d$; Mirror & $1$ & $\Theta(d/\eps)$\\
Supplied-smoothness pairwise recovery \citep{tao2026comparison} & adaptive & $\Theta(d\log(1/\eps))$\\
Unbounded quadratic scale; complete ranking & $R$ & $\Theta(dR(\eps^{-1/R}-1))$\\
\bottomrule
\end{tabularx}
\end{table}

\subsection{Recovery beyond Exact Evenness}
\label{sec:calibration}\label{sec:robust}

For a general smooth objective, the local remainder need not be even\rev{, because} higher-order odd terms $O(x)$ can break antipodal symmetry, while endpoint noise $\xi_{i,\pm}$ can introduce additional asymmetry. As formalized next, an odd cubic perturbation on a single radius-$a$ shell mimics tilting the linear gradient itself, so an uncalibrated exact-even fit can shrink onto the wrong cone and exclude the truth $u$.

\rev{\begin{proposition}[Single-shell ambiguity under an odd remainder]\label{prop:odd-shell}
Let $u,q$ be orthogonal unit vectors, $a>0$, $\mu\ne0$, and $E$ a differentiable even function. The objectives
\begin{equation}\label{eq:odd-equivalence}
f_\mu(x)=u^\top x+E(x)+\mu(q^\top x)\norm x^2,
\qquad
\widetilde f_a(x)=(u+\mu a^2q)^\top x+E(x)
\end{equation}
coincide on $\norm x=a$, but their gradient directions at zero differ by $\Delta=\arctan(|\mu|a^2)$. Any randomized ordinal procedure restricted to this sphere has angular error at most $\eps<\Delta/2$ with probability at most $1/2$ on at least one of these two instances.
\end{proposition}}

To preserve validity beyond exact evenness, we extend the model with a calibrated odd-error allowance. Write $f(x)=b+g^\top x+E(x)+O(x)$ with $O(-x)=-O(x)$ and $\nabla O(0)=0$, and assume bounded endpoint errors $|\xi_{i,\pm}|\le\nu$. Split noise into its even part $\bar\xi_i=(\xi_{i,+}+\xi_{i,-})/(2a\norm g)$ and odd part $\widetilde\xi_i=(\xi_{i,+}-\xi_{i,-})/(2a\norm g)$, and define the shell-normalized even offset $\tilde e_i=E(aW_i)/(a\norm g)+\bar\xi_i$, distinct from \Cref{eq:scores}. The normalized scores are $\tilde e_i\pm(u^\top W_i+o_i)$, where $o_i=O(aW_i)/(a\norm g)+\widetilde\xi_i$. The robust set $\cC_\rho$ retains free even offsets $\tilde e'_i$ and assigns each antipodal pair one shared error variable $b_i\in[-\rho,\rho]$ with opposite signs at $\pm x_i$, where the envelope $\rho$ is supplied externally from prior regularity bounds. Candidate scores $\tilde e'_i\pm(v^\top W_i+b_i)$ retain $u$ whenever $\rho\ge\sup_{\norm w=1}|O(aw)|/(a\norm g)+\nu/(a\norm g)$, while $\bar\xi_i$ inflates effective even shell variation by at most $2\nu/(a\norm g)$.

\rev{\begin{corollary}[Bounded shared-error localization]\label{cor:robust}
Under the uniform query design of \Cref{thm:global} with observed normalized scores $\tilde e_i\pm(u^\top W_i+o_i)$ satisfying $|o_i|\le\rho$ and actual even-offset range at most $\tilde\tau_a$, the true direction belongs to $\cC_\rho$. Moreover, the endpoint count of \Cref{thm:global} guarantees $u\in\cC_\rho\subseteq\cB(u,\eps)$ with probability at least $1-\delta$ whenever $\tilde\tau_a+\rho\le c\eps/\sqrt d$ for a sufficiently small universal $c$ (in particular, whenever $\tau_a+2\nu/(a\norm g)+\rho\le c\eps/\sqrt d$).
\end{corollary}}

Proofs of \Cref{prop:odd-shell,cor:robust} and an additional smoothness-only corollary appear in \Cref{app:bounded-error}.

Calibration follows directly from smoothness bounds\rev{. Given} $\sup_{\norm x\le a}\norm{D^3 f(x)}_{\rm op}\le M_3$ and $\norm g\ge g_{\min}>0$, Taylor's remainder supplies
\begin{equation}\label{eq:odd-noise-tradeoff}
\rho_a^{\rm fit} = \frac{M_3 a^2}{6g_{\min}} + \frac{\nu}{ag_{\min}},
\end{equation}
and an $L$-Lipschitz gradient suffices without $M_3$ under noiseless rankings when $a\le c_0\eps g_{\min}/(L\sqrt d)$ with $\rho=La/(2g_{\min})$ (\Cref{cor:smooth-fixed}). \Cref{eq:odd-noise-tradeoff} explains the U-shaped cap behavior on non-even logistic tasks (\Cref{sec:exp-validity})\rev{, where} shrinking $a$ suppresses cubic odd bias ($O(a^2)$) yet amplifies normalized noise ($O(\nu/a)$). The model assumes transitive shared-pair errors. For equal axis allowances, the initializer's public positivity test requires $\rho<1/\sqrt{d(d+1)}$; a usable chart must also pass its checked enclosure criterion (\Cref{app:robust-chart}).

\section{Experiments from Information to Task Progress}\label{sec:experiments}
\subsection{Datasets and Experimental Setup}\label{sec:exp-setup}
Synthetic objectives vary nuisance scale and rank. Datasets from the University of California, Irvine (UCI) Machine Learning Repository support moving-center ridge optimization (Airfoil and Concrete) and calibrated logistic recovery outside exact evenness (Blood and Banknote). The studies impose ranking interfaces on these ordinary objectives. Direction-set certificates, mean squared angular error (MSE), and relative optimization gaps measure different outcomes; smaller angular errors and optimization gaps are better. Exact-even online recovery uses coordinates and ranks. Numerical gradients, optima, and held-out losses support auditing, while separate offline development runs tune procedures.

Each real-data task uses a fixed 80/20 split and training-only standardization. Ratios compare arithmetic means; resampling preserves repeated settings within each independent target or algorithm seed. Confidence intervals (CIs) quantify neither split nor dataset variation. Exact budgets and calibration are stated with each study. Computation measurements use the shared central processing unit (CPU) and graphics processing unit (GPU) setup in \Cref{app:computation}. Mirror, Reverse, and Full are matched information-layer controls. Ridge-optimization controls are ranking-based zeroth-order stochastic gradient descent (ZO-RankSGD) \citep{tang2024ranking}, stochastic three points (STP, \citealp{bergou2020stp}), covariance matrix adaptation evolution strategy (CMA-ES, \citealp{hansen2001cma}), and incumbent random search (Random). These optimizers use native queries, so those comparisons concern complete procedures rather than an isolated information change (\crefrange{app:structural-study}{app:computation}). Shared query draws yield identical observations only at a common center.

\subsection{When do rankings reveal more than signs?}\label{sec:exp-information}
A checked radius-$U$ cap encloses Reverse and hence Full; two Mirror-feasible directions separated by $2L$ force every enclosing Mirror cap to have radius at least $L$. Thus $U<L$ certifies information beyond signs independently of the selected estimate. Full's exact projection is audited separately in three dimensions (\Cref{app:three-cones}).

\begin{figure}[t]
\centering\includegraphics[width=0.92\linewidth]{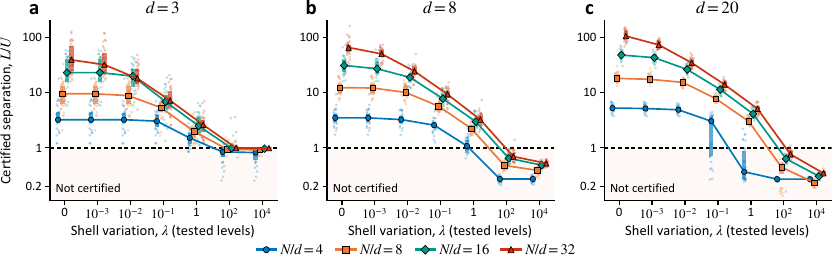}
\caption{When rankings certify more information than signs. Targets use $H=100\lambda(I+0.02zz^\top)$ with $\norm z=1$ (condition number $1.02$ at positive scales, $H=I$ at $\lambda=0$), so $\lambda$ controls shell variation. Small $\lambda$ corresponds to the favorable regime of \Cref{thm:global}; the declining ratio at large $\lambda$ indicates reduced ability to certify separation on finite designs. \Cref{thm:barrier} establishes the minimax barrier separately. Panels use $d=3,8,20$; colors distinguish endpoint budgets $N/d$. The ratio is $L/U$: $L$ lower-bounds the any-center enclosing radius of the same-batch Mirror set $\cC_{\rm mirror}$, and $U$ is a checked outer-cap radius for Reverse and hence Full. Values above the dashed line certify strictly more directional information than Mirror from the same endpoints; values at or below one do not certify separation. Curves and bands show medians and interquartile ranges over 32 shared target--axis units per dimension. Horizontal levels are tested categories rather than equally spaced increments.}\label{fig:scale}
\end{figure}

Uniform antipodal batches use $N/d\in\{4,8,16,32\}$ and $H=100\lambda(I+0.02zz^\top)$, $\norm z=1$. Positive scales have condition number $1.02$ and nominal unit-shell variation $\lambda$; at zero, $H=I$. The design reuses 96 independent target--axis units, 32 at each $d=3,8,20$. Separation $U<L$ holds in all 384 cases at each tested level $\lambda\le10^{-2}$, then drops to 375, 341, 31, and zero for $\lambda=10^{-1},1,10^2,10^4$ (\Cref{fig:scale}).

True-direction feasibility and cap coverage hold throughout. At $\lambda=10^4$, all 64 exact checks at $d=3$, $N/d=8,32$ establish Full--Mirror equality. Information gain can therefore disappear while the model remains valid. All 576 quartic even conditions retain the true direction and certify separation (\Cref{app:scope-results}); \Cref{app:first-batch-audit} gives an independent audit. These tests illustrate the mechanism without estimating asymptotic rates or calibrating the theorem's threshold.

\subsection{Beyond short cycles under higher-rank anisotropy}\label{sec:exp-mechanism}\label{sec:new-high-rank}
\begin{figure}[t]
\centering\includegraphics[width=\linewidth]{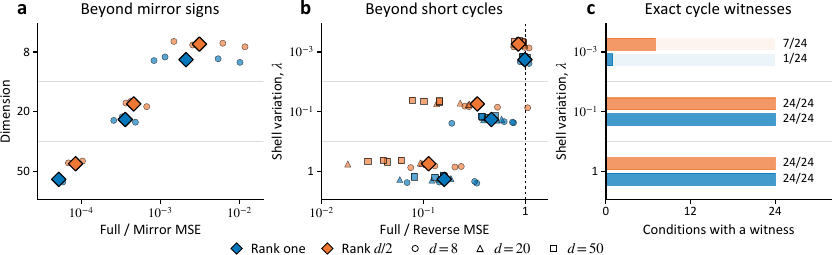}
\caption{\rev{Extra information under matched anisotropic ranks. Colors distinguish rank one and rank $d/2$. (a) Full/Mirror and (b) Full/Reverse squared angular-error ratios below one favor Full. Diamonds show pooled ratios; small points show within-unit ratios, with dimension encoded by shape in (b). Four independent target--reflection--design units per dimension are reused across budgets and scales. (c) Bars count exact negative-cycle witnesses out of 24 conditions per scale and rank. A witness proves strict Full--Reverse inclusion, not an MSE gain; its absence is inconclusive. MSE contrasts are exploratory (\Cref{app:high-rank-statistics}).}}\label{fig:new-high-rank}
\end{figure}

Full's extra constraints also appear beyond rank-one nuisance. By matching the spectral range and condition number, we test whether Full's additional constraints persist when anisotropy is spread across more directions. Across twelve target--reflection--design units (four per dimension, two ranks, two budgets, three scales), all 144 conditions certify Full--Mirror separation. Exact negative cycles establish strict Full--Reverse inclusion in 49/72 rank-one and 55/72 high-rank cases (\Cref{fig:new-high-rank}; full protocol and statistics appear in \Cref{app:high-rank}). Exploratory MSE contrasts remain descriptive.

\subsection{Which information layer helps optimization?}\label{sec:new-tasks}\label{sec:exp-tasks}
UCI Airfoil Self-Noise and Concrete Compressive Strength ridge tasks use $d=6,9$ parameters including the intercept, feature penalty $0.01$, and no Hessian whitening \citep{uci_airfoil,uci_concrete}. Squared loss is exactly linear-plus-even at every center, so we rebuild the recovery model after each move and measure progress by the relative optimization gap $r(\theta)=[F(\theta)-F^\star]/[F(\theta_0)-F^\star]$ using $F^\star$ solely offline.

\paragraph{Matched inference layers.}
Mirror, Reverse, and Full share query draws and the unit-direction step grid $\{0,2^{-10},\ldots,2^1\}$. The step candidates, including the current center, are ranked in a second batch; the best-ranked candidate becomes the next center. Eight updates ($8d$ recovery plus 13 step points each) spend $488/680$ endpoints in 16 rounds on Airfoil/Concrete across two shared radii ($1/8,1/64$) and 32 paired seeds (384 trajectories). Full improves the mean gap over Mirror on Concrete; the Airfoil comparisons are inconclusive, and no Full/Reverse interval excludes one (\rev{\Cref{tab:task-attribution}, Panel~A}).

All external controls use ordinal incumbent updates and disjoint development seeds (32 evaluation seeds per split; \Cref{app:task-baselines}).

\begin{figure}[t]
\centering\includegraphics[width=0.92\linewidth]{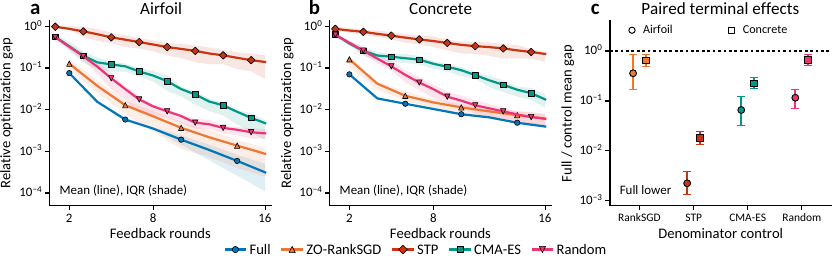}
\caption{Ridge optimization under joint endpoint and sixteen-round caps ($489/681$ endpoints on Airfoil/Concrete). Curves begin at round 2 after round-1 recovery ($8d$ endpoints) and round-2 line search ($13$ points); STP spends only $33$ endpoints within $16$ rounds. (a, b) Mean relative optimization gaps over 32 seeds (shading shows the interquartile range). (c) Full/control ratios with 16-contrast Bonferroni 95\% CIs. Full is the representative ranking pipeline, selecting the same tuned radii ($1/64, 1/16$) as Reverse with no statistically established Full-over-Reverse improvement (\rev{\Cref{tab:task-attribution}, Panel~B}); \rev{Panel~C of \Cref{tab:task-attribution}} reports Reverse's follow-up ratios against Mirror ($0.175/0.417$) and unit-normalized ZO-RankSGD ($0.378/0.667$).}
\label{fig:new-external}
\end{figure}

\paragraph{Do longer cycles explain these gains?}
Independently tuning Reverse with Full's setup selects identical radii ($1/64$ Airfoil, $1/16$ Concrete), matching all 128 paired first batches. Reverse retains adjusted gains over both batch-capable controls, with no statistically established Full-over-Reverse improvement on gap or test root mean squared error (RMSE) (\rev{\Cref{tab:task-attribution}, Panel~B}; \Cref{app:task-inference}). Under joint caps ($489/681$ FEs, 16 rounds; \Cref{fig:new-external}), Full and ZO-RankSGD exhaust budgets whereas sequential STP uses 33.

\paragraph{Two 128-seed control cohorts (\rev{\Cref{tab:task-attribution}, Panel~C}).}
\rev{Two separate 128-seed cohorts isolate the effects of independent tuning and gradient normalization (\Cref{app:task-inference}). Cohort~1 (CPU) applies a common momentum update ($p_t=\operatorname{normalize}(h_t+\beta p_{t-1})$, 24-config budget) across independently tuned Mirror, Reverse, and Full. Cohort~2 (GPU, no momentum, 12 configs) compares against unit-normalized ZO-RankSGD. In contrast to both sign-only and rank-weighted baselines, Reverse and Full consistently attain substantially lower terminal optimization gaps across both cohorts (Reverse ratios $0.175/0.417$ against tuned Mirror and $0.378/0.667$ against unit-normalized ZO-RankSGD), with no statistically established additional improvement from Full over Reverse.}

\begin{table}[t]
\centering\footnotesize
\caption{Which information layer improves ridge optimization? Ratios of mean relative optimization gaps ($<1$ favors numerator). Panel~A: matched draws/radii/steps (32 seeds). Panel~B: independent tuning under joint caps (32 seeds). Panel~C: two unpooled 128-seed follow-up cohorts (Cohort~1: momentum, CPU; Cohort~2: unit-normalized ZO-RankSGD, no momentum, GPU; \Cref{app:task-inference}). Brackets show study-specific Bonferroni 95\% CIs; intervals spanning $1$ do not establish equivalence.}\label{tab:task-attribution}
\begin{tabular}{@{}lcc@{}}
\toprule
Comparison & Airfoil & Concrete \\
\midrule
\multicolumn{3}{@{}l}{\emph{A. Matched inference layers; sixteen-round trajectories}}\\ \hline \\
Full / Mirror, $a=1/8$ & $0.562\ [0.221,1.482]$ & $0.742\ [0.625,0.884]$ \\
Full / Mirror, $a=1/64$ & $0.437\ [0.175,1.237]$ & $0.734\ [0.621,0.864]$ \\
Full / Reverse, $a=1/8$ & $0.892\ [0.592,1.331]$ & $1.027\ [0.952,1.109]$ \\
Full / Reverse, $a=1/64$ & $0.993\ [0.720,1.371]$ & $1.000\ [0.923,1.084]$ \\ 
\midrule
\multicolumn{3}{@{}l}{\emph{B. Independently tuned methods; joint endpoint--round caps}}\\  \hline \\
Full / Reverse & $0.956\ [0.767,1.185]$ & $1.003\ [0.929,1.081]$ \\
Reverse / ZO-RankSGD & $0.374\ [0.179,0.823]$ & $0.645\ [0.512,0.823]$ \\
Reverse / CMA-ES & $0.0689\ [0.0354,0.1250]$ & $0.223\ [0.172,0.292]$ \\ 
\midrule
\multicolumn{3}{@{}l}{\emph{C. Unpooled 128-seed follow-up controls (joint caps)}}\\  \hline \\
Reverse / Mirror (momentum, Cohort~1) & $0.175\ [0.125,0.249]$ & $0.417\ [0.380,0.456]$ \\
Full / Mirror (momentum, Cohort~1) & $0.180\ [0.126,0.262]$ & $0.409\ [0.373,0.448]$ \\
Reverse / Unit-norm.\ ZO-RankSGD (Cohort~2) & $0.378\ [0.264,0.533]$ & $0.667\ [0.593,0.752]$ \\
Full / Unit-norm.\ ZO-RankSGD (Cohort~2) & $0.374\ [0.259,0.539]$ & $0.674\ [0.601,0.757]$ \\
\bottomrule
\end{tabular}
\end{table}

\begin{samepage}
\paragraph{Separate 32-seed A100 timing study.}
Separately from both 128-seed cohorts, a 32-seed warmed A100 study without momentum measures time to relative optimization gap $0.03$ (\rev{\Cref{tab:ridge-time-to-target}}). Reverse and native ZO-RankSGD succeed on all $32/32$ seeds within 16 rounds. Reverse uses fewer rounds ($4.000$ vs.\ $5.125/5.688$), with higher active time ($46.892/62.103\text{ ms}$ vs.\ $3.294/3.556\text{ ms}$). Under the post-hoc model $T=T_{\rm active}+R\ell$, their mean costs cross at hypothetical round delays of $38.75/34.69\text{ ms}$ on Airfoil/Concrete. Timer scope, stricter-target success counts, and optional certification costs appear in \Cref{app:computation}.\par
\end{samepage}

\begin{table}[tbp]
\centering\small
\caption{\rev{Separate 32-seed warmed A100 study of time to relative optimization gap $0.03$ (no momentum, unpooled from the 128-seed follow-up cohorts in \Cref{tab:ridge-followup-controls}; all $32/32$ seeds succeed within 16 rounds). Active time (ms) covers online query generation, evaluation, sorting, fitting, and line search; offline tuning, audits, and certification are excluded. Conditional crossover $\ell_\star$ solves the post-hoc additive projection $T_{\rm active}+R\ell$ against native ZO-RankSGD at this $0.03$ target.}}\label{tab:ridge-time-to-target}
\begin{tabular}{@{}lrrrr@{}}
\toprule
Method & Airfoil rds & Airfoil ms & Concrete rds & Concrete ms \\
\midrule
Reverse & $4.000$ & $46.892$ & $4.000$ & $62.103$ \\
Native ZO-RankSGD & $5.125$ & $3.294$ & $5.688$ & $3.556$ \\
Unit-normalized ZO-RankSGD & $5.500$ & $3.581$ & $5.719$ & $3.710$ \\
\midrule
Latency crossover $\ell_\star$ vs.\ native & \multicolumn{2}{c}{$38.75\text{ ms/rd}$} & \multicolumn{2}{c}{$34.69\text{ ms/rd}$} \\
\bottomrule
\end{tabular}
\end{table}

\paragraph{Removing the round cap.}\label{sec:feedback-boundary}\label{sec:exp-resources}
With endpoint-only caps, STP and CMA-ES attain lower mean gaps than Full on both tasks, using 54--340 rounds (\Cref{tab:round-tradeoff,app:task-resources}). This descriptive reversal places the observed advantage in feedback-scarce settings. Full and Reverse run more slowly than ZO-RankSGD on these inexpensive objectives; lower optimization gaps bring no comparable improvement in test RMSE (\Cref{app:task-inference}).

\begin{table}[t]
\centering\footnotesize
\caption{Optimization without a round cap. Relative optimization gaps are arithmetic means over 32 seeds; parentheses give feedback rounds. Endpoint counts are listed as Airfoil/Concrete (A/C). Bold marks only the smallest observed mean, not statistical superiority (\Cref{tab:task-resource-ledger,app:task-inference}).}\label{tab:round-tradeoff}
\begin{tabular}{@{}lccc@{}}
\toprule
Method & Endpoints (A/C) & Airfoil: gap (rounds) & Concrete: gap (rounds) \\
\midrule
Full & $489/681$ & $3.13\times10^{-4}$ (16) & $3.95\times10^{-3}$ (16) \\
ZO-RankSGD & $489/681$ & $8.75\times10^{-4}$ (16) & $6.11\times10^{-3}$ (16) \\
STP & $489/681$ & $\mathbf{1.43\times10^{-5}}$ (244) & $6.60\times10^{-4}$ (340) \\
CMA-ES & $487/681$ & $1.58\times10^{-5}$ (54) & $\mathbf{1.85\times10^{-4}}$ (68) \\
Random & $489/681$ & $2.57\times10^{-3}$ (17) & $8.02\times10^{-3}$ (17) \\
\bottomrule
\end{tabular}
\end{table}

\subsection{Does calibrated recovery retain the true gradient?}\label{sec:exp-validity}\label{sec:new-noneven}
\begin{figure}[t]
\centering\includegraphics[width=\linewidth]{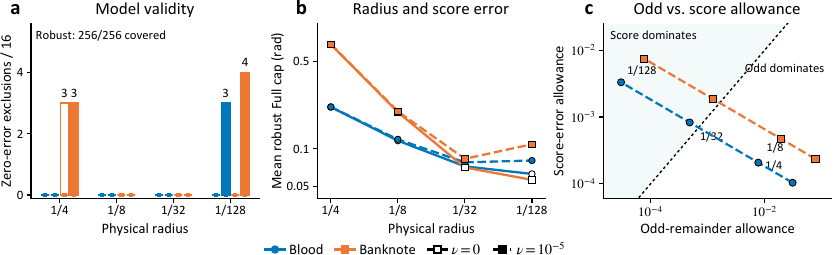}
\caption{\rev{Validity and uncertainty beyond exact evenness on Blood and Banknote. Colors identify tasks; line styles distinguish score-noise levels. (a) Counts show true-gradient exclusions by zero-error Full, out of sixteen seeds per task--radius--noise cell. Robust fits retain and cover the truth in all 256 conditions. (b) Mean robust Full outer-cap radii in radians show that smaller physical radii need not give tighter caps when score noise is present. (c) At $\nu=10^{-5}$, normalized score-error and odd-remainder allowances compete; the diagonal marks equality. All panels evaluate calibrated recovery at fixed centers (\Cref{app:logistic-arms}).}}
\label{fig:new-noneven}
\end{figure}

The ridge tasks satisfy exact evenness at every center. Logistic loss is also linear-plus-even at the zero parameter, so we use fixed centers with nonzero intercepts to test calibrated recovery with an odd remainder. On Blood and Banknote, 32 dataset--seed units across four radii and two noise levels give 256 conditions. \rev{Whereas zero-error Full excludes the true direction in 13 of 256 conditions (including three without score noise), matched robust fits retain and cover the true direction in all 256 conditions (\Cref{fig:new-noneven}). Under score noise ($\nu=10^{-5}$), the outer-cap radius exhibits a clear U-shaped dependence on the query radius $a$, shrinking from $a=1/4$ to $a=1/32$ as cubic odd bias decreases before rising at $a=1/128$ as normalized noise amplifies.} The complete protocol and local-step controls appear in \Cref{app:logistic-tasks}; the initializer's certified domain is narrower than the theorem's.

\subsection{\rev{Computational cost and higher-dimensional scaling}}\label{sec:exp-comp}
\rev{To separate online direction fitting from optional offline certification at higher dimensions, \Cref{tab:computation} reports median one-round wall-clock runtimes and peak GPU memory across $d\in\{20,50,100\}$ and $N/d\in\{8,16\}$ at $\lambda=10^{-3}$ (four independent targets per row; hardware and protocol details appear in \Cref{app:computation}). Analytic-center direction fitting takes $0.54\text{--}1.64\text{ s}$ across all tested settings (reaching $1.64\text{ s}$ at $d=100$ and $N=1600$ endpoints). In contrast, computing the optional residual-checked outer cap ($6.47\text{--}142.55\text{ s}$) exceeds fitting time in every row and is omitted in fit-only optimization modes, even after transitive reduction compresses roughly 640,000 Mirror and Reverse certificate rows down to 836--849 at $d=100$ and $N=1600$. All 24 fits are exactly feasible and all outer caps cover the true gradient direction.}

\begin{table}[t]
\centering\small
\caption{\rev{One-round computation and recovery up to $d=100$. Rows use four independent targets; times are medians in seconds. GPU denotes peak allocation in MiB. All 24 fits are exactly feasible; all caps cover the true direction.}}\label{tab:computation}
\begin{tabular*}{\linewidth}{@{\extracolsep{\fill}}rrrrrl@{}}
\toprule
$(d,N/d)$ & Fit & Certify & Recheck & GPU & Cap range (rad)\\
\midrule
$(20,8)$ & $0.55$ & $6.57$ & $0.33$ & $17.4$ & $[0.0380,0.0408]$\\
$(20,16)$ & $0.54$ & $6.47$ & $0.50$ & $21.2$ & $[0.00664,0.00916]$\\
$(50,8)$ & $0.70$ & $25.86$ & $4.14$ & $50.9$ & $[0.0214,0.0261]$\\
$(50,16)$ & $0.80$ & $27.86$ & $6.88$ & $177.2$ & $[0.00521,0.00584]$\\
$(100,8)$ & $1.16$ & $119.43$ & $29.04$ & $329.7$ & $[0.0162,0.0179]$\\
$(100,16)$ & $1.64$ & $142.55$ & $51.36$ & $1287.7$ & $[0.00391,0.00429]$\\
\bottomrule
\end{tabular*}
\end{table}

\section{Related Work}\label{sec:related}
\paragraph{Ordinal direction recovery and comparison optimization.}
Reconstructing unit vectors from ordered frame measurements was studied via linear and quadratic programming by \citet{nguyen2011fpq} under a purely linear model ($E\equiv 0$), whereas unknown pair-shared even offsets alter the feasible cone geometry. Using binary signs, \citet{cai2022onebit} recover gradient directions via one-bit estimation, while \citet{tao2026comparison} attain the adaptive $\Theta(d\log(1/\eps))$ comparison rate by shrinking query radii across sequential rounds using supplied smoothness and gradient-norm scales. Without supplied scales, adaptive radius selection incurs instance-dependent bounds \citep{scheinberg2026functionfree}, and normalized-Hessian methods target second-order stationarity \citep{wang2026stationary}. Iterative optimization methods include multiway and dueling convex methods \citep{saha2024multiway,saha2025general}, accelerated sign updates \citep{lobanov2024acceleration}, and rank-weighted \textsc{ZO-RankSGD} \citep{tang2024ranking}. Other settings include Riemannian dueling \citep{ren2026riemannian}, surrogate preference search \citep{xu2024preferential,zhang2025pabbo}, and policy or ordinal reward learning \citep{zhang2025zpg,afsharrad2026ordinal}. \rev{Variance-aware and corruption-robust dueling approaches accommodate stochastic or corrupted utilities \citep{di2024variance,cheng2025corruption}, using observation models distinct from our transitive shared-score perturbations. In contrast, our work characterizes all directional constraints implied by pair-shared even-curvature cancellation.}

\paragraph{Numerical antithetic zeroth-order estimators.}
Numerical zeroth-order methods include central-difference SPSA \citep{spall1992multivariate}, derivative-free quadratic models \citep{conn2009introduction}, and randomized two-point estimators \citep{nesterov2017random,duchi2015two,shamir2017optimal}. Other approaches use antithetic evolution strategies \citep{salimans2017evolution} or structured sparse, aligned, low-rank, and Hessian-informed variants \citep{chen2024deepzero,ma2025aligned,zhao2025hizoo,li2026hiso,chen2025lozo}\rev{, alongside analyses of generalization stability and privacy \citep{liu2024stability,zhang2024private}}. For estimators that use numerical symmetric differences $f(x)-f(-x)$, even curvature $E(-x)=E(x)$ cancels directly. \rev{In contrast, our exact-even decoder relies solely on rankings without numerical differences, while our non-even extension additionally uses supplied numerical calibration bounds.}

\section{Discussion and Limitations}
\label{sec:discussion}
Strict set inclusion need not reduce an enclosing radius or improve the selected direction estimate. Full's ridge-optimization gain over Reverse remains unestablished. Reverse supplies the conditional whole-cell guarantee and empirical gains over Mirror and unit-normalized ZO-RankSGD under joint caps.

Because Reverse generates up to $O(n^2)$ constraints versus $O(n)$ for Mirror, runtime depends on feedback latency relative to decoding cost. Direct Mirror-versus-Reverse scaling across $d$ remains unestablished. A post-hoc projection uses $T=T_{\rm active}+R\ell$ at the $0.03$ no-momentum ridge setting, excluding tuning and optional $\sim 1\text{ s}$ certification. It places Reverse's break-even point against native ZO-RankSGD near $\ell\approx 35\text{--}39\text{ ms}$. This is a conditional projection, not a measured delay or a general guarantee.

\subsection*{Limitations}\label{sec:limitations}
The shell-variation condition is sufficient, not necessary, and cannot be verified from rankings alone. Fixed-center constraints do not transfer across iterates; optimization convergence, observable radius selection, and joint $(d,R)$ optimality in the favorable regime remain open. Non-even recovery needs external calibration; the equal-axis initialization test has an $O(1/d)$ allowance scale and does not alone ensure chart acceptance. Empirical tasks span four fixed-split datasets mixing information use with tuning and query geometry.

\section{Conclusion}\label{sec:conclusion}
\rev{This paper establishes when antipodal batch rankings provably outperform within-pair signs for local gradient-direction recovery. By casting \emph{Mirror}, \emph{Reverse}, and \emph{Full} as directed-cycle conditions on a pair-offset difference graph, we show how reversed cross-pair comparisons cancel shared even curvature algebraically without Hessian estimation. Our analysis identifies relative shell variation $\tau_a$, rather than total curvature magnitude or condition number, as a key quantity governing one-round ranking gains. Whenever $\tau_a\le c\eps/\sqrt d$, one-round \emph{Reverse} localizes the entire feasible set within angle $\eps$ using $\widetilde O(d+\sqrt{d/\eps})$ uniform endpoints, breaking the $\Theta(d/\eps)$ sign barrier without knowing $\tau_a$. Conversely, when relative curvature scale is unbounded, complete rankings lose any uniform minimax evaluation advantage over signs across adaptive rounds $R$, while bounded odd-error allowances preserve feasibility under general smooth remainders and noise. Empirically, \emph{Full} separates strictly from \emph{Reverse} via longer cycles, and \emph{Reverse} consistently attains lower terminal optimization gaps than tuned \emph{Mirror} and unit-normalized \textsc{ZO-RankSGD} on ridge tasks under joint evaluation and round caps. Establishing \emph{Full}-over-\emph{Reverse} optimization gains, characterizing \emph{Mirror}-versus-\emph{Reverse} runtime scaling across dimensions, and developing calibration-free radius selection remain promising directions for future work.}

\label{page:main-end}

\clearpage
\subsection*{Artificial intelligence (AI) use statement}
AI assistants supported experimental-code implementation and debugging, figure preparation, feedback on evaluation protocols, and selected mathematical derivations. Writing assistance included structural revision, explanatory passages, and language editing based on author-provided material. The authors retain responsibility for the methodological decisions, scientific claims, and accuracy and integrity of the manuscript and accompanying artifacts.

\subsection*{Ethics statement}

This work develops and evaluates a methodology for comparison-based optimization. The empirical studies use synthetic objectives and existing public benchmark datasets. No new human-subject data were collected as part of this work. The experiments are methodological evaluations and do not involve deployment in clinical or other high-stakes decision-making settings.

\subsection*{Reproducibility statement}
\Crefrange{app:geometry}{app:bounded-error} provide the assumptions and proofs. \Cref{app:certificate} specifies the numerical certificate. The experimental protocols and resource accounting appear in \crefrange{app:structural-study}{app:computation}. The anonymous supplementary material includes companion data, per-run records, seed/split manifests, and executable experiment and checker code. These materials support independent reproduction of the reported numerical audits, summaries, and figures.

\clearpage
\bibliography{iclr2027_conference}
\bibliographystyle{iclr2027_conference}

\clearpage
\appendix
\crefalias{section}{appendix}
\crefalias{subsection}{appendix}
\crefalias{subsubsection}{appendix}
\crefalias{paragraph}{appendix}
\startcontents[appendices]

\begin{center}
    {\LARGE\bfseries Appendix}
\end{center}

\vspace{0.5em}

\section*{Contents}
\printcontents[appendices]{}{1}[2]{}

\vspace{1em}

\section*{Appendix Guide}
\Cref{app:geometry} develops the information model, and \Cref{app:one-round-proof,app:benign-lower,app:scale-proof} establish the two information regimes. Calibration and numerical certificates are in \Cref{app:bounded-error,app:certificate}. The experiments cover matched-endpoint information geometry (\Cref{app:structural-study}), regression optimization and resource attribution (\Cref{app:task-optimization}), non-even logistic validity and local-step controls (\Cref{app:logistic-tasks}), and synthetic validity and initialization (\Cref{app:validity-results}). \Cref{app:computation} reports computational costs. Primary and confirmation cohorts are kept separate.

\paragraph{Audit trail and artifact availability.}\label{app:artifact-scope}
The paths below locate comma-separated values (CSV) and JavaScript Object Notation (JSON) records in the anonymous supplementary material. Reproduction has three stages. First, regenerate the paid coordinates and rankings from the frozen task, split, and seed specifications. Second, check feasibility, cycle witnesses, error contracts, and support bounds independently of the fitting solver. Third, recompute the stated summaries with the original grouping and resampling units. A cycle check needs the candidate direction, ordered edges, and exact weights. A cap check needs the section, support directions, dual multipliers, and residual allowances. The rational example in \Cref{app:cycle-example} is fully specified in the text. 

\clearpage
\section{Model Validity and the Geometry of Ranking Constraints}\label{app:geometry}
\subsection{Normalization, closure, and nuisance gauges}
For a differentiable even $E$, $\nabla E(0)=0$. The directional derivatives in opposite directions are equal by evenness and opposite by differentiability, hence zero. Therefore $g$ in the model is the actual gradient at the fixed center. Dividing all scores by $\norm g>0$ and subtracting $b/\norm g$ gives \Cref{eq:scores}. On one physical shell $x_i=aW_i$, further division by $a$ gives normalized scores $\sigma u^\top W_i+E(aW_i)/(a\norm g)$. Across shells, the common normalization by $\norm g$ retains all physical coordinates and every cumulative ranking relation. Every true ranking relation is satisfied by $(u,e)$. Exact antipodality, rather than equality of nuisance values across different pairs, is the required structure.

The candidate vector $v$ may be any nonzero vector before normalization. The cone $K$ is positively homogeneous in $(v,z)$, so its unit-direction projection is exactly \Cref{eq:joint}. The projection $G=\operatorname{proj}_vK$ of the finite polyhedral cone is polyhedral and closed, so $\cC=G\cap\Sph^{d-1}$ is compact. No boundedness of the joint cone is assumed. Adding a common constant to every $z_i$ leaves all inequalities unchanged. A common gauge, such as $z_1=0$, removes this lineality. If only batch orders are retained, their disconnected components can have separate additive gauges. Cumulative orders connect the observed endpoints and use a common gauge. Repeated physical pairs always share their nuisance coordinate.

Replacing a strict observed order by a weak candidate inequality enlarges the strict consistency set. All upper bounds control this closed relaxation, including boundary directions with candidate ties. Tied observations impose both weak inequalities. No numerical equality is inferred from finite precision. Repeated pairs and mathematical ties do not affect the validity argument.

\paragraph{Instancewise small-radius limit.}\label{app:small-radius}
Differentiability and evenness give $\nabla E(0)=0$ and $E(x)=E(0)+o(\norm x)$. Thus
\[
\Omega_a(E)\le 2\sup_{\norm x=a}|E(x)-E(0)|=o(a),\qquad \tau_a\longrightarrow0.
\]
The quadratic subclass permits arbitrarily large anisotropic curvature, so convergence to this favorable regime is not uniform over the full class.

\subsection{Cycle characterization and short-cycle certificates}
The relation $(i,\sigma)\preceq(j,\zeta)$ is equivalent to
\begin{equation}\label{eq:difference-edge}
z_i\le z_j+w_{ji}(v),\qquad
w_{ji}(v)=\zeta v^\top x_j-\sigma v^\top x_i.
\end{equation}
Create an edge $j\to i$ of this weight. Summing around a directed cycle cancels the potentials, so a feasible $z$ requires every cycle to have nonnegative weight. Conversely, add an auxiliary source with zero-weight edges to all vertices. If there is no negative cycle, shortest-path distances from that source are finite and satisfy \Cref{eq:difference-edge}. They supply a feasible potential. This is the standard difference-constraint characterization \citep{cormen2022algorithms}, including self-loops and disconnected components. For a system of strict inequalities, all directed simple cycles must have positive weight. Subtracting a common positive number smaller than their minimum mean weight reduces sufficiency to the weak case. An acyclic graph permits any positive subtraction.

When positive endpoints are ordered increasingly and their negatives decreasingly, the two inequalities are
\[
v^\top(x_j-x_i)+(z_j-z_i)\ge0,\qquad
v^\top(x_j-x_i)-(z_j-z_i)\ge0.
\]
Their sum gives the Reverse constraint in \Cref{eq:reverse-cancellation}. Orienting either unsigned pair differently also gives rows proportional to $x_i+x_j$. The certifying outer cone retains all valid signed orientations and the self-loop Mirror rows. The localization proofs need only the positive-positive reverse relations. If $E$ is constant on the shell, every strict order of positive linear projections has the reverse order on the negative endpoints. The same unrestricted model consequently inherits all those linear-ranking inequalities, without requiring equal candidate potentials.

\rev{The concrete three-pair longer-cycle example and its exact negative-cycle witness (edges $1\to 2$, $2\to 3$, and $3\to 1$ in \Cref{eq:difference-edge} with weights $3/10$, $3/10$, and $-1$, summing to $-2/5$) are presented directly in \Cref{sec:cycles-main}.}

\subsection{A measurable consistent selector}\label{app:consistent-selector}
For continuous unsigned queries, exact mirror ties have probability zero for fixed $u$. Set $s_i=\sgn(u^\top x_i)$ from the observed mirror orders and $\ell=\sum_i s_ix_i$. Then
$\ell^\top u=\sum_i|u^\top x_i|>0$ almost surely. Maximize $\ell^\top v$ over $G\cap\{\norm v\le1\}$. A maximum exists by compactness and is positive. It has unit norm, since otherwise positive scaling in the cone improves it. The gradient maximizer is unique. Two distinct unit maximizers have a midpoint of norm less than one, which can again be scaled to improve the objective. Its nuisance witness need not be unique.

For each finite order pattern, the graph of this unique optimizer is semialgebraic in the query coordinates. Feasibility and optimality can be expressed using finitely many polynomial inequalities and quantified real variables. It is consequently measurable. Returning one fixed unit vector outside this domain makes the algorithm total and preserves measurability. The same whole-cell conclusion applies to any other measurable consistent selector, including a properly defined analytic-center gradient direction.

\FloatBarrier

\section{Proof of Whole-Cell Localization}\label{app:one-round-proof}
The argument combines regularity of the query distribution, endpoint-disjoint witnesses in two sampling regimes, and a margin-net argument. Both sampling regimes below concern the uniform one-round design of \Cref{thm:global}.
\subsection{Projection regularity for uniform queries}\label{app:regularity}
The density and transverse-tail bounds below have absolute constants and use exact spherical normalization.

\begin{lemma}[Central projections and transverse tails]\label{lem:regularity}
Let $d\ge3$, fix a unit $u$, and let $W$ be uniform on $\Sph^{d-1}$. For $S=u^\top W$ with cumulative distribution function $F$, there are positive constants $a_0,a_1,b_0,p_0$ and $0<q_0\le1/8$ such that a central interval $I$ contains all quantiles in $[1/2-q_0,1/2+q_0]$, and the following bounds hold on $I$ for every unit $w\perp u$:
\begin{equation}\label{eq:regularity}
\begin{gathered}
a_0\sqrt d\le f_S(t)\le a_1\sqrt d,\\
\Prob\{w^\top W\ge b_0/\sqrt d\mid S=t\}\ge p_0,\qquad
\Prob\{w^\top W\le-b_0/\sqrt d\mid S=t\}\ge p_0.
\end{gathered}
\end{equation}
The conditional bounds refer to a density version.
\end{lemma}

\begin{proof}
A coordinate of a uniform sphere point has density
\begin{equation}\label{eq:sphere-density}
f_d(t)=c_d(1-t^2)^{(d-3)/2},\quad |t|<1,\qquad
c_d=\frac{\Gamma(d/2)}{\sqrt\pi\,\Gamma((d-1)/2)}\asymp\sqrt d.
\end{equation}
On $I=[-1/(4\sqrt d),1/(4\sqrt d)]$ this density is bounded above and below by absolute multiples of $\sqrt d$, including $d=3$. Symmetry and integration give a fixed central quantile interval. Conditional on $S=t$, write $W=tu+\sqrt{1-t^2}V$, where $V$ is uniform on the sphere in $u^\perp\simeq\R^{d-1}$. For unit $w$ in that space,
\[
\E(w^\top V)^2=\frac1{d-1},\qquad
\E(w^\top V)^4=\frac3{(d-1)(d+1)}.
\]
Paley--Zygmund applied to $(w^\top V)^2$ gives a fixed positive probability of magnitude at least $1/\sqrt{2(d-1)}$. Central symmetry divides this probability equally between the two signs. Since $\sqrt{1-t^2}$ is bounded below on $I$, \Cref{eq:regularity} follows.

\end{proof}

\subsection{Separated witnesses from dependent rankings}\label{app:witness}
\begin{lemma}[Two-sided witness with fine and coverage regimes]\label{lem:witness}
Use the uniform query law and the projection and tail bounds of \Cref{lem:regularity}. Fix $0<\eta\le1$ and a unit alternative $v$ with $\eta\le\theta=\ang(u,v)\le\pi/2$. For $n$ independent unsigned queries, the probability that no pair satisfies
\begin{equation}\label{eq:two-sided}
u^\top(W_j-W_i)\ge a_*\eta/\sqrt d,\qquad
v^\top(W_j-W_i)\le-b_*\eta/\sqrt d
\end{equation}
is at most
\begin{equation}\label{eq:witness-failure}
Ce^{-cn}+C\exp\{-cn\min(n\eta,1)\}.
\end{equation}
The constants depend only on those in \Cref{lem:regularity}.
\end{lemma}
\begin{proof}
Write $v=\cos\theta\,u+\sin\theta\,w$, $w\perp u$, $\norm w=1$. Choose
\[
q_* =\min\{a_0b_0/16,q_0/8,1/16\},\qquad
a_*=q_* /(2a_1),\qquad b_*=b_0/2.
\]
A pair with both projections in $I$ satisfies \Cref{eq:two-sided} if
\begin{equation}\label{eq:pair-event}
\frac{a_*\eta}{\sqrt d}\le S_j-S_i\le\frac{b_0\eta}{4\sqrt d},\qquad
w^\top W_i\ge\frac{b_0}{\sqrt d},\qquad
w^\top W_j\le-\frac{b_0}{\sqrt d}.
\end{equation}
Indeed, $\sin\theta\ge2\theta/\pi$ and $\theta\ge\eta$ give
\[
v^\top(W_j-W_i)\le\frac{b_0\eta}{4\sqrt d}
-\frac{2b_0\sin\theta}{\sqrt d}\le-\frac{b_0\eta}{2\sqrt d}.
\]
Conditional on all projections, transverse coordinates of distinct endpoints are independent. An endpoint-disjoint matching selected using only the projections therefore supplies independent transverse success events, each of probability at least $p_0^2$.

\paragraph{Fine regime.}
Put $t=n\eta$ and first assume $t\le2/q_*$. The probability transform $U_i=F(S_i)$ is uniform. Its order-statistic spacings have the representation
\begin{equation}\label{eq:exponential-spacings}
U_{(j+1)}-U_{(j)}=Z_{j+1}/T_n,\qquad
T_n=\sum_{\ell=1}^{n+1}Z_\ell,\qquad Z_\ell\stackrel{\rm iid}{\sim}\operatorname{Exp}(1)
\end{equation}
\citep[Section 4.1]{pyke1965spacings}. This representation can be realized by sampling the exponentials, inverting $F$ for the ordered projections, drawing their independent conditional transverse components, and relabeling uniformly. Thus conditioning on all $Z_\ell$ preserves the transverse independence used here.

Choose every other adjacent pair between ranks
$\lceil(1/2-q_0/2)n\rceil$ and $\lfloor(1/2+q_0/2)n\rfloor$.
For $n$ above a constant threshold this gives $M\ge q_0n/4$ endpoint-disjoint pairs and distinct spacing exponentials. Let $H$ be the event that these projections lie in $I$ and $n/2\le T_n\le2n$. The central quantile margin in \Cref{lem:regularity} and binomial Chernoff bounds put all these ranks in $I$ except with probability $Ce^{-cn}$. Exponential Chernoff bounds control both tails of $T_n$, so $\Prob(H^c)\le Ce^{-cn}$.

For a selected spacing, require $q_*t\le Z_{j+1}\le2q_*t$. On $H$, integrating both density bounds gives
\[
\frac{Z_{j+1}}{2a_1n\sqrt d}
\le S_{(j+1)}-S_{(j)}
\le\frac{2Z_{j+1}}{a_0n\sqrt d}.
\]
This exponential interval implies the gap bounds in \Cref{eq:pair-event}. Its probability is
\[
p_t=e^{-q_*t}-e^{-2q_*t}\ge q_*t\,e^{-2q_*t}\ge q_*e^{-4}t.
\]
Let $J$ count selected exponentials in that interval. Conditional transverse independence bounds the probability of no witness on $H$ by $(1-p_0^2)^J$. Removing the indicator of $H$ in the upper bound and using independence of the selected exponentials before division by $T_n$ gives
\begin{align*}
\Prob(\text{no witness})
&\le\Prob(H^c)+\E(1-p_0^2)^J\\
&\le Ce^{-cn}+(1-p_0^2p_t)^M
\le Ce^{-cn}+\exp(-cn^2\eta).
\end{align*}
No independence between $H$ and $J$, or between normalized spacings, was used.

\paragraph{Coverage regime.}
Now $t\ge2/q_*$. Put $p=q_*\eta$, so $np\ge2$ and $p\le q_0/8$. Within $[1/2-q_0,1/2+q_0]$, pack
$K=\lfloor q_0/(2p)\rfloor\ge q_0/(4p)$ blocks of width $4p$. In a block starting at $a_\ell$, use intervals
\[
I_\ell^-=[a_\ell,a_\ell+p],\qquad
I_\ell^+=[a_\ell+2p,a_\ell+3p].
\]
Match as many endpoints as possible between the two intervals, choosing matches deterministically by original indices. Every matched gap in probability coordinates lies between $p$ and $3p$. The density bounds imply
\[
\frac{q_*\eta}{a_1\sqrt d}\le S_j-S_i
\le\frac{3q_*\eta}{a_0\sqrt d}\le\frac{b_0\eta}{4\sqrt d},
\]
which implies the gap conditions in \Cref{eq:pair-event}. These separated-block pairs need not be adjacent in the full ordering. Their true quantile positions are used only in the proof.

Let $N_\ell^-,N_\ell^+$ be the occupancy counts and
$M=\sum_\ell\min(N_\ell^-,N_\ell^+)$. Multinomial moments give
$\E N_\ell^\pm=np$ and $\E(N_\ell^--N_\ell^+)^2=2np$, regardless of dependence across bins. Consequently,
\[
\E\min(N_\ell^-,N_\ell^+)
=np-\tfrac12\E|N_\ell^--N_\ell^+|
\ge np-\sqrt{np/2}\ge np/2,
\]
and $\E M\ge Knp/2\ge q_0n/8$. Replacing one independent $U_i$ changes $M$ by at most two. The exposure martingale therefore has conditional increment ranges of length at most two. Iterating Hoeffding's exponential-moment bound \citep[Eq.~(4.16)]{hoeffding1963bounded} gives $\Prob(M-\E M\le-t)\le e^{-t^2/(2n)}$. Taking $t=q_0n/16$ yields
\[
\Prob(M<q_0n/16)\le\exp(-q_0^2n/512).
\]
Conditional on all projections, each of these disjoint pairs has transverse success probability at least $p_0^2$. Hence
\[
\Prob(\text{no witness})\le e^{-q_0^2n/512}+e^{-p_0^2q_0n/16}\le2e^{-cn}.
\]
Combining the regimes and enlarging $C$ to cover bounded $n$ establishes \Cref{eq:witness-failure}.
\end{proof}

\subsection{Whole-cell localization}\label{app:localization}
\subsubsection{Turning a separated witness into an observed inequality}
For this one-shell argument, use $e_i=E(aW_i)/(a\norm g)$ and $s_{i,\pm}=e_i\pm u^\top W_i$. For any actual nuisance vector with range at most $\tau$, a pair satisfying the first part of \Cref{eq:two-sided} and $\tau\le a_*\eta/(2\sqrt d)$ has
\[
s_{j,+}-s_{i,+}=u^\top(W_j-W_i)+(e_j-e_i)>0,\qquad
s_{j,-}-s_{i,-}=-u^\top(W_j-W_i)+(e_j-e_i)<0.
\]
Thus the observed ranking supplies a reverse-order halfspace excluding the witness alternative with a negative margin. Candidate potentials remain unrestricted. The witnessing event depends only on the fixed $u$ and the query directions. On this event, the implication holds simultaneously for all actual nuisance values of the stated range. The target is fixed before the random design is drawn.

\subsubsection{Proof of the global theorem}
Take a deterministic Euclidean net of the sphere with mesh
$\zeta=c_{\rm net}\eps/\sqrt d$ and at most $(C\sqrt d/\eps)^d$ points. Choose $c_{\rm net}$ sufficiently small relative to $a_*,b_*$. A feasible direction farther than $\eps$ from $u$ has a nearest net point separated by at least $\eps/2$. For net points at angle at most $\pi/2$, apply \Cref{lem:witness} with $\eta=\eps/2$. Taking $\tau\le a_*\eps/(4\sqrt d)$ makes the true gap enforce a reverse-order relation. The alternative margin is at most $-b_*\eps/(2\sqrt d)$. Replacing the net point by the original candidate changes the inner product with $W_j-W_i$ by at most $2\zeta$, so the same relation excludes that candidate.

For larger angles, use disjoint independent pairs of uniform endpoints. Their difference $D$ is rotationally invariant. Its direction is uniform and independent of its length, and $\Prob(\norm D\ge1)\ge1/2$. For a uniform direction $V$, let $Z$ be its projection length onto a plane containing $u$ and the alternative $v_0$. Spherical moments give $\E Z^2=2/d$ and $\E Z^4=8/(d(d+2))$. Thus $\Prob(Z\ge1/\sqrt d)\ge1/8$. Independently of $Z$, the polar angle is uniform. If $\ang(u,v_0)\ge\pi/3$, a polar interval of half-width $1/8$ about $(u-v_0)/\norm{u-v_0}$ gives
\[
u^\top D\ge1/(4\sqrt d),\qquad
v_0^\top D\le-1/(4\sqrt d)
\]
whenever $Z\ge1/\sqrt d$ and $\norm D\ge1$. For antipodal $u,v_0$, choose any plane through their line. This event has probability bounded below by an absolute positive constant. The inequality survives the nuisance and net perturbations after reducing the constants. The probability that these independent pairs yield no witness is at most $Ce^{-cn}$.

A union bound over the net gives
\begin{equation}\label{eq:global-union}
\Prob\{\cC_{\rm rev}\not\subseteq\cB(u,\eps)\}
\le C(C\sqrt d/\eps)^d
\left[e^{-cn}+e^{-cn\min(n\eps,1)}\right].
\end{equation}
The two requirements $n\ge CL_\eps$ and $n^2\eps\ge CL_\eps$ make this at most $\delta$. They follow from \Cref{eq:global-count}. Together with \Cref{eq:inclusions}, this establishes \Cref{thm:global}.

\subsubsection{Graceful degradation and the continuing value of mirrors}\label{app:degradation}
Let $c$ be small enough for \Cref{thm:global} and set
$\eps_{\rm out}=\max\{\eps,c^{-1}\sqrt d\,\tau\}\le1/10$.
The sample requirement in \Cref{eq:global-count} decreases as its precision argument increases, so the sample size for $\eps$ also suffices for $\eps_{\rm out}$. Applying the theorem at $\eps_{\rm out}$ gives $\cC_{\rm rev}\subseteq\cB(u,\eps_{\rm out})$ with probability at least $1-\delta$.

This is not a minimax accuracy floor. For any even nuisance, a mirror sign is $\sgn(u^\top W)$. Against an alternative at angle $\theta$, a uniform query has, with probability at least $c\theta$, opposite projections of magnitude at least $c\theta/\sqrt d$. Project onto the target--alternative plane, require planar radius at least $1/\sqrt d$, and choose a subwedge centered at $(u-v)/\norm{u-v}$. Its angular width is a constant multiple of $\theta$. A margin net then gives whole-cell recovery from mirrors whenever
\begin{equation}\label{eq:mirror-random}
n\ge \frac C\eps\left[d\log(C\sqrt d/\eps)+\log(C/\delta)\right],
\end{equation}
without any restriction on $\tau$. Full enforces these constraints as well. The sharper adaptive minimax achievability is proved in \Cref{app:mirror}.

For a quadratic nuisance, the Rayleigh-quotient extrema give
$\Omega_a(E)=a^2(\lambda_{\max}(H)-\lambda_{\min}(H))/2$, proving the expression following \Cref{eq:tau}. Isotropic curvature does not enter the fast condition.

\FloatBarrier

\section{Lower Bounds in the Small-Variation Regime}\label{app:benign-lower}
\begin{proposition}[Known isotropic subclass]\label{prop:benign-lower}
For $d\ge3$, $0<\eps\le1/10$, and $R\ge1$, every algorithm satisfying \Cref{eq:success} on $q_u(x)=u^\top x+\norm x^2/2$, with the Hessian known, obeys the following bound, where $A_\eps=(d-1)\log(1/\eps)$:
\begin{equation}\label{eq:benign-lower}
N\ge\max\left\{\frac{d\cos^2\eps}{2},\ \sqrt{\frac{(d-1)R}{2e}(\eps^{-1/R}-1)},\ \frac{A_\eps}{\log(2+A_\eps)}\right\}.
\end{equation}
The algorithm may use cumulative orders, arbitrary radii, and adaptive batch sizes.
\end{proposition}
For $R=1$, the middle term in \Cref{eq:benign-lower} is
$\sqrt{(d-1)(\eps^{-1}-1)/(2e)}$, which gives the stated
$\Omega(\sqrt{d/\eps})$ contribution.
\begin{proof}
Draw $u$ uniformly and fix a seed. As in \Cref{app:barrier}, unfold the finite tree. Nontrivial equality sets have zero spherical measure. Repeats add no partition. With $D=d-1$, the number of regions formed on $\Sph^{d-1}$ by $m\ge1$ affine hyperplanes is at most
$2\sum_{j=0}^D\binom{m-1}{j}\le(1+2em/D)^D$.
To see the first estimate, successively add a hyperplane and count the regions induced on its intersection sphere. The circle base case has at most $2m$ arcs. Empty, coincident, and tangent intersections only reduce the count. The binomial estimate is as in \Cref{eq:arrangements}. Set the count to one for $m=0$.

A round with $p$ old endpoints and $n$ new endpoints has at most $pn+\binom n2\le Nn$ new pairwise equalities. Each is an affine hyperplane in $u$, even for different physical radii. The remaining-budget recursion therefore bounds the number of leaves by
\[
\max_{\sum n_r\le N}\prod_{r=1}^R(1+2eNn_r/D)^D
\le(1+2eN^2/(DR))^{DR}.
\]
The cap bound from \Cref{app:barrier} gives average success at most
$\eps^D(1+2eN^2/(DR))^{DR}/2$.
Integrating over the independent seed and imposing average success at least $3/4$ establishes the square-root term in \Cref{eq:benign-lower}.

For the dimension term, give the algorithm the stronger oracle of exact numerical values of the known quadratic and allow adaptation after every endpoint. This supplies exact linear measurements of $u$. It suffices to treat $N<d$. Orthogonalizing and adding unused measurements when needed can only increase information, so the stronger transcript contains $N$ adaptively chosen orthonormal coordinates. Conditional on their values and the seed, the remaining component is uniform on a sphere in the unmeasured subspace. If $v_N$ is the measured projection and $r^2=1-\norm{v_N}^2$, reflection of the residual pairs two equally probable targets separated by angle $2\arcsin r$. When $r>\sin\eps$, no one estimate is accurate for both.

The conditional expected square of measurement $j+1$ is $(1-\norm{v_j}^2)/(d-j)$. Induction gives $\E\norm{v_N}^2=N/d$ under adaptive selection. Consequently
\[
\Prob(\mathrm{success})\le\tfrac12+\tfrac12\Prob(\norm{v_N}^2\ge\cos^2\eps)
\le\tfrac12+\frac{N}{2d\cos^2\eps}.
\]
Success probability $3/4$ requires $N\ge d\cos^2\eps/2$, also immediate for $N\ge d$. This applies a fortiori to rankings.

For the last term, fix a seed and label endpoints by evaluation order, using a fixed order within each batch. Pad stopped histories to $N$ labels. Outside the null equality sets of the finite tree, distinct transcript leaves induce distinct final permutations. Indeed, at their first differing feedback the queried points coincide and their prefix orders differ. Restriction of a common final permutation could not recover both. Identical-point ties are deterministic at that node and can be broken by label for this counting argument. Thus there are at most $N!$ leaves, even with fully sequential feedback. The same cap bound gives average success at most $N!\eps^{d-1}/2$. Averaging over seeds and using $3/4$ success yields $\log(N!)\ge A_\eps+\log(3/2)$, hence $N\log N\ge A_\eps$. If $N<A_\eps$, this implies $N>A_\eps/\log(2+A_\eps)$. If $N\ge A_\eps$, the displayed bound is immediate.
\end{proof}
In one round the bound is $\Omega(d+\sqrt{d/\eps})$. Every shell-variation class considered in \Cref{thm:global} contains this zero-variation subclass. The permutation bound also prevents unlimited feedback from removing precision dependence altogether. 

\FloatBarrier

\begin{corollary}[One-round Mirror lower bound]\label{cor:mirror-lower}
Let $d\ge3$ and $0<\eps\le1/10$. Any one-round procedure whose feedback consists only of antipodal signs and which satisfies \Cref{eq:success} on the known isotropic family $q_u(x)=u^\top x+\norm x^2/2$ must use
\begin{equation}\label{eq:mirror-one-round-lower}
N\ge\frac{d-1}{2e}\left[(3/2)^{1/(d-1)}\eps^{-1}-1\right]
=\Omega(d/\eps)
\end{equation}
endpoints. The same lower bound applies to every favorable class that contains this zero-variation family.
\end{corollary}
\begin{proof}
Fix the private seed and draw $u$ uniformly on $\Sph^{d-1}$. A one-round transcript with at most $N$ antipodal signs partitions the sphere with at most $C(N,d)$ homogeneous sign regions, where $C$ is defined in \Cref{eq:arrangements}. One output is associated with each region. The cap bound in \Cref{app:barrier} therefore limits the conditional success probability to $\eps^{d-1}C(N,d)/2$. Requiring average success of at least $3/4$ and using \Cref{eq:arrangements} gives \Cref{eq:mirror-one-round-lower}. Averaging over the private seed preserves the inequality, so failure below this budget occurs for at least one fixed $u$. The isotropic nuisance has $\tau=0$ on every shell.
\end{proof}

\section{The Unbounded-Scale Endpoint--Round Barrier}\label{app:scale-proof}
The lower bound permits full cumulative rankings. The matching construction uses only mirror signs and therefore also works with separate batch rankings and arbitrary even nuisance.
\subsection{The cumulative-ranking lower bound}\label{app:barrier}
\subsubsection{A finite tree before choosing an instance}
Fix an algorithm with hard budgets $N,R$ and include all randomization, including decoding, in its private seed $\omega$. Each weak ordering of at most $N$ endpoint labels belongs to a finite alphabet. For a fixed seed, unfolding every possible ordinal history through $R$ rounds therefore gives a finite tree. Pad stopped histories with empty rounds and use finite dummy queries on unreachable histories, enforcing the same hard budget. Let $\mathcal X(\omega)$ be the finite set of points appearing anywhere in this tree. It depends on the algorithm and seed, not on its subsequently chosen instance.

Fix $\kappa>1$ and put $\delta_\kappa=(\kappa-1)/(2(\kappa+1))\in(0,1/2)$. Independently draw $A=I+S$, with $S$ absolutely continuously distributed on the symmetric Frobenius ball $\norm S_F<\delta_\kappa$. Then
\[
\cond(A)<\frac{1+\delta_\kappa}{1-\delta_\kappa}
=\frac{3\kappa+1}{\kappa+3}<\kappa.
\]
For fixed $\omega$ and $xx^\top\ne yy^\top$, the equality
$\tr(A(xx^\top-yy^\top))=0$ is a proper affine hyperplane in the symmetric matrix variable. It has measure zero. Taking the finite union over tree pairs and integrating over the seed shows that
\begin{equation}\label{eq:hiding-variable}
\Gamma(\omega,A)=\max_{\substack{x,y\in\mathcal X(\omega)\\xx^\top\ne yy^\top}}
\frac{2\norm{x-y}}{|\tr(A(xx^\top-yy^\top))|}
\end{equation}
is measurable and almost surely finite, with an empty maximum equal to zero.

Choose one deterministic multiplier $t>0$ such that $\Prob_{\omega,A}(\Gamma(\omega,A)\ge t)\le1/8$. Draw $u$ independently and uniformly on $\Sph^{d-1}$ and set
$q(x)=u^\top x+(t/2)x^\top Ax$. The multiplier may depend on the prescribed algorithm, budgets, and $\kappa$, but not on the realized seed, matrix, or target. No moment bound on $\Gamma(\omega,A)$ is required. This is a distribution on fixed instances in $\cQ$.

On $\mathcal G=\{\Gamma(\omega,A)<t\}$, every distinct-projector pair anywhere in the tree obeys
\[
\left|\frac t2\tr(A(xx^\top-yy^\top))\right|>\norm{x-y}\ge|u^\top(x-y)|
\]
simultaneously for all unit $u$. Its score order is fixed by $A$. If $xx^\top=yy^\top$, then $y=x$ or $y=-x$ (including the zero case). Repeats are fixed ties. A nonzero antipode contributes only $\sgn(u^\top x)$. For a uniform $u$, directional ties have probability zero after a finite union over the tree. For fixed $(\omega,A)\in\mathcal G$, all remaining gradient information is consequently a tree of homogeneous sign tests.

At a node adding $n$ endpoints, at most $n$ previously unknown antipodal signs become available. Each newly completed equal-projector class requires at least one new endpoint, but its opposite endpoint may have been evaluated in an earlier round. This explains why a count of $n/2$ would be incorrect for cumulative rankings.

\subsubsection{Arrangements with history-dependent batch sizes}
Put $D=d-1$. The number of strict spherical regions formed by $n\ge1$ homogeneous hyperplanes in $\R^d$ is at most
\begin{equation}\label{eq:arrangements}
C(n,d)=2\sum_{j=0}^{d-1}\binom{n-1}{j}
\le(1+2en/D)^D,\qquad C(0,d)=1.
\end{equation}
The combinatorial formula follows by adding one hyperplane and counting the regions induced on it by its predecessors, with the one-dimensional and one-hyperplane base cases. For $1\le n\le D$, use $2^n\le(1+n/D)^D$. For $n\ge D+1$, the binomial bound
$\sum_{j=0}^D\binom{n-1}{j}\le(e(n-1)/D)^D$ follows by setting $z=D/(n-1)\le1$ in
$z^D\sum_{j\le D}\binom{n-1}{j}\le(1+z)^{n-1}\le e^D$.
These inequalities establish the displayed estimate. Restricting to a previous history does not increase the number of regions.

Let $L(N,R)$ bound the leaves of a sign tree with remaining budgets $N,R$. At a node with $n$ new endpoints, each of at most $C(n,d)$ children has at most $L(N-n,R-1)$ leaves. Induction yields
\[
L(N,R)\le\max_{n_1+\cdots+n_R\le N}\prod_{r=1}^R C(n_r,d)
\le(1+2eN/(DR))^{DR}.
\]
The final step uses concavity of $\log(1+2ex/D)$. The maximizing sequence is an upper bound on all subtrees, not an assumption that the actual algorithm fixes its allocation in advance. Empty rounds allow stopping.

\subsubsection{Cap coverage and randomized fixed-instance quantifiers}
The normalized measure of an angular $\eps$-cap on $\Sph^{d-1}$ is at most $\eps^D/2$. Indeed, using polar angle variable $\vartheta$,
\[
\frac{\int_0^\eps\sin^{D-1}\vartheta\,d\vartheta}{\int_0^\pi\sin^{D-1}\vartheta\,d\vartheta}
\le\frac{\eps^D/D}{2/D},
\]
using $\sin\vartheta\le\vartheta$ in the numerator. For the denominator, symmetry and $\cos\vartheta\le1$ on $[0,\pi/2]$ give
\[
\int_0^\pi\sin^{D-1}\vartheta\,d\vartheta
=2\int_0^{\pi/2}\sin^{D-1}\vartheta\,d\vartheta
\ge2\int_0^{\pi/2}\sin^{D-1}\vartheta\cos\vartheta\,d\vartheta=\frac2D.
\] For fixed $(\omega,A)\in\mathcal G$, there is one output per leaf. A union bound on its possible success caps gives conditional success probability at most $\eps^DL(N,R)/2$.

The event $\mathcal G$ involves only $\omega,A$, so $u$ is still uniform and independent after conditioning. Allowing success probability one on its complement gives
\[
\Prob(\mathrm{success})\le\tfrac18+\tfrac12\eps^D(1+2eN/(DR))^{DR}.
\]
If \Cref{eq:success} holds on every fixed member of $\cQ$, it holds on average under the independent instance distribution. Hence
\begin{equation}\label{eq:explicit-lower}
N\ge\frac{DR}{2e}\left[(5/4)^{1/(DR)}\eps^{-1/R}-1\right]
\ge\frac{DR}{2e}(\eps^{-1/R}-1).
\end{equation}
Violating this bound therefore forces failure on at least one fixed quadratic. Shrinking the open matrix support as $\kappa\downarrow1$ changes the quantile $t$, but no constant in the arrangement or cap argument. This establishes the uniformity claimed in \Cref{thm:barrier}.

\subsection{A finite Mirror construction for matching achievability}\label{app:mirror}
The construction in this section supplies the upper half of \Cref{thm:barrier}. It uses only the identity $f(aw)-f(-aw)=2a\norm g\,u^\top w$ and thus works for arbitrary even nuisance.

\subsubsection{First-round ratio information and subsequent refinement}
Draw a Haar-distributed orthonormal basis $(a_1,\ldots,a_d)$ and write $G_i=u^\top a_i$. Choose subdivisions $M_1,\ldots,M_s\ge2$, where $s\le R$ and $M_1$ is even. In the first batch, query mirrors of all $d$ axes. On every edge $(i,j)$ with $i\in\{1,2,3\}$ and $i<j\le d$, also query both lines
\begin{equation}\label{eq:mirror-grid}
\cos\theta\,a_i+\sin\theta\,a_j,\qquad
\sin\theta\,a_i-\cos\theta\,a_j,
\quad\theta=\frac{\ell\pi}{2M_1},\quad1\le\ell<M_1.
\end{equation}
There are $3d-6$ edges. The grid includes $\pi/4$, so after feedback the three candidate edges determine $h=\argmax_{i\le3}|G_i|$. All edges incident to this selected coordinate have already been queried in the same batch.

For each $i\ne h$, let $\psi_i=\arctan(G_i/G_h)\in(-\pi/2,\pi/2)$. Axis signs give its half-domain. To check the grid information, orient an edge temporarily as $(j,i)$ and write $\phi=\arctan(|G_i|/|G_j|)$. If the two coordinate signs agree, the second line in \Cref{eq:mirror-grid} compares $\phi$ with $\theta$. Otherwise the first compares it with $\pi/2-\theta$. Reversing the stored edge exchanges $\phi$ and $\pi/2-\phi$. These reflections preserve the grid. Thus each selected signed interval has width $\pi/(2M_1)$, with no separate initialization round.

In round $r\ge2$, subdivide each retained interval into $M_r$ equal cells. At every internal threshold $\theta$, query mirrors of
$z_{i,\theta}=\cos\theta\,a_i-\sin\theta\,a_h$. With $s_h=\sgn(G_h)$,
\[
s_hu^\top z_{i,\theta}=\frac{|G_h|}{\cos\psi_i}\sin(\psi_i-\theta),
\]
so its sign identifies the correct subinterval. With $Q=\prod_{r=1}^sM_r$, the final width is $\Delta=\pi/(2Q)$. For signed midpoints $m_i$, return the normalization of
$s_h(a_h+\sum_{i\ne h}\tan m_i\,a_i)$.
All coordinate zeros and the finitely many threshold equalities have probability zero under the Haar basis. Returning one fixed unit vector on such exceptional histories defines a measurable total algorithm without changing the conclusion. The paid cost is
\begin{equation}\label{eq:mirror-cost}
N=2d+4(3d-6)(M_1-1)+2(d-1)\sum_{r=2}^s(M_r-1).
\end{equation}

\subsubsection{Risk of the actual finite midpoint decoder}
For a uniform partition of $[0,\pi/2]$ with width $\Delta$, let $m$ be the midpoint of the cell containing $\phi<\pi/2$. Then
\begin{equation}\label{eq:midpoint}
|\tan m-\tan\phi|\le\Delta(1+\tan^2\phi).
\end{equation}
To verify this even in the last cell, use $|m-\phi|\le\Delta/2$, $m\le\pi/2-\Delta/2$, and
$\cos\phi\le\cos m+\sin m\sin(\Delta/2)\le2\cos m$.
The tangent difference identity then gives
$|\tan m-\tan\phi|=|\sin(m-\phi)|/(\cos m\cos\phi)\le\Delta/\cos^2\phi$.

Reflect signed cells onto their absolute half-domain and apply \Cref{eq:midpoint}. In basis coordinates, use $v_h=G_h$ and $v_i=G_h\tan m_i$ for analysis only. Normalizing this vector is exactly the implemented decoder. The coordinate error obeys
\begin{equation}\label{eq:mirror-error}
\norm{v-G}^2\le2\Delta^2\left((d-1)G_h^2+\sum_{i\ne h}\frac{G_i^4}{G_h^2}\right).
\end{equation}
Write $G=Z/\norm Z$ with independent standard Gaussian coordinates and $M=\max_{j\le3}|Z_j|$. The first term in parentheses has expectation at most 3. The two unselected candidate coordinates contribute at most 1. For $p=d-3>0$ and $S=\sum_{i=4}^dZ_i^2$,
\[
\sum_{i=4}^d\frac{G_i^4}{G_h^2}
\le\frac1{M^2}\frac{\sum_{i=4}^dZ_i^4}{S}.
\]
The two factors are independent. Since $M^2\ge(Z_1^2+Z_2^2+Z_3^2)/3$, the first expectation is at most $3\E[(\chi_3^2)^{-1}]=3$. Gaussian radial decomposition gives the second expectation $3p/(p+2)\le3$. When $p=0$ its contribution is absent. Thus the bracket expectation is at most $3+1+9=13$, with no assumption that the selected hub is independent of normalized coordinates.

The reverse triangle inequality and the unit chord formula imply
$\ang(v/\norm v,G)\le\pi\norm{v-G}$, so the actual finite decoder satisfies
\begin{equation}\label{eq:mirror-risk}
\E\ang(\widehat u,u)^2\le26\pi^2\Delta^2=\frac{26\pi^4}{4Q^2}.
\end{equation}

\subsubsection{Uniform optimization over a round cap}
Set $K_0=\pi^2\sqrt{26}$, $a_\eps=\log(1/\eps)$, and
$s=\min\{R,\lceil a_\eps\rceil\}$. Choose all $M_r=M$, the least even integer at least $(K_0/\eps)^{1/s}$. Then $Q\ge K_0/\eps$ and \Cref{eq:mirror-risk} is at most $\eps^2/4$. Markov's inequality gives \Cref{eq:success}. The cost is at most $12dsM\le Cds\eps^{-1/s}$. If $R\le a_\eps$, then $s=R$ and $\eps^{-1/R}\ge e$, yielding $CdR(\eps^{-1/R}-1)$. If $R>a_\eps$, then $s=\lceil a_\eps\rceil$ and the cost is $O(da_\eps)$. Use $R(e^{a_\eps/R}-1)\ge a_\eps$. This establishes the matching upper bound uniformly over $R$ and $\kappa$.

\FloatBarrier

\section{Model Error, Ambiguity, and Valid Retention}\label{app:bounded-error}

\subsection{Calibrated localization: formal statements}\label{app:robust-statements}
On a fixed shell, write the observed normalized scores as $e_i\pm(u^\top W_i+o_i)$, with $|o_i|\le\rho$ and actual even-offset range at most $\tau_a$. The shared-error cell $\cC_\rho$ consists of unit candidates $v$ for which free even offsets $e_i'$ and shared $b_i\in[-\rho,\rho]$ explain the ranking as $e_i'\pm(v^\top W_i+b_i)$. \rev{\Cref{cor:robust} is stated in \Cref{sec:calibration} and proved below; here we also state a fixed-center smoothness corollary that avoids third-derivative bounds.}

\begin{corollary}[Fixed-center recovery with a known smoothness scale]\label{cor:smooth-fixed}
Take $d,\eps,\delta$ as in \Cref{thm:global}. Suppose $\nabla f$ is $L$-Lipschitz on the Euclidean ball $\{x:\norm x\le a_0\}$, $L>0$, and $\norm{\nabla f(0)}\ge g_{\min}>0$. For a sufficiently small universal $c_0>0$, choose
\[
0<a\le\min\{a_0,c_0\eps g_{\min}/(L\sqrt d)\},\qquad
\rho=La/(2g_{\min}).
\]
Under exact score rankings, the shared-error model on $\{\pm aW_i\}$ contains the true gradient direction and satisfies the whole-cell conclusion and endpoint-complexity rate of \Cref{thm:global}. Exact evenness and a third-derivative bound are unnecessary.
\end{corollary}

\subsection{Why an exact-even fit can fail outside its model}
\begin{proof}[Proof of \Cref{prop:odd-shell}]
On the sphere, $\norm x^2=a^2$. The even term and cubic term have zero gradient at the origin, giving the stated directions and angle. Identical values imply identical transcript and output distributions, even for adaptive queries. The two angular success caps are disjoint, so their probabilities sum to at most one.
\end{proof}
Thus a valid certificate for the fitted direction set need not cover the true gradient outside the even-nuisance model. Shrinking the physical radius reduces this cubic bias, but retention is monotone under the add-only rule. If $u\notin\cC_1$, then $u\notin\cC_r\subseteq\cC_1$. Restoring validity in that case requires changing the model or discarding invalid constraints.

\subsection{A full shared-error model and its localization result}
At the fixed center, consider
\[
f(x)=b+g^\top x+E(x)+O(x),\qquad E(-x)=E(x),\quad O(-x)=-O(x),\quad\nabla O(0)=0.
\]
Thus $g$ remains the gradient at zero. On a radius-$a$ shell, let
$\rho_a=\sup_{\norm w=1}|O(aw)|/(a\norm g)$ and suppose a supplied envelope $\rho\ge\rho_a$ is available. Normalization gives the score model and shared-error cell $\cC_\rho$ defined in \Cref{app:robust-statements}. An adjacent-edge relaxation with independently assigned allowances need not enforce one globally shared score per endpoint and can be larger. This distinction does not assert strict enlargement for every complete pairwise relaxation.

\begin{proof}[Proof of \Cref{cor:robust}]
The true $u$, actual potentials, and actual odd values form a feasible witness. Consider $D=W_j-W_i$. If
\[
u^\top D>\tau_a+2\rho,
\]
then the positive scores increase from $i$ to $j$ and the negative scores decrease. Both inequalities hold irrespective of the allowed even and odd values. Adding the corresponding inequalities for any feasible candidate cancels $e_j'-e_i'$ and gives
\begin{equation}\label{eq:robust-reverse}
v^\top D\ge b_i-b_j\ge-2\rho.
\end{equation}
Hence a pair satisfying $u^\top D>\tau_a+2\rho$ and $v^\top D<-2\rho$ excludes $v$ from the Full shared-error model.

Use the separated-witness event of \Cref{app:witness,app:localization}, reserving a fixed fraction of its two margins for perturbations. For candidates at angular distance at least $\eps$, that event supplies positive and negative margins of order $\eps/\sqrt d$. The fine-spacing and coverage-block arguments give their stated common failure bound before any nuisance or error values are chosen. Taking $\tau_a+\rho\le c\eps/\sqrt d$ with $c$ smaller than both reserved margin constants makes the first inequality strict and the candidate inequality incompatible with \Cref{eq:robust-reverse}. The same finite net and its margin allowance then exclude all far candidates simultaneously. Increasing the universal sample constant absorbs the fixed margin reduction without changing the endpoint-complexity rate.

No independence among ranking edges or among bounded errors is used.
\end{proof}

More generally, let the actual odd envelope be $\rho_{\rm true}$ and the fitted envelope be $\rho_{\rm fit}\ge\rho_{\rm true}$. The forcing margin is $\tau_a+2\rho_{\rm true}$ and the exclusion margin is $2\rho_{\rm fit}$. The proof therefore gives the same count under
$\tau_a+\rho_{\rm true}+\rho_{\rm fit}\le c\eps/\sqrt d$.
At a nominal resolution $\eps$, applying this result at
$\max\{\eps,C\sqrt d(\tau_a+\rho_{\rm true}+\rho_{\rm fit})\}$ gives graceful degradation whenever that radius is at most $1/10$. This sufficient localization scale need not be an accuracy floor. The cubic ambiguity in \Cref{prop:odd-shell} supplies an instance-specific obstruction of order $\rho_a$.

For several physical shells, use physical budgets $\omega_i$ for normalized scores
$e_i\pm(u^\top x_i+b_i)$, so $|b_i|\le\omega_i$. A radius-$a$ shell has normalized envelope $\max_i\omega_i/a$. If all budgets are valid and all shared variables are retained, the full unit-direction sets are nested and contain the same $u$. Large but valid old error budgets do not create the exclusion mechanism of \Cref{prop:odd-shell}.

A measurable consistent selector exists. Bound the potential gauges as in \Cref{app:robust-chart} and take the lexicographically first unit direction in the compact feasible projection. Successive coordinate minima have semialgebraic graphs for each finite order pattern. For an implemented chart with $c/\norm h\ge1/2$, \Cref{eq:robust-sandwich} enlarges the fitted envelope by at most two, absorbed by the same one-round localization constants.

\begin{proposition}[Resolution of a robust uniform-query cell]\label{prop:error-cell-floor}
Fix a unit $u$, $n\ge1$, $d\ge3$, $\rho>0$, and $0<\delta<1$. For $n$ independent uniform query directions, observe the noiseless linear scores $\pm u^\top W_i$ and fit $\cC_\rho$. With probability at least $1-\delta$, its enclosing angular radius about any center is at least
\[
\frac12\min\!\left\{1,\rho\sqrt{\frac{d}{2\log(2n/\delta)}}\right\}.
\]
\end{proposition}

\begin{proof}
Write $\Delta=\min\{1,\rho\sqrt{d/[2\log(2n/\delta)]}\}$. Fix a unit $v$ with $\norm{v-u}=\Delta$ before drawing the queries. For unit $z$, the sphere moments satisfy
$\E(z^\top W)^{2k}=(2k-1)!!/[d(d+2)\cdots(d+2k-2)]\le(2k-1)!!/d^k$.
The moment generating function is therefore at most $\exp(t^2/(2d))$, giving
$\Prob\{|z^\top W|>t\}\le2\exp(-dt^2/2)$.
Take $z=(u-v)/\Delta$ and union-bound the $n$ projections. With probability at least $1-\delta$, every $b_i=(u-v)^\top W_i$ has magnitude at most $\rho$. The candidate scores $\pm(v^\top W_i+b_i)$ then equal all observed scores exactly, so both $u$ and $v$ belong to $\cC_\rho$. Their angular separation is $2\arcsin(\Delta/2)\ge\Delta$. The triangle inequality gives the radius bound.
\end{proof}
This is a finite-design whole-cell statement for the uniform query law, not a minimax lower bound against all adaptive designs. It shows why the $\sqrt d\,\rho$ resolution term cannot generally be removed from this shared-error cell analysis, up to logarithms, even when the actual instance has zero odd error.

\subsection{From score noise and local smoothness to an error envelope}
Suppose each stored endpoint score has error $\xi_{i,\pm}$ satisfying $|\xi_{i,\pm}|\le\nu$. On a radius-$a$ shell, set $\zeta=\nu/(a\norm g)$ and split the normalized errors as
\[
\bar\xi_i=\frac{\xi_{i,+}+\xi_{i,-}}{2a\norm g},\qquad
\widetilde\xi_i=\frac{\xi_{i,+}-\xi_{i,-}}{2a\norm g}.
\]
The first term is absorbed by the free even potentials and increases their range by at most $2\zeta$. The second is shared with opposite signs and has magnitude at most $\zeta$. Thus a fitted odd envelope $\rho\ge\rho_a+\zeta$ is valid, and the sufficient localization condition becomes
$\tau_a+2\zeta+\rho\le c\eps/\sqrt d$.
The observed ranking is the weak order of these stored noisy scores, with both relations retained for ties. This covers stochastic score perturbations while preserving transitivity. It does not model arbitrary cyclic pairwise judgments or partial rankings.

For a horizon of at most $N$ evaluations, conditionally $\sigma$-sub-Gaussian score errors satisfy
\[
\max_{j\le N}|\xi_j|\le\sigma\sqrt{2\log(2N/\delta_0)}
\]
with probability at least $1-\delta_0$, by conditional tail bounds and a union bound. This envelope also holds for adaptive query locations under the stated conditional tail assumption. Applying the localization result requires its own independent uniform query design and yields failure probability at most $\delta+\delta_0$; the envelope alone does not establish adaptive localization. Reusing a stored endpoint retains its realized noisy score. These error contracts refer to a calibrated score scale. An arbitrary unknown monotone link need not preserve their numerical size.

For a thrice differentiable objective with $\sup_{\norm x\le a}\norm{D^3f(x)}_{\rm op}\le M_3$, define
$E(x)=[f(x)+f(-x)]/2$ and $O(x)=[f(x)-f(-x)]/2-g^\top x$, absorbing the constant into $E$. Taylor's integral remainder gives $|O(aw)|\le M_3a^3/6$. A known lower bound $\norm g\ge g_{\min}>0$ therefore supplies the usable envelope
\[
\rho\ =\ \frac{M_3a^2}{6g_{\min}}+\frac{\nu}{ag_{\min}}.
\]
The two terms have opposite radius dependence. The even shell-variation condition remains separate. A third-derivative bound does not bound the anisotropic Hessian scale. The argument is at one fixed center and does not transfer retained constraints to a changing gradient.

\begin{proof}[Proof of \Cref{cor:smooth-fixed}]
Set $g=\nabla f(0)$ and write the first-order remainder as
$r(x)=f(x)-f(0)-g^\top x$. Smoothness gives $|r(\pm aw)|\le La^2/2$. Its even and odd parts
$E(x)=[r(x)+r(-x)]/2$ and $O(x)=[r(x)-r(-x)]/2$
satisfy $\Omega_a(E)\le La^2$ and $|O(aw)|\le La^2/2$. Thus $\tau_a\le La/g_{\min}$ and $\rho_a\le\rho$. Taking $c_0$ small enough in \Cref{cor:robust} establishes the result.
\end{proof}
The scale inputs are supplied rather than inferred from ordinal feedback, as in calibrated comparison-based recovery \citep{tao2026comparison}. The unknown-curvature hard class has no common finite $L$, so this consequence does not circumvent its uniform-budget barrier. With score error $\nu$, the sufficient condition instead includes $\nu/(ag_{\min})$, making arbitrarily small physical radii unsuitable.

\FloatBarrier

\section{Fitting and Certifying the Direction Set}\label{app:certificate}
The certificate encloses a projection of the fitted model. Its basic support inequality is
\begin{equation}\label{eq:dual-main}
c^\top v\le\alpha+M\norm{c+A^\top\lambda-\alpha\ell}_1,
\end{equation}
valid on a section with $Av\ge0$, $\ell^\top v=1$, $\norm v_\infty\le M$, and any $\lambda\ge0$. The section bound and residual-checked angular enclosure follow below.
\subsection{A bounded section in gradient variables}
Let $X$ have the unsigned stored query vectors as rows, and assume they span $\R^d$. Let $s_i$ be their exact observed Mirror signs, $A$ contain the Mirror and valid reverse-order rows, and $G_{\rm rev}=\{v:Av\ge0\}$. It contains the Full direction cell. Define $\ell=X^\top s$. For every nonzero $v\in G_{\rm rev}$,
\[
\ell^\top v=\sum_i s_i(Xv)_i>0.
\]
The summands are nonnegative, and a zero sum would imply $Xv=0$, contradicting full column rank. Every feasible ray therefore intersects $\ell^\top v=1$. On this section,
$\norm{Xv}_\infty\le1$ because each nonnegative mirror projection is bounded by their sum.

Suppose $L$ is a proposed left inverse with the exactly checked induced norm
$q=\norm{I-LX}_\infty<1$. Then
\[
\norm v_\infty\le\norm L_\infty\norm{Xv}_\infty+q\norm v_\infty,
\qquad
\norm v_\infty\le M:=\frac{\norm L_\infty}{1-q}.
\]
All quantities can be verified over the exact rationals represented by stored floating-point coordinates. The proposed inverse is used with its residual included in the bound. This bounds the gradient section independently of the nuisance potentials.

\begin{proposition}[Residual-compensated support and angular bounds]\label{prop:certificate}
Assume the preceding section bound. For any objective vector $c$, any $\lambda\ge0$, and any real $\alpha$, define
\begin{equation}\label{eq:certified-support}
U(c;\lambda,\alpha)=\alpha+M\norm{c+A^\top\lambda-\alpha\ell}_1.
\end{equation}
Then $c^\top v\le U(c;\lambda,\alpha)$ for every feasible section point. Let $h\ne0$, $P=I-hh^\top/(h^\top h)$, and obtain coordinate intervals $l_i\le(Pv)_i\le u_i$ and a lower bound $h^\top v\ge p>0$ from these supports. Then every feasible direction lies within angular radius
\begin{equation}\label{eq:angle-certificate}
\beta=\min\left\{4,\frac{\norm h\sqrt{\sum_i\max\{|l_i|,|u_i|\}^2}}{p}\right\}
\end{equation}
of $h/\norm h$, with outward rounding. If $p\le0$ or a bounded section has not been established, the trivial radius $4$ remains valid.
\end{proposition}
\begin{proof}
Write $r=c+A^\top\lambda-\alpha\ell$. On the section,
$c^\top v=\alpha-\lambda^\top Av+r^\top v\le\alpha+M\norm r_1$.
Use objectives $P_{i,:},-P_{i,:}$ for the intervals and $-h$ for the parallel lower bound. When $p>0$, the angle is below $\pi/2$ and
\[
\tan\ang(h,v)=\frac{\norm h\norm{Pv}}{h^\top v}
\le\frac{\norm h\sqrt{\sum_i\max\{|l_i|,|u_i|\}^2}}p.
\]
Since $\ang(h,v)\le\tan\ang(h,v)$, \Cref{eq:angle-certificate} follows. The radius $4>\pi$ covers all cases.
\end{proof}

The numerical optimizer proposes multipliers and primal points. Rational arithmetic independently checks multiplier nonnegativity, matrix products, norm bounds, and residual allowances, without relying on solver-reported dual feasibility or exact stationarity. A sparsified multiplier proposal is admissible only with its recomputed residual. Positive square roots and the reported radius are rounded outward. This produces a certificate for all directions allowed by the stored-coordinate model, rather than a statement based on the measured error at the true target.

\subsection{Transitive reduction and certificate completion}
The reverse constraints can be represented as comparisons among signed linear projections. Their strictly observed ordering is acyclic. If a directed comparison has a path through other retained comparisons, summing the endpoint differences along the path recovers its row. Removing that edge therefore leaves the gradient cone unchanged. The implementation checks that every removed edge has a retained path. This reduction leaves the Full decoder unchanged.

Coordinate boxes need not give tight angular caps even when the underlying cone is small. We tighten them using the exact tangent identity $h^\top Pv=0$. Whenever $h_i\ne0$,
\begin{equation}\label{eq:tangent-completion}
(Pv)_i=-\sum_{j\ne i}(h_j/h_i)(Pv)_j.
\end{equation}
Choosing each coordinate's upper or lower support according to its coefficient sign gives another bound. Intersecting with the existing interval preserves validity. The reported completion procedure uses two sweeps.

This operation also composes dual certificates. If $c=\sum_j a_jc_j$ with $a_j\ge0$, let $\lambda=\sum_j a_j\lambda_j$ and $\alpha=\sum_j a_j\alpha_j$. Then $\lambda\ge0$ and the exact residual is the corresponding linear combination. The triangle inequality makes the resulting support no larger than the same nonnegative combination of previous certified bounds. Applying \Cref{eq:tangent-completion} by selecting objectives $P_{j,:}$ or $-P_{j,:}$ keeps all coefficients nonnegative. The checker can therefore verify completed supports in the original certificate format.

\paragraph{An any-center Mirror lower bound.}\label{app:mirror-lower}
To obtain a lower bound $L$ on the Mirror cell's enclosing radius, optimize all positive and negative coordinate directions on its normalized section, with barrier levels $10^{-2}$ through $10^{-7}$. Mix each proposed section point with the strict center at weights $0.999$ and $0.001$. Only exactly Mirror-feasible witnesses are admitted. Exact integer Gram comparisons identify the most separated witness pair. A separate rational cosine check certifies a downward-rounded angle for this pair. Set $L$ to half this certified angle. The triangle inequality makes it a lower bound on the enclosing radius about \emph{any} center. If a checked upper-cap radius $U$ for Full satisfies $U<L$, the same paid ranking therefore certifies an information separation. This computation uses no new endpoint evaluations and does not depend on the pointwise error of a selected estimate.

\subsection{Joint analytic centers and their sections}\label{app:analytic-center}
\rev{We now detail the implementation and boundedness of the analytic-center program \Cref{eq:decoder-center} introduced in \Cref{sec:decoder}.} Full uses $2n-1$ adjacent-order rows and $d+n-1$ variables, enforcing cycles without enumerating them. Reverse uses $d$ variables and permits exact transitive reduction of redundant constraints. Returned directions are checked for feasibility, including the existence of shared offsets for Full.

For $n$ unsigned physical vectors, form features $(x_i,\mathbf e_i)$ and $(-x_i,\mathbf e_i)$, where $\mathbf e_i$ is the pair-indicator vector with one gauge coordinate omitted. Adjacent differences in the full order give a matrix $D$ in $d+n-1$ variables. Let $B$ be its row-normalized version. Every fit retains all rows.

The first fit uses the positive section normal
$e=B^\top w/\norm{B^\top w}$, where $w_i=\lambda_i+0.01/m$, $m$ is the row count, and $\lambda\ge0$ is the unit-ball margin solver's dual proposal. Its purpose is to choose a section and a strictly feasible point. If the initial point is not strictly feasible, invertible right whitening or a strict phase-I solve supplies another proposal. The original rows determine acceptance. For the final fit in $d>3$, use the canonical choice $e=B^\top\boldsymbol1/\norm{B^\top\boldsymbol1}$ and strict phase I. The $d=3$ fit uses the first section before the geometric selection below.

Both sections are bounded when the physical query matrix spans $\R^d$. Indeed, $By=0$ makes all endpoint scores equal. Subtracting each mirror pair gives $Xy_g=0$, hence $y_g=0$. The fixed gauge then forces all potentials to zero. Thus $B$ is injective. For any strictly positive $w$, the section $w^\top By=1$, $By\ge0$ bounds each slack by $1/w_i$, and injectivity bounds $y$. The log barrier has a unique minimizer in a nonempty strict section.

The center minimizes $-\sum_i\log(B_i^\top y)$ subject to $e^\top y=1$. With an orthonormal section basis $Q$, QR factorization of $\operatorname{diag}(By)^{-1}BQ$ computes Newton steps without forming normal equations. Line search preserves every strict inequality. There are at most 150 iterations, an internal decrement target of $10^{-7}$, and a final acceptance threshold of $10^{-5}$. The gradient direction must additionally admit an exact shared-potential witness.

The barrier's local metric is not, by itself, a representation-independent certificate of the full direction set's size. Repeating every row $m$ times leaves the feasible set and analytic center unchanged but multiplies the barrier Hessian by $m$. Its inverse shrinks by $1/m$ without reducing the direction set. Different row or section representations can change the selected center. Angular MSE comparisons therefore complement, rather than replace, the representation-independent inclusion witnesses and residual-checked caps.

\subsection{Feasible enclosing centers in three dimensions}\label{app:enclosing-center}
The spanning Mirror cone is pointed. We enumerate the rays obtained by intersections of pairs of boundary planes of the Reverse outer cone and retain only rays satisfying every row, using exact integer or rational arithmetic. Their nonnegative hull is the entire outer cone. For a proposed center $h$ whose inner products with all rays $r_j$ are positive, define
\begin{equation}\label{eq:ray-cap}
q(h)=\max_j\frac{\norm h^2\norm{r_j}^2-(h^\top r_j)^2}{(h^\top r_j)^2}.
\end{equation}
For every nonnegative ray combination, its tangent norm is at most $\sqrt{q(h)}$ times its parallel component, by the triangle inequality. An outward upper bound on $\sqrt{q(h)}$ therefore encloses the full direction set, since $\arctan(t)\le t$ for $t\ge0$.

The normalized rays, sums of ray pairs, and both normals through triples give candidate spherical enclosing centers. Choose a proposal that maximizes the minimum dot product with the rays. Mix this proposal with the analytic-center direction using weights $0.9999,0.99,0.9,0.5$. Normalize each mixture and solve the resulting strict difference constraints for the pair potentials. The margin proposals are $2^{-30}\min_i\norm{x_i}$ and $2^{-40}\min_i\norm{x_i}$. Only candidates satisfying every original ranking edge in exact arithmetic are admitted. The analytic center is also retained. Among these feasible candidates, choose the smallest verified ray cap with a deterministic tie rule. This defines a finite, observation-based selector.

Membership is checked for the returned normalized direction itself. An exact positive rational rescaling of the fitted potentials provides its witness, and every original adjacent edge is rechecked. These checks establish feasibility and cap coverage for the direction actually returned by the implementation.

\subsection{A computable chart that contains the robust Full model}\label{app:robust-chart}
The unit-sphere constraint in $\cC_\rho$ prevents directly reusing the exact-even homogeneous-cone implementation. We use a certified chart and retain the shared errors explicitly. The construction below takes coordinates, ranks, and error budgets. No target gradient or cardinal score enters it.

\subsubsection{Observable axis bound and a checked initialization criterion}
Express coordinates and physical error budgets in units of the first public probe radius. Pay for the standard-axis pairs $\pm e_j$ in these units in the first batch. Let $s_j\in\{-1,1\}$ be an observed mirror sign (choosing either sign for a tie) and $\omega_j\ge0$ its normalized physical error allowance. For every feasible unit direction $v$, sign disagreement implies $|v_j|\le\omega_j$. The simpler sufficient bound
\[
\ell^\top v\ge\norm v_1-2\sum_j\omega_j\ge1-2\sum_j\omega_j,
\qquad \ell=(s_1,\ldots,s_d),
\]
remains valid. For $S_2:=\sum_j\omega_j^2<1$, the observable-initialization experiments use the stronger bound
\begin{equation}\label{eq:robust-axis-section}
\ell^\top v\ge m_{\rm ax}:=\sqrt{1-S_2}-S_1,
\qquad S_1:=\sum_j\omega_j.
\end{equation}
To establish it, let $J_-$ be the coordinates with $s_jv_j<0$. Their squared norm is at most $S_2$, whereas the matching coordinates have Euclidean norm at least $\sqrt{1-S_2}$ and hence at least that much $\ell_1$ norm. The sum of the disagreeing absolute coordinates is at most $S_1$. Subtraction gives \Cref{eq:robust-axis-section}.

When $m_{\rm ax}>0$, set $m=m_{\rm ax}$ or use any positive outward-safe lower bound $m\le m_{\rm ax}$. Every original feasible direction then has a representative on $\ell^\top w=1$ with $\norm w_\infty\le1/m$. Mirror and reverse-order allowances on this section are $\omega_i/m$ and $(\omega_i+\omega_j)/m$. Intersect the implied rows with the coordinate box and apply residual-checked supports. A resulting enclosure with $h^\top v\ge c>0$ can define the next chart.

Each numerical refinement must preserve \emph{every} direction of the original shared-error model. A support-based norm or cosine tightening is accepted only after its residual-compensated bound is checked. These refinements consume no additional endpoint evaluations. A usable chart is accepted only if its checked error-envelope inflation is at most two. Otherwise the implementation reports that no usable chart was certified.
Keep the section normal $\ell$ fixed and initialize the outward norm bound $M=1/m$. Build its Mirror/Reverse polytope with physical allowances $M\omega_i$ and coordinate box $[-M,M]^d$. Compute an interior analytic-center proposal and certify its enclosure. Accept a center $h$ as soon as its checked cosine numerator $c>0$ satisfies $\norm h^2\le4c^2$. Otherwise tighten $M$ using the support box as follows. Write $z=P_{h^\perp}w$, let $[l_j,u_j]$ enclose each $z_j$, and let $[L_z,U_z]$ enclose $(P_{h^\perp}\ell)^\top z$. For $\ell^\top h\ne0$,
\[
\norm w^2\le\sum_j\max\{l_j^2,u_j^2\}
+\frac{\norm h^2}{(\ell^\top h)^2}
\max\{(1-L_z)^2,(1-U_z)^2\}.
\]
This follows from $w=z+h(1-\ell^\top z)/(\ell^\top h)$. Round the square root upward, verifying its square against the rational bound, and update only when it strictly reduces $M$. Recompute the interior point after an update. The first phase allows six such updates with the analytic center as enclosure proposal. A second phase allows three with proposal $\ell$. If a phase cannot tighten further, it ends. Every acceptance and update is checked against the enclosing polytope, so unsuccessful numerical proposals cannot remove original feasible directions.

The stronger bound is positive when $S_1^2+S_2<1$. Equal allowances $\omega_j=\rho$ therefore require $\rho<1/\sqrt{d(d+1)}$, improving the simpler $\rho<1/(2d)$ test but still having order $1/d$. It is not implied uniformly by every allowed $\rho=O(\eps/\sqrt d)$ in the information-theoretic localization result. The separate error-contract cohorts use their reported chart checks, rather than this refinement rule.

\subsubsection{Shared-error chart, retention, and support verification}
On $h^\top w=1$, fit the entire ranking with
\begin{equation}\label{eq:robust-chart}
\norm w_\infty\le1/c,\qquad |b_i|\le\omega_i/c,\qquad
\text{scores }e_i\pm(w^\top x_i+b_i),
\end{equation}
and free shared even potentials. If $h$ is unit, the normalized directions of this fitted set satisfy the sandwich
\begin{equation}\label{eq:robust-sandwich}
\cC_{\omega}\ \subseteq\ \cC_{\rm chart}\ \subseteq\ \cC_{\omega/c}.
\end{equation}
For the first inclusion, divide any unit candidate and its latent scores by $h^\top v\ge c$. For the second, $h^\top w=1$ gives $\norm w\ge1$. Normalizing a chart candidate can only reduce each error magnitude. With a nonunit stored $h$, replace the last envelope by $\omega\norm h/c$. Thus a certified cap with $c/\norm h\ge1/2$ costs at most a constant factor in the robust localization theorem. Adding $2d$ paid axis endpoints preserves this endpoint-complexity rate.

To obtain a bounded optimization problem, fix $e_1=0$ and choose a potential box without restricting projected directions. For fixed $w,b$, the ranking constraints are difference constraints in $e$. If $\norm w_\infty\le M$, $|b_i|\le B$, and $X_1=\max_i\norm{x_i}_1$, each edge weight has magnitude at most $2(MX_1+B)$. When the system is feasible, shortest-path potentials from a zero-weight supersource have range at most $2n(MX_1+B)$. Shifting by $e_1$ therefore permits the box $|e_i|\le2n(MX_1+B)+1$. This removes unbounded potential translations while preserving the candidate directions.

An analytic-center proposal uses all adjacent ranking constraints, the pair-error bounds, and these boxes on the chosen section. QR Newton steps avoid normal equations. A returned coefficient vector must pass exact checks of every ranking inequality and shared-error bound. The same initial chart is reused across later shells, retaining each endpoint's physical budget and every earlier constraint. These fitted models are nested and always contain the true direction when the contracts hold.

For certification, eliminate the even potentials using Mirror and reverse-order relations, now with their bounded-error allowances. On a section $\ell^\top w=1$, let $Gw\le b$ be this gradient outer polytope and $\norm w_\infty\le M$. Any $\lambda\ge0$ gives the exactly checkable support bound
\begin{equation}\label{eq:robust-dual}
q^\top w\le\lambda^\top b+\alpha+
M\norm{q-G^\top\lambda-\alpha\ell}_1.
\end{equation}
Tangent-coordinate bounds and a positive parallel bound yield a cap as in \Cref{app:certificate}. If its squared tangent ratio is $Q$ about stored center $h$, then any positive rational $c$ with $c^2(1+Q)\le\norm h^2$ is a valid lower bound on $h^\top v$ for unit candidates. All support products, residual allowances, and this last inequality are recomputed in rational arithmetic. Loose bounds affect informativeness, not validity.

\FloatBarrier

\section{Information Geometry: Scale, Nuisance Rank, and Cycles}\label{app:structural-study}
These experiments measure information under matched endpoints. Scale and nuisance-family controls assess when cross-pair orders are useful. Exact cycle witnesses distinguish Full from Reverse. The cubic branches test model validity and are reported in \Cref{app:validity-results}.
\subsection{Structural experiments: protocol}\label{app:scope-protocol}
\subsubsection{Targets, paired designs, and endpoint accounting}
The structural study has 96 independent target--axis units, 32 at each $d\in\{3,8,20\}$, separate from \Cref{app:first-batch-audit}. Normalize independent Gaussian vectors to obtain $u,z$, and normalize $z-(z^\top u)u$ to obtain $q$. Generate $16d$ uniform unsigned directions from normalized Gaussian draws. Budget $N$ uses the first $N/2$ directions and their antipodes, sharing targets, axes, and prefixes across coefficients and families. Exact seed formulas for all studies are in the companion \texttt{seed\_manifest.csv}.

The experiment matrix uses the following settings.
\begin{itemize}
\item Quadratic settings use $N/d\in\{4,8,16,32\}$ and $\lambda\in\{0,10^{-3},10^{-2},10^{-1},1,10^2,10^4\}$, giving 2,688 conditions.
\item Quartic even settings use $N/d\in\{8,32\}$ and $\lambda\in\{10^{-3},10^{-1},1\}$, giving 576 conditions.
\end{itemize}
These 3,264 even-family conditions reuse 96 independent units; MSEs average all 32 targets per dimension. The odd-remainder branch uses the same units and is specified with its retention policies and diagnostics in \Cref{app:odd-experiments}.

\subsubsection{Function families and physical shells}
For positive $\lambda$, the quadratic family is
\begin{equation}\label{eq:scope-quadratic}
f_\lambda(x)=u^\top x+\tfrac12t\norm x^2+\lambda(z^\top x)^2,
\qquad t=100\lambda.
\end{equation}
For unit $u,z$, $H=t(I+0.02zz^\top)$ has condition number $1.02$ and $\tau_1=\lambda$. At $\lambda=0$, $t=1$ gives the isotropic baseline. The quartic family uses
\begin{equation}\label{eq:scope-quartic}
f_\lambda(x)=u^\top x+\tfrac12\norm x^2+\lambda(z^\top x)^4.
\end{equation}
Its nominal unit-shell variation is also $\lambda$, with no quadratic representation of the last term. Both families use one uniform antipodal batch at physical radius one. The joint decoder is given no coefficient, curvature, or nuisance-family information.

All coordinates and polynomial coefficients are stored in float64. Exact checks treat these stored values as binary rationals. Consequently nominal unit norms and orthogonality need not hold as real-arithmetic identities at the saved points. The evaluator uses their actual norms and coordinates.

\subsubsection{Recovery and numerical verification}
The Full solver and selector are specified in \Cref{app:certificate}. Python 3.12 with NVIDIA's CUDA platform computes numerical proposals in float64 precision. Exact CPU arithmetic checks their feasibility and bound residuals. A float64 score enclosure detects ambiguous polynomial orders, which are resolved at exact stored-coordinate values. The decoder receives only coordinates and integer ranks.

For each proposed or true direction, exact difference constraints produce shared potentials or a negative-cycle witness, testing membership in the Full projection rather than its Reverse outer cone. Independent checks verify returned-direction feasibility, outer caps, and Mirror lower-bound witnesses (\Cref{app:certificate,app:mirror-lower}). All planned branches are checked, and completed even fits meet the analytic-center residual requirement. Projection and replay checks use stored observations, adding neither endpoints nor independent targets.

\subsubsection{Exact comparisons of three direction sets}\label{app:three-cones}
The projection audit uses all 32 three-dimensional targets at $N/d=8,32$ and $\lambda=0,\allowbreak10^{-3},\allowbreak10^{-1},\allowbreak1,\allowbreak10^4$, for 320 conditions. It starts from the exact Mirror and Reverse cones. Each extreme ray of the Reverse cone is tested against all ranking difference constraints. An infeasible ray supplies a negative cycle and hence a valid gradient halfspace. Add that halfspace, recompute the exact extreme rays, and continue until every ray admits shared potentials.

At termination, the resulting cone equals the Full gradient projection. Every cut is implied by the ranking, so the Full projection is contained in the computed cone. Conversely, feasibility of all generating rays puts their nonnegative hull inside the convex projected cone. A removed Reverse ray establishes a strict inclusion. Equality with the Mirror cone is checked by exact canonical extreme rays. These operations use the saved endpoints and require no new evaluations.

For each cone, the most separated ray pair supplies an any-center angular-radius lower bound. A common strictly feasible unit center supplies a tangent-ratio upper bound. If every ray $v$ has $h^\top v>0$, then
\[
\ang(h,v)=\arctan\!\left(\frac{\norm{v-(h^\top v)h}}{h^\top v}\right)
\le\frac{\norm{v-(h^\top v)h}}{h^\top v}.
\]
Exact Gram comparisons and outward rounding certify the maximum ratio. It also bounds the cone's normalized convex combinations. The reported Full/Reverse cap ratio uses this same center. It is not the ratio of optimal enclosing radii.

\subsection{Structural experiments: supporting results}\label{app:scope-results}
\subsubsection{Scale dependence and finite-budget recovery}
\Cref{tab:scope-quadratic} reports absolute angular MSEs, complementary to the information ratios in \Cref{fig:scale}. Larger budgets reduce MSE at every tested scale, but large-scale errors remain higher despite condition number $1.02$. All 2,688 quadratic conditions preserve true-direction feasibility and cap coverage. Losing ranking gain is not model invalidity or loss of all mirror information.

\begin{table}[tbp]
\centering\small
\caption{One-round Full angular MSE for quadratic nuisance (rad$^2$). Each entry averages 32 independent targets per dimension. Targets and query prefixes are shared across budgets and shell-variation levels.}\label{tab:scope-quadratic}
\begin{tabular*}{\linewidth}{@{\extracolsep{\fill}}rrcccc@{}}
\toprule
$d$ & $\lambda$ & $N/d=4$ & $8$ & $16$ & $32$ \\
\midrule
3 & $0$ & $1.669\!\times\!10^{-2}$ & $1.044\!\times\!10^{-3}$ & $4.299\!\times\!10^{-5}$ & $3.877\!\times\!10^{-6}$ \\
3 & $10^{-3}$ & $1.669\!\times\!10^{-2}$ & $1.044\!\times\!10^{-3}$ & $4.291\!\times\!10^{-5}$ & $4.040\!\times\!10^{-6}$ \\
3 & $10^{-2}$ & $1.694\!\times\!10^{-2}$ & $9.970\!\times\!10^{-4}$ & $3.978\!\times\!10^{-5}$ & $6.087\!\times\!10^{-6}$ \\
3 & $10^{-1}$ & $1.999\!\times\!10^{-2}$ & $1.307\!\times\!10^{-3}$ & $1.867\!\times\!10^{-4}$ & $3.647\!\times\!10^{-5}$ \\
3 & $1$ & $4.119\!\times\!10^{-2}$ & $5.796\!\times\!10^{-3}$ & $1.344\!\times\!10^{-3}$ & $2.233\!\times\!10^{-4}$ \\
3 & $10^2$ & $2.039\!\times\!10^{-1}$ & $4.622\!\times\!10^{-2}$ & $1.895\!\times\!10^{-2}$ & $3.042\!\times\!10^{-3}$ \\
3 & $10^4$ & $2.202\!\times\!10^{-1}$ & $7.419\!\times\!10^{-2}$ & $2.836\!\times\!10^{-2}$ & $5.807\!\times\!10^{-3}$ \\
\addlinespace
8 & $0$ & $5.496\!\times\!10^{-3}$ & $1.502\!\times\!10^{-4}$ & $8.985\!\times\!10^{-6}$ & $5.078\!\times\!10^{-7}$ \\
8 & $10^{-3}$ & $5.452\!\times\!10^{-3}$ & $1.506\!\times\!10^{-4}$ & $9.195\!\times\!10^{-6}$ & $5.091\!\times\!10^{-7}$ \\
8 & $10^{-2}$ & $5.651\!\times\!10^{-3}$ & $1.457\!\times\!10^{-4}$ & $1.323\!\times\!10^{-5}$ & $9.758\!\times\!10^{-7}$ \\
8 & $10^{-1}$ & $4.805\!\times\!10^{-3}$ & $1.995\!\times\!10^{-4}$ & $2.082\!\times\!10^{-5}$ & $2.246\!\times\!10^{-6}$ \\
8 & $1$ & $1.060\!\times\!10^{-2}$ & $3.325\!\times\!10^{-4}$ & $3.472\!\times\!10^{-5}$ & $3.840\!\times\!10^{-6}$ \\
8 & $10^2$ & $2.112\!\times\!10^{-1}$ & $2.134\!\times\!10^{-2}$ & $4.510\!\times\!10^{-3}$ & $4.854\!\times\!10^{-4}$ \\
8 & $10^4$ & $3.552\!\times\!10^{-1}$ & $7.441\!\times\!10^{-2}$ & $2.484\!\times\!10^{-2}$ & $7.232\!\times\!10^{-3}$ \\
\addlinespace
20 & $0$ & $9.471\!\times\!10^{-4}$ & $3.739\!\times\!10^{-5}$ & $1.525\!\times\!10^{-6}$ & $1.007\!\times\!10^{-7}$ \\
20 & $10^{-3}$ & $9.405\!\times\!10^{-4}$ & $3.813\!\times\!10^{-5}$ & $1.600\!\times\!10^{-6}$ & $1.179\!\times\!10^{-7}$ \\
20 & $10^{-2}$ & $8.800\!\times\!10^{-4}$ & $3.299\!\times\!10^{-5}$ & $2.087\!\times\!10^{-6}$ & $1.968\!\times\!10^{-7}$ \\
20 & $10^{-1}$ & $7.888\!\times\!10^{-4}$ & $4.340\!\times\!10^{-5}$ & $3.904\!\times\!10^{-6}$ & $3.534\!\times\!10^{-7}$ \\
20 & $1$ & $1.267\!\times\!10^{-3}$ & $6.208\!\times\!10^{-5}$ & $5.090\!\times\!10^{-6}$ & $4.788\!\times\!10^{-7}$ \\
20 & $10^2$ & $9.628\!\times\!10^{-2}$ & $6.874\!\times\!10^{-3}$ & $4.720\!\times\!10^{-4}$ & $4.504\!\times\!10^{-5}$ \\
20 & $10^4$ & $3.431\!\times\!10^{-1}$ & $7.938\!\times\!10^{-2}$ & $2.423\!\times\!10^{-2}$ & $5.577\!\times\!10^{-3}$ \\
\bottomrule
\end{tabular*}
\end{table}

\subsubsection{Nonquadratic even nuisance}
\begin{figure}[tbp]
\centering\includegraphics[width=\linewidth]{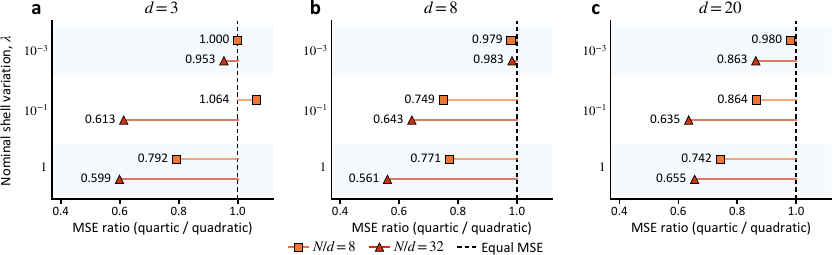}
\caption{Recovery under quartic and quadratic even nuisance. Labels show quartic/quadratic MSE ratios over 32 matched targets at matched nominal shell variation. Shell-variation levels are tested categories, not a continuous logarithmic axis. Segments connect estimates to equality; they are not confidence intervals.}\label{fig:scope-quartic}
\end{figure}

\Cref{fig:scope-quartic} compares all 18 matched dimension--budget--variation configurations. At $\lambda=1$, quartic MSE is lower in every configuration. Shell variation gives a worst-case guarantee but does not uniquely determine realized information. All 576 quartic conditions preserve true-direction feasibility, cap coverage, and $U<L$ without fitting either nuisance family. Absolute quartic MSEs are retained in the companion \texttt{table04\_quartic\_mse.csv}.

\subsubsection{The additional information in longer cycles}

\begin{figure}[t]
\centering\includegraphics[width=\linewidth]{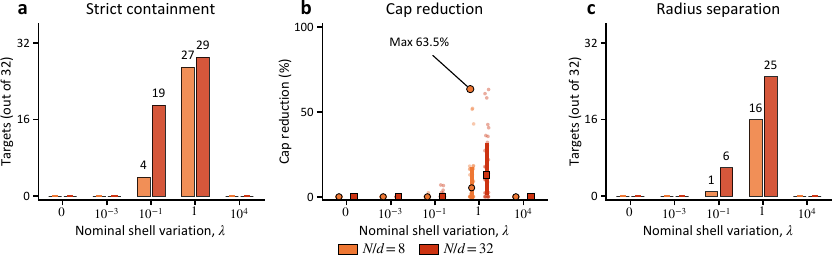}
\caption{Exact three-dimensional projected-cone geometry. Each shell-variation level and budget uses 32 targets. (a) Counts show strict inclusion of Full in Reverse. (b) $100(1-U_F/U_R)$ gives the percentage reduction in upper cap radii about a common center. Small points show targets, large marks medians, and bars interquartile ranges. (c) Counts show $U_F<L_R$, where $L_R$ is the Reverse any-center lower bound. Shell-variation levels are tested categories. All 320 conditions use matched endpoints.}\label{fig:scope-cycles}
\end{figure}

The 79 strict inclusions and 48 radius separations in \Cref{fig:scope-cycles} are distinct diagnostics. Available bounds need not detect every smaller cell. At the largest scale, all 64 checked Full cones equal their Mirror cones. These are finite-design observations; \Cref{thm:barrier} supplies the minimax conclusion.

\subsection{Matched rank-one and higher-rank anisotropy}\label{app:high-rank}

Recovery under rank-one anisotropy can become easier as the dimension grows. To test a harder nuisance geometry, we compare
\begin{equation}\label{eq:high-rank-family}
f_k(x)=g^\top x+50\lambda\norm x^2+\lambda\norm{P_kQ^\top x}^2,
\qquad k\in\{1,d/2\},
\end{equation}
where $Q$ is a rational orthogonal reflection and $P_k$ selects $k$ coordinates, giving anisotropic rank $k$. Both families have Hessian eigenvalues ranging from $100\lambda$ to $102\lambda$ and nominal unit-shell variation $\lambda$. At fixed $\lambda/\norm g$, the nuisance-to-signal standard-deviation ratio under uniform spherical queries decays as $d^{-1/2}$ for rank one but stays of constant order for rank $d/2$.

Twelve independent target--reflection--design units, four at each $d=8,20,50$, are reused across $N/d=8,16$, $\lambda=10^{-3},10^{-1},1$, and both ranks. All 432 model outputs are feasible, and all 144 conditions certify separation between Full and Mirror. Exact negative cycles establish strict inclusion of Full in Reverse in 49/72 rank-one and 55/72 high-rank conditions (\Cref{fig:new-high-rank}). A separate cohort, evaluated under the same frozen rules, confirms both types of information gain.

MSE also depends on the analytic-center selector; the exact cycle witnesses do not. With four independent units per dimension, rank-dependent MSE contrasts are exploratory (\Cref{app:high-rank-statistics}).

\subsubsection{Design and the controlled difficulty change}
The matched-rank study uses \Cref{eq:high-rank-family} with twelve independent target--reflection--design units, four at each $d=8,20,50$. The settings are $N/d\in\{8,16\}$, $\lambda\in\{10^{-3},10^{-1},1\}$, and $k\in\{1,d/2\}$. Target and query directions are independent isotropic draws. The 144 family conditions give 432 Mirror/Reverse/Full outputs. All three models use the same endpoints within a condition.
The reflection is $Q=I-2zz^\top/(z^\top z)$ for a stored nonzero integer vector $z$, with rational entries. Gaussian targets and query vectors are normalized in float64. Exact ordering and support checks use those stored coordinates. The seed manifest and source snapshots fix target, reflection, and query-prefix generation. The companion file \path{high_rank_all_outputs.csv} contains all 432 model outputs. The file \path{high_rank_all_groups.csv} contains 108 model--dimension--budget--scale--rank summaries ($54\times2$, one per Mirror/Reverse/Full rank level). Both reside under \path{data/integrated_20260918/}.

A separate confirmation retains the frozen query, fit, and certificate rules. It uses twelve additional units, with four per dimension, at $N=8d$ and $\lambda=0.1$ (base seed $8818091901$). All 72 model runs complete. All 24 rank-family conditions certify separation between Full and Mirror. Each also admits a Reverse-feasible direction that Full excludes through a negative cycle. For the high-rank family, the confirmation-only pooled Full/Mirror and Full/Reverse MSE ratios are $0.0009903$ and $0.4784$. These ratios use a different aggregation scheme from the primary figure. The cohort is reported separately in \path{confirmation_high_rank_outputs.csv}.

The comparison matches the minimum and maximum eigenvalues and the condition number while changing the rank of the anisotropic part. The ideal uniform-sphere moments in \Cref{app:regularity} explain the resulting change in difficulty. For unit $W$ uniform on the sphere and any fixed orthogonal $Q$,
\[
\operatorname{Var}\!\left(\norm{P_kQ^\top W}^2\right)
=\frac{2k(d-k)}{d^2(d+2)},\qquad
\operatorname{Var}(g^\top W)=\frac{\norm g^2}{d}.
\]
Consequently the standard-deviation ratio of the anisotropic nuisance to the linear signal is
\[
\frac{\lambda}{\norm g}\sqrt{\frac{2k(d-k)}{d(d+2)}}.
\]
For $k=1$ this decreases as $d^{-1/2}$. For $k=d/2$ it stays of constant order at fixed $\lambda/\norm g$. These ideal-sphere moments describe typical, rather than worst-case, variation and require no Haar assumption on $Q$.

\subsubsection{Point loss, set witnesses, and computation}\label{app:high-rank-statistics}
Under high-rank nuisance, Full's pooled MSE is $1.447$ times its value under rank-one nuisance (marginal paired 95\% interval $[1.175,2.127]$). This contrast is exploratory. Unlike exact cycle witnesses, MSE also depends on the analytic-center selector.

The marginal interval above uses 10,000 dimension-stratified target resamples, preserving both budgets, all three scales, and both ranks within an independent unit. With four independent units per dimension, repeated conditions increase within-unit coverage of settings but do not create additional independent targets.

In \Cref{fig:new-high-rank}(a), Full/Mirror MSE ratios pool both budgets and all three scales \emph{within each unit and rank}. In panel (b), Full/Reverse ratios pool both budgets within each unit, rank, and fixed scale. If $A_j$ and $B_j$ are these paired mean squared angular errors for the numerator and denominator models, the pooled statistic is $\sum_j A_j/\sum_j B_j$. Within-unit ratios $A_j/B_j$ are paired diagnostics; their arithmetic mean generally differs from the pooled statistic. Panel (a) has four such units per dimension and panel (b) twelve per scale. Intervals concern pooled estimates, not individual settings. Shell-variation levels in \Cref{fig:new-high-rank} are tested categories. Panel (b)'s small-point shapes indicate dimension, and connecting lines are not confidence bands.

At $\lambda=10^{-3}$, the exact negative-cycle witness counts are 1/24 for rank-one nuisance and 7/24 for high-rank nuisance. At each of $\lambda=10^{-1},1$, both families have 24/24. A witness provides a direction allowed by Reverse but ruled out by a longer ranking cycle. Failure to find one does not establish equality. Unlike the three-dimensional exhaustive projection audit, this experiment does not enumerate the complete high-dimensional projected cone or its optimal enclosing radius.

The Full and Reverse selectors use different analytic-center barriers (\Cref{app:certificate}), and the high-dimensional Full upper cap encloses Reverse. At $d=50$, median Full/Reverse fitting times are $0.323/0.080$ seconds, whereas both certificate times are approximately $17.68$ seconds. Hardware and instance differences preclude a direct timing comparison with \Cref{app:computation}.

\FloatBarrier

\subsection{Additional independent first-batch information audit}\label{app:first-batch-audit}
This audit uses the saved first batches of a separate two-round cohort. Its objective is
\begin{equation}\label{eq:experiment-target}
f(x)=u^\top x+\tfrac12\norm x^2+\lambda(z^\top x)^2,\qquad \norm u=\norm z=1.
\end{equation}
Independent normalized Gaussians give a fixed target--axis pair $(u,z)$. There are 256 independent units at each $d\in\{3,8,20\}$, separate from the structural cohort. Each is reused across the original total-budget labels $B/d\in\{16,32\}$ and $\lambda\in\{0.001,0.1\}$, giving 768 units and 3,072 first-batch conditions. A saved first batch consists of $B/4$ independent uniform unsigned directions and their antipodes at radius one, totaling $B/2\in\{8d,16d\}$ endpoints. Only this first batch contributes to the reported observations and endpoint costs. The companion seed manifest specifies the fixed designs.

Stored float64 normalizations need not satisfy ideal unit norms exactly. For the polynomial defined by stored $u,z$, the ideal radius-$a$ relative shell variation is $a\lambda\norm z^2/\norm u$. Actual stored-coordinate evaluation also includes the isotropic term. Float64 polynomial scores use a $10^{-12}$ coarse enclosure, with ambiguous adjacent orders resolved by exact binary-rational values. The decoder receives only coordinates and ranks. The numerical fit and independent feasibility/support checks use \Cref{app:certificate}.

The saved first-batch upper bound $U$ encloses the Reverse cone and hence Full. In $d>3$ its center is the first-section joint analytic-center direction. In $d=3$, the saved center is the feasible midpoint between the analytic-center and enclosing-center directions. Rescale the two joint coefficient vectors to unit gradient parts, average them with equal weight, and normalize the resulting gradient part. Positive combinations preserve strict homogeneous ranking inequalities. The actual rounded coefficient vector and its ray cap are checked again. This retains the original certificate center rather than recomputing the reported bounds with a different selector. The any-center Mirror lower bound $L$ is constructed by \Cref{app:mirror-lower}. Both bounds use the same first-batch endpoints, with no new oracle evaluations.

The audit certifies $U<L$ in all 1,536 conditions at $\lambda=0.001$ and in 1,533 of 1,536 at $\lambda=0.1$. The three unresolved comparisons are all at $d=20,B=16d,\lambda=0.1$, with maximum $U/L=6.061$. Nonseparation does not establish cone equality. All first-batch caps cover the true direction. Median $U/L$ decreases with budget and increases with nuisance coefficient in every dimension. These geometric results are descriptive. Complete grouped summaries are identified by \texttt{table09\_information.csv}.

\FloatBarrier

\section{Ranking-Only Optimization on Real-Data Quadratic Tasks}\label{app:task-optimization}
\subsection{Tasks, information access, and the local model}\label{app:task-data}
The tasks are UCI Airfoil Self-Noise and Concrete Compressive Strength \citep{uci_airfoil,uci_concrete}. Each uses a fixed 80/20 split and training-only standardization. Hessian whitening is not applied. Including the unpenalized intercept gives $d=6$ and $d=9$. The reported objective is half the mean squared training residual plus a ridge penalty of weight $0.01$ on feature coefficients. Writing $R_{\rm quad}$ for the quadratic regularizer, the objective is
\[
F(\theta)=\frac{1}{2m}\norm{X\theta-y}^2+R_{\rm quad}(\theta),
\qquad R_{\rm quad}(\theta)=\frac{0.01}{2}\norm{\theta_{1,\ldots,d-1}}^2.
\]
For any center $\theta_c$ its exact expansion is
\[
F(\theta_c+x)=F(\theta_c)+\nabla F(\theta_c)^\top x+\tfrac12x^\top Hx.
\]
Thus recovery is within the even model at every center, with the data covariance unchanged.
For dataset identifier $j$, NumPy's default generator with seed $202609170+j$ permutes the rows. The first $\lfloor0.8m_{\rm total}\rfloor$ are training data. Feature and response means and population standard deviations are estimated on training rows only. The standardized response is $(y-\overline y_{\rm train})/s_{y,\rm train}$. Test RMSE is converted back to the original response scale by multiplying by $s_{y,\rm train}$. The initial feature coefficients are zero and the unpenalized intercept is one. Exact split indices and stored scaling constants are supplied in \path{dataset_split_indices.csv} and \path{training_scalings.csv}.

Only endpoint coordinates and rankings enter the online query and step decisions. The exact optimum and held-out test scores are reserved for evaluation. Hyperparameters are selected using terminal training objectives on separate development seeds. At every new optimization center, the gradient consistency model is rebuilt from scratch.

The primary optimization metric is the relative optimization gap
\[
r(\theta)=\frac{F(\theta)-F^\star}{F(\theta_0)-F^\star}.
\]
It measures the fraction of the initial optimization gap remaining. The 32-seed intervals hold the train/test split fixed and quantify algorithmic randomness. They do not establish uncertainty over alternate data splits or a task population.

\subsection{Information-layer trajectories}\label{app:task-layers}
Mirror, Reverse, and Full receive $8d$ recovery endpoints at every center and thirteen paid dyadic step candidates, including the current center. Eight updates therefore spend $8(8d+13)=488/680$ endpoints on Airfoil/Concrete, with sixteen feedback rounds. The study uses radii $1/8,1/64$ and 32 algorithm seeds per task, paired across all methods and radii, giving $2\times2\times32\times3=384$ completed trajectories.

Panel A of \Cref{tab:task-attribution} reports all four task--radius comparisons. Intervals use 20,000 paired algorithm-seed resamples. Bonferroni adjustment covers eight contrasts. Ratios compare arithmetic means of relative optimization gaps; they are not means of per-seed ratios. Concrete test RMSE is $11.0048$ for Mirror and approximately $10.9618$ for Full.
At each center, all three recovery models return unit directions. The step grid is $\{0,2^{-10},2^{-9},\ldots,2^1\}$. It is applied directly to those directions without gradient-magnitude scaling. \path{ridge_model_absolute_results.csv} gives all twelve task--radius--method rows, including mean relative optimization gaps and test RMSEs. \path{ridge_model_contrasts.csv} gives the eight paired gap contrasts and paired test-RMSE differences. All per-iteration outcomes are retained in \path{ridge_all_trajectories.csv}.

\subsection{External optimizer implementations and development}\label{app:task-baselines}
The controls are ZO-RankSGD \citep{tang2024ranking}, STP \citep{bergou2020stp}, CMA-ES \citep{hansen2001cma}, and local incumbent random search. Each method uses its own queries and returns the best paid point. ZO-RankSGD uses Gaussian ranking estimates and an ordinal line search. STP uses fixed or decreasing steps. CMA-ES uses native EvoTorch 0.6.1 GPU execution with rank fitness and its default active covariance adaptation.

For each task and resource profile, twelve configurations per method are evaluated on four development seeds. CMA-ES additionally includes four configurations with population size $\lfloor(B-1)/16\rfloor$, enabling effective use of the round cap. Selection minimizes the terminal development training objective without test-based tuning. Final evaluation uses 32 algorithm seeds, five methods, two tasks, and two resource profiles, giving 640 completed trajectories.
With $m=8d$ Gaussian samples $z_i$ and ranks $r_i\in\{0,\ldots,m-1\}$ increasing with loss, ZO-RankSGD uses $\sum_i(2r_i-m+1)z_i/\binom m2$. This estimator is not normalized before line search. Full uses a unit direction. Both evaluate $\{0,b2^{-11},\ldots,b\}$ times their direction, where $b$ is the selected step parameter. CMA-ES minimizes within-batch ranks. Its covariance update receives no cardinal losses. The configuration grids and selected task--resource settings are stored in \path{external_development_plan.json}, \path{external_cma_budget_grid.json} and \path{external_selected_configurations.json}. Development and evaluation use disjoint base seeds $1709186201$ and $1709187201$, respectively, with the dataset and replicate offsets recorded in the manifests.

The Reverse attribution control repeats the Full grid and development budget, but replaces the Full joint fit by the Mirror and reverse-order rows defining $\cC_{\rm rev}$. Exact transitive reduction removes only rows implied by retained paths. The resulting fit uses an analytic-center direction selector and the same unit-direction line search. All 192 development trajectories complete. The deterministic selection rule chooses query radius $1/64$ on Airfoil and $1/16$ on Concrete, with unit step scale, for both resource profiles. These are also the configurations selected independently for Full. The Reverse development optimum is tied across the two tested step maxima in each task. The fixed configuration-index rule selects the smaller maximum.

Confirmation uses the same 32 algorithm seeds per task and resource profile, giving 128 additional complete trajectories. Full and Reverse have identical initial centers, random direction arrays, paid first batches, and observed first-batch scores. Their later centers and queries may diverge. Before confirmation analysis, the targeted study defines one family of four Full/Reverse contrasts across terminal relative optimization gap and test RMSE, and another of sixteen Reverse/control terminal-gap contrasts across both resource profiles. No result is pooled with the earlier information-layer trajectories.

\paragraph{Independent tuning and gradient-normalization controls.}
Two separate 128-seed ridge cohorts assess tuning and gradient normalization (\Cref{app:task-inference,tab:ridge-followup-controls}). The first compares independently tuned Mirror, Reverse, and Full under a common momentum update ($p_t=\operatorname{normalize}(h_t+\beta p_{t-1})$) with equal 24-configuration budgets. The second evaluates unit-normalized ZO-RankSGD under a 12-configuration budget. The ZO-RankSGD baseline in \rev{Panel~B of \Cref{tab:task-attribution}} uses its unnormalized estimate before line search. Random-search proposal distributions and step scales are specified in companion records. Both cohorts are analyzed separately from the initial 32-seed runs.

\subsection{Endpoint and feedback-round accounting}\label{app:task-resources}

For Full and Reverse, the external comparison charges a separate initial incumbent evaluation in addition to eight recovery/step updates, giving $B=1+8(8d+13)=489/681$, one endpoint more than the information-layer trajectories (\Cref{tab:task-resource-ledger}).

\begin{table}[tbp]
\centering\small
\caption{Realized resources for external comparisons. Entries give Airfoil/Concrete endpoint and feedback-round counts under joint endpoint--round caps or an endpoint cap alone. Common caps need not be fully spent by every method.}\label{tab:task-resource-ledger}
\begin{tabular}{lrrrr}
\toprule
& \multicolumn{2}{c}{Joint caps} & \multicolumn{2}{c}{Endpoint cap only} \\
\cmidrule(lr){2-3}\cmidrule(lr){4-5}
Method & Endpoints & Rounds & Endpoints & Rounds \\
\midrule
Full & 489 / 681 & 16 / 16 & 489 / 681 & 16 / 16 \\
Reverse & 489 / 681 & 16 / 16 & 489 / 681 & 16 / 16 \\
ZO-RankSGD & 489 / 681 & 16 / 16 & 489 / 681 & 16 / 16 \\
STP & 33 / 33 & 16 / 16 & 489 / 681 & 244 / 340 \\
CMA-ES & 481 / 673 & 16 / 16 & 487 / 681 & 54 / 68 \\
Random & 481 / 673 & 16 / 16 & 489 / 681 & 17 / 17 \\
\bottomrule
\end{tabular}
\end{table}

The initial incumbent is appended to the first submitted batch before any scores are returned. That call charges all its endpoints and increments the round counter once. Each subsequent call similarly submits its entire batch before feedback. Reconstructing the original 30,848 stored batches across 640 runs confirms the endpoint totals, round totals, and first-batch incumbent in every run. The Reverse control adds 2,048 independently recomputed batches. This gives 32,896 audited batches in total. All 128 Reverse outputs are the best paid point in their history. All 128 first batches match Full exactly. \path{external_batch_ledger.csv} records the original batch index, new and cumulative endpoint counts, and preceding feedback dependency. \path{external_reverse_audit.json} records the targeted audit.

Sequential STP stores the incumbent and spends only two new endpoints per update, so it cannot use the endpoint cap within sixteen rounds. Its poor round-capped result is therefore not evidence of poor sequential FE efficiency. CMA-ES includes round-aware populations, and ZO-RankSGD spends essentially the same FE budget as Full, supplying batch-capable controls.

\subsection{Effects and statistical scope}\label{app:task-inference}
For the Full/control comparison in \Cref{fig:new-external}, 40,000 paired algorithm-seed resamples give percentile intervals with Bonferroni adjustment over sixteen contrasts spanning both resource profiles. The eight displayed joint-cap intervals lie below one. Full/ZO-RankSGD terminal mean-gap ratios are $0.358\,[0.168,0.845]$ on Airfoil and $0.646\,[0.498,0.837]$ on Concrete. Full/CMA-ES ratios are $0.0658\,[0.0328,0.1226]$ and $0.223\,[0.177,0.289]$, respectively. \path{external_all_contrasts.csv} reports the numerical results for every contrast in the comparison family.

Panel B of \Cref{tab:task-attribution} reports the independently tuned Full/Reverse and Reverse/control terminal-gap ratios. The Full/Reverse family has four contrasts, spanning two tasks and terminal gap/test RMSE. Its additional test-RMSE ratios are $0.99970\,[0.99925,1.00011]$ on Airfoil and $1.00000\,[0.99907,1.00095]$ on Concrete. These intervals use the same 40,000 paired resamples and four-contrast Bonferroni adjustment. None of the four Full/Reverse intervals excludes one. Reverse/control intervals use their separate sixteen-contrast family spanning both resource profiles, not the four-contrast adjustment.

Absolute results and both attribution families are identified by \path{external_reverse_all_groups.csv}, \path{external_full_reverse_contrasts.csv}, and \path{external_reverse_control_contrasts.csv}. Development selections and audit receipts appear in \path{external_reverse_audit.json}. Under endpoint-only caps, Full and Reverse complete eight updates in sixteen rounds. STP and CMA-ES use additional rounds and attain lower mean gaps (\Cref{tab:round-tradeoff,tab:task-resource-ledger}). These endpoint-only comparisons are descriptive.

Mean Airfoil/Concrete runtimes are $0.62/0.82$ seconds for Full, $0.43/0.61$ for Reverse, and $0.10/0.05$ for ZO-RankSGD. Full and ZO-RankSGD test RMSEs are $4.7102/10.9544$ and $4.7190/11.0208$, respectively. Reverse's test RMSEs are $4.7116/10.9544$. Reverse's mean whole-run times are 30\% and 26\% lower than Full's in these implementations. The separately executed runs provide implementation-level whole-run timing comparisons.

The external-run timer starts before loading the processed task and constructing GPU tensors, and ends after optimization, diagnostic metrics, and compressed query-record output. It includes direction fitting, rank sorting, query generation, and the algorithm's objective evaluations. It is a whole-run measurement, not a certificate-only or optimizer-kernel timer. No warm-up is subtracted. Objective calls synchronize the GPU before and after evaluation and also have a separately recorded timer. Independent audit programs run afterward and are excluded. Whole-run times and the separately reported component costs in \Cref{app:computation} therefore have different scopes.

\paragraph{Two separate 128-seed ridge control cohorts.}
Two distinct follow-up cohorts (\Cref{tab:ridge-followup-controls}) each evaluate 128 new algorithm seeds per task under the same 16-round, 489/681-FE caps: a CPU momentum cohort and a GPU no-momentum cohort. Neither is pooled with the initial 32-seed studies or with the other.
\emph{Cohort 1 (common momentum update, CPU):} Mirror, Reverse, and Full all update via $p_t=\operatorname{normalize}(h_t+\beta p_{t-1})$, where $h_t$ is the recovered unit direction and $p_{t-1}$ the previous accepted unit update. Memory starts at zero and resets after a zero step. Each method uses step grid $\{0,2^{-11},\ldots,1\}$ and independently tunes 24 configurations ($a\in\{1,1/4,1/16,1/64,1/256,1/1024\}$, $\beta\in\{0,0.25,0.5,0.75\}$) on four development seeds per task. Selected configurations are frozen before evaluation on 128 new seeds. Intervals use 40,000 paired bootstrap resamples with 16-contrast Bonferroni adjustment. Reverse and Full achieve substantially lower mean gaps than tuned Mirror, with no statistically established Full-over-Reverse improvement.
\emph{Cohort 2 (unit-normalized ZO-RankSGD, GPU, no momentum):} The control retains native Gaussian queries and the ranking estimator, normalizing $\widehat g$ before line search. The methods use a 12-configuration tuning budget with 4 development seeds, followed by evaluation on 128 new seeds with 10-contrast Bonferroni intervals. Full and Reverse attain lower gaps than unit-normalized ZO-RankSGD. Both 128-seed cohorts are separate from the 32-seed A100 timing study in \Cref{app:computation}.

\begin{table}[tbp]
\centering\small
\caption{Two separate 128-seed follow-up ridge cohorts (16 rounds, 489/681 FEs; unpooled from each other, from the original studies in \rev{Panels~A and~B of \Cref{tab:task-attribution}}, and from \Cref{tab:ridge-time-to-target}). Cohort 1 uses a common momentum update on CPU ($p_t=\operatorname{normalize}(h_t+\beta p_{t-1})$, 24 configs, 16-contrast Bonferroni 95\% CIs). Cohort 2 uses no momentum on GPU and compares Full and Reverse against unit-normalized ZO-RankSGD (12 configs, 10-contrast Bonferroni 95\% CIs).}\label{tab:ridge-followup-controls}
\begin{tabular}{@{}lll@{}}
\toprule
Comparison / Metric & Airfoil & Concrete \\
\midrule
\multicolumn{3}{@{}l}{\emph{A. Common momentum update (mean terminal relative optimization gap)}}\\
Mirror (momentum) & $0.000252325$ & $0.004662082$ \\
Reverse (momentum) & $0.000044192$ & $0.001942583$ \\
Full (momentum) & $0.000045350$ & $0.001905175$ \\
Reverse / Mirror (95\% CI) & $0.175\ [0.125,0.249]$ & $0.417\ [0.380,0.456]$ \\
Full / Mirror (95\% CI) & $0.180\ [0.126,0.262]$ & $0.409\ [0.373,0.448]$ \\
\midrule
\multicolumn{3}{@{}l}{\emph{B. Original method / Unit-normalized ZO-RankSGD (95\% CI)}}\\
Full / Unit-normalized ZO-RankSGD & $0.374\ [0.259,0.539]$ & $0.674\ [0.601,0.757]$ \\
Reverse / Unit-normalized ZO-RankSGD & $0.378\ [0.264,0.533]$ & $0.667\ [0.593,0.752]$ \\
\bottomrule
\end{tabular}
\end{table}

The task interface combines within-batch ranks with an ordinal incumbent update. The evaluator privately compares each new batch's best point with the stored incumbent and exposes the coordinates of the best paid point, but not its score. The saved-batch audit independently verifies that every final output is the best evaluated point across the complete history.

\FloatBarrier

\section{Non-Even Logistic Recovery and Local-Step Controls}\label{app:logistic-tasks}

UCI Blood Transfusion Service Center (Blood) and Banknote Authentication (Banknote) ($d=4,5$) use nonzero intercepts to avoid the linear-plus-even expansion at zero \citep{uci_blood,uci_banknote}. Thirty-two dataset--seed units across four radii and two noise levels yield 256 conditions. Zero-error Full, robust Reverse, and robust Full share $8d$ antipodal endpoints, including initialization, in one recovery round. Supplied error envelopes preserve true-direction feasibility and cap coverage in every robust fit. Zero-error Full excludes the true direction in thirteen conditions, including three without score noise.

At $\nu=10^{-5}$, mean robust Full outer-cap radii decrease as $a$ decreases from $1/4$ to $1/32$, then increase at $1/128$, consistent with competing remainder and score-error allowances (\Cref{fig:new-noneven,app:logistic-arms}). The paid initializer enforces a checked factor-two error-inflation bound; its certified domain is narrower than the full robust-localization regime (\Cref{app:robust-chart,app:initialization-results}). \Cref{app:logistic-tasks,app:validity-results} give calibration, local-step controls, and synthetic validity tests.

Recovery validity and downstream local-step outcomes use the same task cohort but different paid observation stages. Validity is assessed before step feedback. The downstream comparisons evaluate one recovery-and-step procedure at each fixed center.
\subsection{Tasks and nonzero recovery centers}\label{app:logistic-data}
The two classification datasets are UCI Blood Transfusion Service Center and Banknote Authentication \citep{uci_blood,uci_banknote}. Fixed 80/20 splits and training-only standardization yield $d=4,5$ parameters including the intercept. In Blood, exact affine-duplicate detection using training data removes Monetary, which is 250 times Frequency. Every compared method uses the same resulting features. Feature coefficients start at zero, and the unpenalized intercept starts at $0.5$.

For a scalar logit $t$,
\[
\log(1+e^t)-yt=\log\!\left(2\cosh(t/2)\right)+(1/2-y)t.
\]
At the all-zero parameter, the logistic objective is itself linear-plus-even. The nonzero-center experiment avoids this special expansion. The numerical audit reports a nonzero odd remainder in every tested condition at the prescribed nonzero center.

\subsection{Public smoothness, third derivative, and gradient calibration}\label{app:logistic-calibration}
Write $X_i$ for augmented training features, $m$ for training size, $\sigma$ for the logistic sigmoid, and $\overline y$ for the training-label mean. The supplied outward-rounded constants are
\[
L=\frac{1}{4m}\sum_i\norm{X_i}^2+0.01,\qquad
M_3=\frac{1}{6\sqrt3\,m}\sum_i\norm{X_i}^3,
\qquad 0<g_{\min}<|\sigma(0.5)-\overline y|.
\]
They use task metadata and the prescribed center, not the realized full gradient or ranking margins. The intercept derivative at this center is exactly $\sigma(0.5)-\overline y$, so its absolute value is a gradient-norm lower bound. It need not remain a lower bound after moving the center.

For the scalar logistic loss, its second derivative is at most $1/4$, and the absolute third derivative is at most $1/(6\sqrt3)$. The implemented training objective is
\[
F(\theta)=\frac1m\sum_i\bigl[\log(1+e^{X_i^\top\theta})-y_iX_i^\top\theta\bigr]
+\frac{0.01}{2}\norm{\theta_{1,\ldots,d-1}}^2.
\]
The unpenalized intercept is the last coordinate, so the regularizer contributes exactly $0.01$ to the Hessian operator-norm bound. Summing feature-tensor bounds gives the displayed $L$ and $M_3$. The stored $(L,M_3,g_{\min})$ values are
\[
\begin{aligned}
\text{Blood:}\quad &(1.0100000000000005,1.1487774432367968,0.3850011272219215),\\
\text{Banknote:}\quad &(1.2600000000000013,1.2690047823460824,0.16940553815718729).
\end{aligned}
\]
Interval arithmetic verifies that the first two are rounded outward and that the last is strictly below the absolute intercept derivative.

With physical radius $a$, stored score error bounded by $\nu$, and numerical allowance $10^{-10}$, the fitted shell-normalized shared-error envelope is
\begin{equation}\label{eq:logistic-envelope}
\rho_a^{\rm fit}=\frac{M_3a^2/6+(\nu+10^{-10})/a}{g_{\min}}.
\end{equation} In common physical-score coordinates it is $a\rho_a^{\rm fit}=[M_3a^3/6+\nu+10^{-10}]/g_{\min}$. The two representations must not be interchanged when retaining several shells. The cubic term controls odd remainder, not even Hessian anisotropy. A valid error contract alone does not verify the fast-localization sufficient condition on $\tau_a$.

For $\nu>0$, each endpoint receives one stored uniform error in $[-\nu,\nu]$. It keeps that realized error in the ranking. The resulting observations remain transitive. The public constants are expressed on the task's calibrated score scale, so the experiment is not scale-free under an arbitrary unknown monotone score transformation.
We verify the numerical contract with 192-bit Arb ball arithmetic, treating the stored float64 features, labels, and regularizer coefficient as exact inputs. The audit encloses dot products, exponentials, logarithms, sums, and the calibration formulas. For recovery, it evaluates the ideal points $\theta_0\pm aW_i$, thereby including coordinate-construction rounding in the score-error check. All 256 conditions pass. The largest recovery-batch clean-score discrepancy upper bound is $3.88\times10^{-16}<10^{-10}$, and every observed pair satisfies its gradient-normalized shared-error allowance. Detailed receipts are in \path{logistic_interval_and_centers.csv} and \path{logistic_interval_summary.json}.

\subsection{Matched recovery-stage validity}\label{app:logistic-arms}
Sixteen query/noise seeds per task are reused across $a\in\{1/4,1/8,1/32,1/128\}$ and $\nu\in\{0,10^{-5}\}$. The resulting 256 conditions reuse 32 dataset--seed units. For the recovery-stage comparison, zero-error Full, robust Reverse, and robust Full receive identical $8d$ antipodal endpoints, with paid axis initialization included in that first batch. This stage costs one feedback round and $32/40$ endpoints on Blood/Banknote. The local-step arms in \Cref{app:logistic-step-controls} add a separate paid round. No observations from that additional round enter the recovery-stage validity or cap summaries.

Every robust initialization meets the checked factor-two chart criterion, and every robust fit retains the true direction and every cap covers it. The zero-error Full model excludes the true direction in thirteen conditions, including three with $\nu=0$. The fitted cell and its outer cap are checked separately. Coverage alone would not establish that a misspecified model contains the true direction.

\Cref{fig:new-noneven} shows all radius/noise conditions. In panel (b), solid and dashed lines denote zero and nonzero score-error bounds, respectively. For $\nu=10^{-5}$, the mean caps on Blood/Banknote are $0.2164/0.6842$ at $a=1/4$. They are $0.0780/0.0833$ at $a=1/32$ and $0.0807/0.1089$ at $a=1/128$. \Cref{fig:new-noneven}(c) compares the two terms in the implemented allowance. The diagonal marks equality. Their score-error/odd-remainder ratio is $6(\nu+10^{-10})/(M_3a^3)$; the factor $g_{\min}$ cancels. Ratios below/above one indicate odd-remainder/score-error dominance. A crossing of one is consistent with the observed small-radius upturn, but it does not define a universally optimal radius. All 256 conditions, including every zero-error exclusion, remain in these recovery summaries. The corresponding recovery and local-step records are identified by \path{logistic_all_outputs.csv} and \path{logistic_absolute_results.csv}.

\subsection{Matched-budget local-step comparisons and physical-step control}\label{app:logistic-step-controls}
\paragraph{What is matched.}
The four local-step arms are zero-error Full, robust Reverse, robust Full, and ZO-RankSGD. The first three reuse the identical $8d$ antipodal recovery endpoints described above. ZO-RankSGD receives $8d$ Gaussian endpoints with matched root-mean-square radius. Later step candidates can differ with the fitted direction. Comparisons with ZO-RankSGD instead match the budget but use different recovery-query geometry.

Each arm then evaluates thirteen candidates on the fixed dyadic multiplier grid $\{0,2^{-10},2^{-9},\ldots,2^1\}$ with unit base scale. This adds a second feedback round, giving $8d+13$ endpoints, or 45 on Blood and 53 on Banknote. All 1,024 runs complete, and the zero-error arm's exclusions remain recorded. Zero-error Full and both robust models use unit directions. ZO-RankSGD uses its unnormalized complete-ranking Gaussian estimate. Identical multipliers therefore do not imply identical physical step lengths. No arm uses $L$ or $g_{\min}$ to tune the step grid. The robust models use calibration to construct their feasible sets. The reported interval audit also checks the saved line-search and ZO-RankSGD observations under the numerical contract in \Cref{app:logistic-calibration}. Those checks do not change the recovery-stage sample count.

\paragraph{Unit-normalized physical-step control.}
A separate diagnostic normalizes the saved ZO-RankSGD estimate $\widehat g$ when $\norm{\widehat g}>10^{-12}$ and otherwise returns the current center. The fallback is unused in all 256 conditions, where $\norm{\widehat g}\in[0.881,1.388]$. Reusing the recovery output, it evaluates the thirteen prescribed physical step lengths with paired line-search noise. Its per-condition recovery-plus-step cost remains $8d+13$ endpoints and two rounds. The diagnostic adds 3,328 step evaluations across the cohort. It changes only the norm entering the step grid, not the angular error or the recovery-query geometry.

\subsection{Direction accuracy and downstream loss are separate outcomes}\label{app:logistic-inference}
Forty thousand paired seed resamples provide percentile intervals with Bonferroni correction across the original 64-contrast family. This family contains two controls (robust Reverse and ZO-RankSGD), two outcomes, and sixteen task--radius--noise cells. These are ratios of arithmetic means, not means of per-seed ratios. Repeated radii and noise levels do not create new independent units. The comparison counts appear in \Cref{tab:logistic-contrast-counts}.

\begin{table}[!htbp]
\centering\small
\caption{Non-even logistic direction and task comparisons. Counts give task--radius--noise cells whose Bonferroni-adjusted 95\% ratio intervals favor robust Full. They are not success probabilities on new tasks.}\label{tab:logistic-contrast-counts}
\begin{tabularx}{\linewidth}{@{}X>{\centering\arraybackslash}X>{\centering\arraybackslash}X@{}}
\toprule
Robust Full compared with & Angular MSE intervals below one & Task-loss intervals below one \\
\midrule
ZO-RankSGD & 16 / 16 & 8 / 16 \\
Robust Reverse & 6 / 16 & 2 / 16 \\
\bottomrule
\end{tabularx}
\end{table}

With matched recovery endpoints, the robust Full/Reverse results provide limited evidence of a local-step benefit. The remaining intervals do not establish equivalence. \path{logistic_all_64_contrasts.csv} supplies every ratio and its marginal and adjusted interval. \path{logistic_absolute_results.csv} gives all 64 task--radius--noise--method summaries, including absolute MSE and training/test loss. All 1,024 individual outputs are retained in \path{logistic_all_outputs.csv}.

The unit-normalized ZO-RankSGD control forms a separate diagnostic family of sixteen task-loss contrasts. Using 40,000 paired seed resamples, Bonferroni-adjusted intervals favor robust Full in $8/16$ cells; none favors the control. All sixteen point ratios are below one. The same eight cells remain below one under the 64-comparison threshold as a conservative sensitivity check, not a joint old-plus-new family. Positive normalization preserves angular MSE exactly and improves the pooled Blood regularized training objective from $0.52057$ to $0.51343$, while robust Full still has the lower objective (\Cref{tab:ranksgd-scale-control}). Estimator and recovery-query differences remain after this norm control.

\begin{table}[!htbp]
\centering\small
\caption{Physical-step control for ZO-RankSGD. The first three result columns give arithmetic mean post-step regularized training objectives $F$ over 128 matched conditions per task. The last gives the robust Full/unit-normalized ZO-RankSGD ratio. All 3,328 new candidates pass a 192-bit interval audit. The maximum clean-score discrepancy upper bound is $2.45\times10^{-16}$.}\label{tab:ranksgd-scale-control}
\begin{tabularx}{\linewidth}{@{}l*{4}{>{\centering\arraybackslash}X}@{}}
\toprule
Task & Robust Full & ZO-RankSGD & Unit-normalized ZO-RankSGD & Robust Full / unit-normalized \\
\midrule
Blood & 0.48848 & 0.52057 & 0.51343 & 0.9514 \\
Banknote & 0.36632 & 0.38180 & 0.38322 & 0.9559 \\
\bottomrule
\end{tabularx}
\end{table}

\FloatBarrier

\section{Validity Tests under Odd Remainders and Score Noise}\label{app:validity-results}
The cubic experiment isolates invalid retention. Separate calibrated cohorts test shared-error recovery, smooth-radius calibration, and paid observable initialization. Non-even logistic recovery and supplementary local-step controls are evaluated in \Cref{app:logistic-tasks}.
\subsection{Odd remainders: containment, coverage, and history}\label{app:odd-experiments}
\subsubsection{Protocol and retention policies}
This branch reuses the 96 independent target--axis units in \Cref{app:scope-protocol}, with 32 at each $d\in\{3,8,20\}$. It crosses $N/d\in\{8,32\}$ with $\mu\in\{0,0.003,0.03,0.3\}$, giving 768 conditions, each evaluated under four policies. Together with the 2,688 quadratic and 576 quartic conditions, the shared structural cohort comprises 4,032 conditions and 6,336 policy branches, not additional independent targets. All planned denominators are retained, including the 337 empty retained models.

The odd family uses \Cref{eq:odd-equivalence} with $E(x)=\norm x^2/2$. Large shell evaluates all $N$ endpoints at radius one. Small shell evaluates them at radius $1/8$. Retained evaluates the first $N/2$ endpoints at radius one and the remaining $N/2$ at radius $1/8$, then fits their complete cumulative ranking. Restart uses the same two batches and pays the same $N$ evaluations, but its final fit retains only the second batch's $N/2$ endpoints. Both two-round designs use uniform directions. They isolate the effect of retained observations, not an adaptive query-width policy. In particular, the single-shell shifted-direction order is independently checked on every saved odd design rather than assumed from ideal normalization.

\subsubsection{Diagnosing an empty retained model}
To distinguish a solver failure from model inconsistency, reconstruct pure-gradient rows $a_i^\top v\ge0$ implied by the observed ranking. A nonnegative dependence satisfies $\sum_i w_i a_i=0$, $w_i\ge0$, and $\sum_iw_i=1$. Every feasible $v$ must then satisfy $a_i^\top v=0$ for all positive-weight rows. If those rows span $\R^d$, only $v=0$ is possible, so the direction set is empty even under the weak closed model.

For all 337 empty retained models, an independent checker reconstructs the original implied rows and verifies nonnegative weights, exact cancellation, and full positive-support rank. These certificates establish nonzero-gradient impossibility, not merely failure to find a strict interior.

\subsubsection{Containment, coverage, and retained history}

All 2,304 single-shell branches, including Restart's retained small batch, have the same exact ranking as the corresponding shifted-gradient even instance. Its direction is feasible and covered in every branch. These are consistency and coverage checks. They do not imply that finite rankings uniquely determine that direction. The true gradient direction at the center is the normalized stored $u$ and is assessed separately in \Cref{tab:scope-odd}.

A certificate encloses the fitted model. It can cover the true direction even after the model has excluded it, because the outer bound may be loose. \Cref{tab:scope-odd} shows why containment and coverage require separate diagnostics. On a fixed shell, more observations can expose model mismatch by narrowing the feasible set around a biased direction. Restart keeps fewer constraints than Small shell, so its higher containment or coverage count does not establish better precision.

\begin{table}[tbp]
\centering\small
\caption{Odd-remainder recovery diagnostics. Each cell gives true-direction feasibility/coverage counts out of 32 planned targets. Coverage counts only returned certificates; empty direction sets count as neither feasible nor covered. Restart pays $N$ endpoints but retains only $N/2$.}\label{tab:scope-odd}
\begin{tabular*}{\linewidth}{@{\extracolsep{\fill}}rrrcccc@{}}
\toprule
$d$ & $N/d$ & $\mu$ & Large shell & Small shell & Retained & Restart \\
\midrule
3 & 8 & $0$ & 32 / 32 & 32 / 32 & 32 / 32 & 32 / 32 \\
3 & 8 & $0.003$ & 32 / 32 & 32 / 32 & 32 / 32 & 32 / 32 \\
3 & 8 & $0.03$ & 10 / 22 & 32 / 32 & 23 / 27 & 32 / 32 \\
3 & 8 & $0.3$ & 0 / 0 & 29 / 31 & 0 / 0 & 32 / 32 \\
\addlinespace
3 & 32 & $0$ & 32 / 32 & 32 / 32 & 32 / 32 & 32 / 32 \\
3 & 32 & $0.003$ & 5 / 9 & 32 / 32 & 21 / 28 & 32 / 32 \\
3 & 32 & $0.03$ & 0 / 0 & 22 / 31 & 0 / 0 & 30 / 32 \\
3 & 32 & $0.3$ & 0 / 0 & 2 / 3 & 0 / 0 & 16 / 28 \\
\addlinespace
8 & 8 & $0$ & 32 / 32 & 32 / 32 & 32 / 32 & 32 / 32 \\
8 & 8 & $0.003$ & 10 / 32 & 32 / 32 & 21 / 32 & 32 / 32 \\
8 & 8 & $0.03$ & 0 / 29 & 30 / 32 & 2 / 30 & 32 / 32 \\
8 & 8 & $0.3$ & 0 / 0 & 4 / 32 & 0 / 0 & 24 / 32 \\
\addlinespace
8 & 32 & $0$ & 32 / 32 & 32 / 32 & 32 / 32 & 32 / 32 \\
8 & 32 & $0.003$ & 0 / 7 & 24 / 32 & 0 / 19 & 29 / 32 \\
8 & 32 & $0.03$ & 0 / 0 & 1 / 32 & 0 / 0 & 13 / 32 \\
8 & 32 & $0.3$ & 0 / 0 & 0 / 1 & 0 / 0 & 0 / 31 \\
\addlinespace
20 & 8 & $0$ & 32 / 32 & 32 / 32 & 32 / 32 & 32 / 32 \\
20 & 8 & $0.003$ & 0 / 32 & 30 / 32 & 11 / 32 & 31 / 32 \\
20 & 8 & $0.03$ & 0 / 27 & 11 / 32 & 0 / 16 & 23 / 32 \\
20 & 8 & $0.3$ & 0 / 0 & 0 / 32 & 0 / 0 & 5 / 32 \\
\addlinespace
20 & 32 & $0$ & 32 / 32 & 32 / 32 & 32 / 32 & 32 / 32 \\
20 & 32 & $0.003$ & 0 / 0 & 8 / 32 & 0 / 0 & 21 / 32 \\
20 & 32 & $0.03$ & 0 / 0 & 0 / 32 & 0 / 0 & 0 / 32 \\
20 & 32 & $0.3$ & 0 / 0 & 0 / 0 & 0 / 0 & 0 / 32 \\
\bottomrule
\end{tabular*}
\end{table}

Of 768 retained conditions, 302 admit the true direction and 466 exclude it. Every exclusion is already present in the first batch. In 274 of those 466, the new small batch alone admits the true direction, yet keeping the old constraints prevents its re-entry (\Cref{tab:scope-retention}). Of the 466, 337 have no feasible nonzero gradient at all. The remaining 129 return another direction. Thus reducing the new shell's cubic bias and restoring the retained model's validity are distinct operations.

\begin{table}[tbp]
\centering\small
\caption{True-direction exclusion under retained history. Each odd-remainder coefficient has 192 conditions (96 targets, two budgets). Membership columns record (first batch, new batch, retained model), with 1 denoting inclusion; the three patterns exhaust the data. The final column counts independently certified empty retained sets, a subset of the two true-direction exclusion patterns.}\label{tab:scope-retention}
\begin{tabular*}{\linewidth}{@{\extracolsep{\fill}}lrrrr@{}}
\toprule
$\mu$ & $(1,1,1)$ & $(0,1,0)$ & $(0,0,0)$ & Empty retained set \\
\midrule
$0$ & 192 & 0 & 0 & 0 \\
$0.003$ & 85 & 92 & 15 & 43 \\
$0.03$ & 25 & 105 & 62 & 105 \\
$0.3$ & 0 & 77 & 115 & 189 \\
\midrule
Total & 302 & 274 & 192 & 337 \\
\bottomrule
\end{tabular*}
\end{table}

\subsection{Matched error-contract experiments}\label{app:error-experiments}
This study uses 96 independent targets, with 32 at each $d=3,8,20$. Targets, odd axes, and stored noise are reused across $N/d=16,32$ and three perturbations. This design gives 576 matched conditions. Each condition gives a first fit at $N/2$ and a cumulative fit at $N$ endpoints. Solver parameters are identical throughout. Target, query, and noise seeds are in the companion manifest.

The objective and stored observations are
\[
f(x)=u^\top x+\norm x^2/2+\mu(q^\top x)\norm x^2,
\qquad y(x)=f(x)+\xi(x),
\]
where $u$ is a normalized Gaussian vector, $q$ is a normalized orthogonalized Gaussian vector, and each endpoint error is uniform on $[-\nu,\nu]$. The three $(\mu,\nu)$ settings are odd-only $(0.003,0)$, noise-only $(0,0.0003)$, and combined $(0.003,0.00001)$. The combined setting uses both error sources but a lower noise amplitude than the noise-only setting, so the comparison is a sensitivity check rather than a factorial decomposition. A stored endpoint retains its original error. The first $d$ unsigned queries are the standard axes, and the rest are uniform spherical draws. The first half use physical radius one. The second half use radius $1/8$ for odd-only and combined conditions, and radius one for noise-only conditions. Every axis and antipode counts toward the budget.

The decoder receives coordinates, integer ranks, and public budgets
\[
\omega_i=\frac{\mu\,\overline{\norm{x_i}}^{\,3}+\nu+10^{-12}}{0.99},
\]
rounded outward, where $\overline{\norm{x_i}}$ is a checked upper norm bound. The supplied gradient lower bound is $0.99$. The small numerical allowance covers the stored normalizations. It is not adjusted using the realized projection on $q$. Exact binary-rational polynomial scores plus stored errors determine the order. Objective parameters and cardinal scores enter only the evaluator and audits. The second fit retains the same chart and each endpoint's error budget.

An independent rational checker reconstructs observations, verifies every Full shared-error inequality and true physical-score witness, and recomputes supports and caps. True-direction membership in the zero-error model is checked independently on the same endpoints and ranks, isolating the model rather than acquisition.

Retained/first MSE ratios compare $N$ with $N/2$ endpoints, not equal budgets. Marginal 95\% percentile intervals use 20,000 paired target resamples, retaining pairing across budgets and perturbations within each dimension.

All 1,152 fits across the 576 conditions converge and are feasible; all caps cover the true direction. The zero-error model excludes the true direction in 333 final conditions (\Cref{tab:bounded-errors}). Every marginal refinement interval lies below one, and the largest chart-envelope inflation is $1.047994$, below the factor two allowed by the theory. Caps shrink in 574/576 conditions. Outer radii about changing centers need not be monotone even for nested sets.

\begin{table}[t]
\centering\small\setlength{\tabcolsep}{3pt}
\caption{Shared-error recovery. Each row uses 32 independent targets. Error settings are Odd $(\mu,\nu)=(0.003,0)$, Noise $(0,0.0003)$, and Combined $(0.003,0.00001)$. Combined uses less noise than Noise. All robust fits are feasible, and all caps cover the true direction. ``Excluded'' counts true-direction exclusions by the zero-error model on the same final observations. MSE is the final mean squared angular error (rad$^2$). Ratios compare final $N$-endpoint with first-batch $N/2$-endpoint MSE. Brackets give marginal paired 95\% percentile confidence intervals (CIs). Cap radii are final medians in radians.}\label{tab:bounded-errors}
\begin{tabular*}{\linewidth}{@{\extracolsep{\fill}}rrlcccc@{}}
\toprule
$d$ & $N/d$ & Error & Excluded & MSE & Ratio [95\% CI] & Median cap\\
\midrule
3 & 16 & Odd & $3/32$ & $1.30\times10^{-4}$ & $0.158\ [0.0883,0.291]$ & $0.0359$\\
3 & 16 & Noise & $2/32$ & $6.35\times10^{-5}$ & $0.0814\ [0.0516,0.130]$ & $0.0173$\\
3 & 16 & Combined & $3/32$ & $1.30\times10^{-4}$ & $0.158\ [0.0886,0.291]$ & $0.0361$\\
3 & 32 & Odd & $11/32$ & $1.90\times10^{-5}$ & $0.180\ [0.0845,0.382]$ & $0.0113$\\
3 & 32 & Noise & $3/32$ & $3.43\times10^{-6}$ & $0.0540\ [0.0225,0.117]$ & $0.00490$\\
3 & 32 & Combined & $11/32$ & $1.90\times10^{-5}$ & $0.180\ [0.0845,0.383]$ & $0.0114$\\
8 & 16 & Odd & $17/32$ & $4.78\times10^{-5}$ & $0.229\ [0.182,0.295]$ & $0.0347$\\
8 & 16 & Noise & $7/32$ & $7.37\times10^{-6}$ & $0.0431\ [0.0297,0.0626]$ & $0.0149$\\
8 & 16 & Combined & $17/32$ & $4.75\times10^{-5}$ & $0.228\ [0.181,0.294]$ & $0.0350$\\
8 & 32 & Odd & $31/32$ & $6.11\times10^{-6}$ & $0.238\ [0.176,0.315]$ & $0.0123$\\
8 & 32 & Noise & $15/32$ & $6.76\times10^{-7}$ & $0.0917\ [0.0600,0.142]$ & $0.00628$\\
8 & 32 & Combined & $31/32$ & $5.80\times10^{-6}$ & $0.227\ [0.169,0.296]$ & $0.0132$\\
20 & 16 & Odd & $31/32$ & $1.54\times10^{-5}$ & $0.255\ [0.210,0.312]$ & $0.0338$\\
20 & 16 & Noise & $24/32$ & $2.22\times10^{-6}$ & $0.0646\ [0.0519,0.0806]$ & $0.0163$\\
20 & 16 & Combined & $31/32$ & $1.47\times10^{-5}$ & $0.243\ [0.202,0.295]$ & $0.0370$\\
20 & 32 & Odd & $32/32$ & $1.30\times10^{-6}$ & $0.0760\ [0.0659,0.0870]$ & $0.00889$\\
20 & 32 & Noise & $32/32$ & $2.77\times10^{-7}$ & $0.125\ [0.102,0.155]$ & $0.00869$\\
20 & 32 & Combined & $32/32$ & $1.43\times10^{-6}$ & $0.0834\ [0.0706,0.0994]$ & $0.0118$\\
\bottomrule
\end{tabular*}
\end{table}

\subsection{A non-even smooth objective with a public radius}\label{app:smooth-experiment}
To test \Cref{cor:smooth-fixed}, use the strongly convex shifted quartic
\[
f(x)=u_0^\top x+\norm x^2/2+0.01\sum_{j=1}^d(z_j+x_j)^4,
\qquad \norm{u_0}=1,\quad z_j\in[-1,1].
\]
At the origin, its actual gradient is $g=u_0+0.04z^{\odot3}$, not $u_0$, and its odd remainder is $0.04\sum_jz_jx_j^3$. The gradient is $1.48$-Lipschitz throughout the unit ball. At the recovery center, $\norm g\ge1-0.04\sqrt d>0.8$ for $d\le20$; this is not a gradient-norm lower bound throughout the ball. Supply $L=1.5$, $g_{\min}=0.5$, and $a=0.01g_{\min}/(8L\sqrt d)$. The divisor eight is an experimental choice, not a calibrated theorem constant.

Four independent $(u_0,z)$ draws per $d=3,8,20$ give 12 one-round conditions, each paying $32d$ endpoints including standard-axis pairs. After rescaling by $a$, the decoder receives coordinates, exact polynomial ranks, and outward-rounded budgets $La\norm{W_i}^2/(2g_{\min})$. All fits meet the convergence threshold, and exact witness/support checks confirm true-direction retention and cap coverage. Caps range from $0.00409$ to $0.00795$ rad and point errors from $0.000355$ to $0.003014$ rad.

\subsection{Observable initialization and retained recovery}\label{app:initialization-results}\label{sec:new-initialization}
The observable initializer constructs a shared-error chart from paid observations without a supplied direction cap. Its first batch has $d$ coordinate pairs and $3d$ uniform pairs, for $8d$ endpoints. It uses the stronger axis bound and residual-checked refinements in \Cref{app:robust-chart}. All refinements are numerical. Failure to certify the factor-two criterion is reported without fitting.

The archived summary reports 116 conditions from 25 target units at $d=8,20,50$. It partitions them into 68 conditions on thirteen units and 48 conditions on twelve additional units. The per-unit reconstruction of the 68-condition subset resides in the companion records and is not recoverable from the manuscript alone.
\Cref{tab:new-initialization} summarizes the independently enumerated 48-condition subset. All reported conditions complete under one common rule, with maximum chart inflation $1.99428$. All 232 first- and second-batch fits converge and are feasible; every cap covers the true direction. A second batch adds $8d$ uniform endpoints at physical radius $1/8$, retaining the original chart and all prior constraints. Every final cap is tighter than its first cap. The manifest and \path{initialization_all_conditions.csv} retain the cohort identifiers.

\begin{table}[tbp]
\centering\small
\caption{Observable robust initialization and retention. Four independent targets per dimension (twelve total) are reused across two error budgets $d\rho\in\{0.40,0.55\}$ and two observation modes, giving sixteen conditions per row. Cap radii are medians in radians. The MSE ratio compares $16d$ with $8d$ endpoints, so budgets differ. All charts satisfy the factor-two criterion; every fit contains the true direction and every cap covers it.}\label{tab:new-initialization}
\begin{tabular}{rrrrr}
\toprule
$d$ & Conditions & First cap & Final cap & Final / first MSE \\
\midrule
8 & 16 & 0.6872 & 0.05834 & 0.03206 \\
20 & 16 & 0.4927 & 0.05129 & 0.03786 \\
50 & 16 & 0.3822 & 0.04092 & 0.02303 \\
\bottomrule
\end{tabular}
\end{table}

The implementation stores physical coordinates $x_i=aW_i$ and physical budgets $B_i=\rho a^3$, with the odd remainder bounded by $0.99B_i$. This bound excludes the linear gradient contribution to the antisymmetric score. Reducing the radius from one to $1/8$ therefore reduces stored budgets by $1/512$. Dividing coordinates and budgets by $a$ gives the shell-normalized factor $1/64$. This synthetic continuation has no fixed-amplitude score-noise term, whose normalized allowance would instead grow as $\nu/(a\norm g)$. The two observation modes use zero odd error or bounded shared odd error under the same public budget.

Continuation here jointly increases observations and decreases the physical radius. Its loss and cap changes do not isolate either intervention or establish an advantage over a matched final-only design. The finite success counts do not establish initialization throughout the full parameter region of \Cref{cor:robust}.

\subsection{Task-structured gradient calibration for noisy squared loss}\label{app:task-initialization}
On the standardized Airfoil and Concrete squared-loss tasks, set the unpenalized intercept to one and all feature coefficients to zero. Then
\[
\partial_{\rm intercept}F=1-\overline y_{\rm train}.
\]
Training-response centering supplies the public $g_{\min}=0.99$ without revealing the full gradient. This is a center-specific metadata-derived bound, not a lower bound along an optimization trajectory. With physical radius $a=1/512$, score-error bound $\nu=10^{-4}$, and shell-normalized allowances $(\nu+10^{-10})/(ag_{\min})$, all 64 initializations across 32 query/noise seeds on each task satisfy the factor-two criterion. The maximum inflation is $1.99547$. Independent checks confirm retention and coverage in every robust fit.

These fixed-center initialization experiments are separate from the noiseless moving-center ridge trajectories in \Cref{sec:new-tasks}. A 192-bit interval audit of all 3,840 initial observations verifies the $10^{-10}$ numerical allowance, including coordinate-construction error. The maximum clean-score discrepancy upper bound is $4.76\times10^{-16}$. The audit also verifies every public intercept bound and shared-error inequality.
The stored training responses give exact rational intercept derivatives $1-\overline y_{\rm train}$. \path{ridge_public_gradient_proof.json} contains these fractions and their comparison with $0.99$. \path{ridge_initialization_conditions.csv} identifies all 64 conditions. \path{ridge_interval_checks.csv} gives the interval checks. \path{protocols.json} records query/noise seeds and the factor-two gate. Numerical chart refinements use no new oracle feedback. Their computation is recorded separately from the paid first batch and subsequent step evaluations.

\FloatBarrier

\section{Computational Cost and Higher-Dimensional Recovery}\label{app:computation}

We measure fitting, certification, independent checking, and memory on controlled one-round instances.
\subsection{Stored constraints and numerical work}
For $n\ge d$ distinct physical pairs and a strict ranking, the exact-even model has $p=d+n-1$ variables after fixing the potential gauge and $2n-1$ adjacent-ranking inequalities. Tied adjacencies require opposing inequalities, and duplicate physical pairs share offsets. These changes preserve the stated asymptotic orders. Each dense QR Newton iteration costs $O(np^2)$ arithmetic operations and $O(np+p^2)$ storage. These per-iteration bounds exclude iteration counts and rational bit complexity. The shared-error chart has $p=d+2n-1$, the same ranking-row convention, and linear-size coordinate/error boxes, with the same dense-factorization order.

For the strict, distinct-pair convention, the gradient certificate initially forms at most $n^2$ Mirror/Reverse rows, using $O(n^2d)$ storage; tied relations require both orientations. Zero-error transitive reduction preserves their implications, not necessarily the lengths of individual cycles in the full comparison multigraph. Reachability uses $O(n^2)$ additional storage and $O(n^3)$ Boolean/dense arithmetic. The implementation solves $2d+1$ support objectives with $d-1$ gradient-only section variables. Shared-error certificates have a larger chart and explicitly retain their allowances without zero-error reduction.

Barrier-based sparse dual proposals are accepted only with nonnegative multipliers and the exact residual allowance in \Cref{eq:dual-main}. Singular or unsuccessful proposals are discarded. Proposal optimality is unnecessary. The crossover recipe is retained in the companion implementation notes.

\subsection{A controlled dimension--budget study}
Twelve independent targets, four at each $d=20,50,100$, use matched one-round budgets $N/d=8,16$. Independent normalized Gaussians give $u,z$ in \Cref{eq:experiment-target}, with $\lambda=10^{-3}$. Uniform unit-shell antipodal queries use nested budget prefixes. All 24 conditions use the same exact ordering, float64 QR fit, dual crossover, rational supports, and independent tangent completion.

Each worker uses one allocated logical CPU and one shared A100 80-GB GPU. In \Cref{tab:computation}, CUDA-synchronized certificate time includes dual proposals and rational checks. Separate recheck time includes independent verification and tangent completion. GPU allocation excludes cached reserved memory. Peak host resident memory, including Python/CUDA, is at most 1.27 GiB. These timings characterize shared hardware, not isolated-device throughput.

\Cref{tab:computation} shows that certification alone, and certification plus independent checking, exceed fitting time in every row. Rechecking alone does not exceed fitting at $d=20$. Larger budgets tighten caps but increase GPU workspace. At $d=100,N=1600$, about 640,000 Mirror/Reverse rows reduce to 836--849. These measurements are not a matched Mirror-versus-Reverse timing comparison and cannot establish a computational advantage of Reverse.

\paragraph{Evaluation cost and feedback latency.}
With $P$ parallel workers, $n_r$ paid endpoints in round $r$, endpoint time $t_{\rm eval}$ and feedback latency $t_{\rm fb}$, an illustrative total time is $\sum_r\lceil n_r/P\rceil t_{\rm eval}+Rt_{\rm fb}+T_{\rm sort}+T_{\rm decode}$. Here $T_{\rm decode}$ includes fitting and any certification or independent checking selected for the operating mode; a fit-only mode omits the latter stages. Thus $N=\sum_r n_r$ and $R$ describe different resources, neither of which includes sorting or decoding.

\paragraph{Separate 32-seed A100 time-to-target study.}
Independent of the two 128-seed cohorts in \Cref{tab:ridge-followup-controls}, a 32-seed study on a warmed A100 measures time to a common target without momentum (\Cref{tab:ridge-time-to-target}; two timing repetitions per paired seed). Active time covers online query generation, evaluation, sorting, fitting/initialization, and line search, excluding data loading, trajectory logging, offline tuning, and audits. All three methods in \Cref{tab:ridge-time-to-target} reach relative optimization gap $0.03$ across all $32/32$ seeds within 16 rounds: Reverse takes fewer rounds ($4.000$ vs.\ $5.125/5.688$) at higher active time ($46.892/62.103\text{ ms}$ vs.\ $3.294/3.556\text{ ms}$) than native ZO-RankSGD. We substitute mean active times and round counts into the additive model $T=T_{\rm active}+R\ell$, where $\ell$ is a hypothetical equal delay per round. The crossover delays are $\ell_\star=38.75\text{ ms}$ (Airfoil) and $34.69\text{ ms}$ (Concrete) at the $0.03$ no-momentum target. These projected crossovers apply to this target and these configurations; the study did not measure network delays. At the stricter target $0.001$ on Airfoil, Full, Reverse, Mirror, native ZO-RankSGD, and unit-normalized ZO-RankSGD reach the threshold on $32/32, 32/32, 29/32, 21/32$, and $22/32$ seeds within 16 rounds. On Concrete, $0/32$ succeed. Times conditioned on success do not estimate unconditional expected hitting times. Optimization runs made zero certificate calls. In a separate 8-seed/task first-batch diagnostic, median ridge certification times are $1.365\text{ s}$ (Airfoil) and $0.874\text{ s}$ (Concrete), plus $0.0093/0.0239\text{ s}$ for independent rechecking and completion.

\end{document}